\documentclass{article}

\usepackage{microtype}
\usepackage{graphicx}
\usepackage{booktabs} % for professional tables

\usepackage{hyperref}

\usepackage[accepted]{icml2024}

\usepackage{amsmath}
\usepackage{amssymb}
\usepackage{mathtools}
\usepackage{amsthm}

\usepackage{amsfonts}
\usepackage{multirow}
\usepackage{multicol}

\usepackage{color}
\usepackage{algorithm}
\usepackage{algorithmic}

\usepackage{xcolor,colortbl}
\definecolor{LightCyan}{rgb}{0.88,1,1}

\usepackage{subfig}
\usepackage{tcolorbox}

\usepackage[capitalize,noabbrev]{cleveref}

\theoremstyle{plain}
\newtheorem{theorem}{Theorem}[section]

\newtheorem{lemma}[theorem]{Lemma}

\theoremstyle{definition}

\newtheorem{assumption}[theorem]{Assumption}
\theoremstyle{remark}
\newtheorem{remark}[theorem]{Remark}

\usepackage[textsize=tiny]{todonotes}

\icmltitlerunning{ Faster Adaptive Decentralized Learning Algorithms }

\begin{document}

\twocolumn[
\icmltitle{ Faster Adaptive Decentralized Learning Algorithms }

% It is OKAY to include author information, even for blind
% submissions: the style file will automatically remove it for you
% unless you've provided the [accepted] option to the icml2024
% package.

% List of affiliations: The first argument should be a (short)
% identifier you will use later to specify author affiliations
% Academic affiliations should list Department, University, City, Region, Country
% Industry affiliations should list Company, City, Region, Country

% You can specify symbols, otherwise they are numbered in order.
% Ideally, you should not use this facility. Affiliations will be numbered
% in order of appearance and this is the preferred way.
\icmlsetsymbol{equal}{*}

\begin{icmlauthorlist}
\icmlauthor{Feihu Huang}{1,2}
\icmlauthor{Jianyu Zhao}{1}
%\icmlauthor{Firstname3 Lastname3}{comp}
%\icmlauthor{Firstname4 Lastname4}{sch}
%\icmlauthor{Firstname5 Lastname5}{yyy}
%\icmlauthor{Firstname6 Lastname6}{sch,yyy,comp}
%\icmlauthor{Firstname7 Lastname7}{comp}
%%\icmlauthor{}{sch}
%\icmlauthor{Firstname8 Lastname8}{sch}
%\icmlauthor{Firstname8 Lastname8}{yyy,comp}
%%\icmlauthor{}{sch}
%%\icmlauthor{}{sch}
\end{icmlauthorlist}

\icmlaffiliation{1}{College of Computer Science and Technology, Nanjing University of Aeronautics and Astronautics, Nanjing, China}
\icmlaffiliation{2}{MIIT Key Laboratory of Pattern Analysis and Machine Intelligence, Nanjing, China}

\icmlcorrespondingauthor{Feihu Huang}{huangfeihu2018@gmail.com}

% You may provide any keywords that you
% find helpful for describing your paper; these are used to populate
% the "keywords" metadata in the PDF but will not be shown in the document
\icmlkeywords{Machine Learning, ICML}

\vskip 0.3in
]

% this must go after the closing bracket ] following \twocolumn[ ...

% This command actually creates the footnote in the first column
% listing the affiliations and the copyright notice.
% The command takes one argument, which is text to display at the start of the footnote.
% The \icmlEqualContribution command is standard text for equal contribution.
% Remove it (just {}) if you do not need this facility.

\printAffiliationsAndNotice{}  % leave blank if no need to mention equal contribution
%\printAffiliationsAndNotice{\icmlEqualContribution} % otherwise use the standard text.

\begin{abstract}
Decentralized learning recently has received increasing attention in machine learning due to its advantages in implementation simplicity and system robustness, data privacy. Meanwhile,
the adaptive gradient methods show superior performances in many machine learning tasks such as training neural networks.
Although some works focus on studying decentralized optimization algorithms with adaptive learning rates,
these adaptive decentralized algorithms still suffer from high sample complexity.
To fill these gaps, we propose a class of faster adaptive decentralized algorithms (i.e.,
AdaMDOS and AdaMDOF) for distributed nonconvex stochastic and finite-sum optimization, respectively. Moreover, we provide a solid convergence analysis framework for our methods.
In particular, we prove that our AdaMDOS obtains a near-optimal sample complexity of $\tilde{O}(\epsilon^{-3})$
for finding an $\epsilon$-stationary solution of nonconvex stochastic optimization.
Meanwhile, our AdaMDOF obtains a near-optimal sample complexity of $O(\sqrt{n}\epsilon^{-2})$ for finding an $\epsilon$-stationary solution of nonconvex finite-sum optimization, where $n$ denotes the sample size. To the best of our knowledge, our AdaMDOF algorithm is the first adaptive decentralized algorithm for nonconvex finite-sum optimization.
Some experimental results demonstrate  efficiency of our algorithms.
\end{abstract}

\section{Introduction}
With the rapidly increasing dataset sizes and the high dimensionality of the machine learning problems, training large-scale machine learning models has been increasingly concerned. Clearly, training large-scale models by a single centralized machine has become inefficient and unscalable. Due to addressing the efficiency and scalability challenges, recently distributed machine learning optimization
is widely studied. In particular, decentralized optimization~\citep{lian2017can} has received increasing attention in recent years in machine learning due to liberating the centralized agent with large
communication load and privacy risk.
In the paper, we study decentralized learning algorithms to solve the distributed \textbf{stochastic} problem over a communication network $G=(V,E)$, defined as
\begin{align} \label{eq:1}
 \min_{x \in \mathbb{R}^d} & \ F(x)\equiv\frac{1}{m}\sum_{i=1}^m f^i(x), \ f^i(x)= \mathbb{E}_{\xi^i}\big[f^i\big(x;\xi^i\big)\big]
\end{align}
where for any $i\in[m]$, $f^i(x)$ denotes the objective function in
$i$-th client, which is a differentiable and possibly nonconvex function. Here $\xi^i$ for any $i\in[m]$ is an independent random variable following an unknown distribution $\mathcal{D}^i$, and for any $i,j\in [m]$ possibly $\mathcal{D}^i \neq \mathcal{D}^j$. $G=(V,E)$ is a communication network including $m$ computing agents, where
any agents $i,j\in V$ can communicate only if $(i,j)\in E$.
 Meanwhile, we also consider decentralized learning algorithms for solving  the distributed \textbf{finite-sum} problem over a communication network $G=(V,E)$, defined as
\begin{align} \label{eq:2}
 \min_{x \in \mathbb{R}^d} & \ F(x)\equiv\frac{1}{m}\sum_{i=1}^mf^i(x), \ f^i(x)=\frac{1}{n}\sum_{k=1}^n f^i_k(x)
\end{align}
where $f^i_k(x) = f^i(x;\xi^i_k)$ for $k=1,2\cdots,n$. Here $\{\xi^i_k\}_{k=1}^n$ can be seen as $n$ samples drawn from distribution $\mathcal{D}^i$ for $i=1,2,\cdots,m$.
In fact, Problems~(\ref{eq:1}) and (\ref{eq:2}) frequently appear many machine learning applications such as training Deep Neural Networks (DNNs)~\citep{lian2017can} and reinforcement learning~\citep{chen2022sample}.

\begin{table*}
  \centering
  \caption{ \textbf{Sample} and \textbf{Communication} complexities comparison of the representative \textbf{adaptive} decentralized stochastic algorithms for finding
  an $\epsilon$-stationary point of Problem~(\ref{eq:1}) or~(\ref{eq:2}), i.e., $\mathbb{E}\|\nabla F(x)\|\leq \epsilon$
  or its equivalent variants. \textbf{Note that} the AdaRWGD~\citep{sun2022adaptive} relies on the random walk instead of parallel framework used in other algorithms. For fair comparison, here we do not consider some specific cases such as sparse stochastic gradients. }
  \label{tab:1}
   \resizebox{0.95\textwidth}{!}{
\begin{tabular}{c|c|c|c|c}
  \hline
  % after \\: \hline or \cline{col1-col2} \cline{col3-col4} ...
  \textbf{Problem}  & \textbf{Algorithm} & \textbf{Reference} & \textbf{Sample Complexity} & \textbf{Communication Complexity}   \\ \hline
 \multirow{4}*{\textbf{Stochastic}} & DADAM  & \cite{nazari2022dadam}  & $O(\epsilon^{-4})$ & $O(\epsilon^{-4})$ \\   \cline{2-5}
   & AdaRWGD  & \cite{sun2022adaptive}  & $O(\epsilon^{-4})$ & $O(\epsilon^{-4})$ \\   \cline{2-5}
  & DAMSGrad/DAdaGrad  & \cite{chen2023convergence} & $O(\epsilon^{-4})$ & $O(\epsilon^{-4})$   \\   \cline{2-5}
  & AdaMDOS  & Ours & {\color{red}{$\tilde{O}(\epsilon^{-3})$}} & {\color{red}{$O(\epsilon^{-3})$}} \\  \hline \hline
  \textbf{Finite-Sum} & AdaMDOF  & Ours & {\color{red}{$O(\sqrt{n}\epsilon^{-2})$}} & {\color{red}{$O(\epsilon^{-2})$}}  \\  \hline
\end{tabular}
 }
 \vspace*{-8pt}
\end{table*}

Many decentralized stochastic gradient-based algorithms recently have been developed
to solve the above stochastic Problem~(\ref{eq:1}).
For example, \cite{lian2017can} proposed an efficient decentralized stochastic gradient descent (D-PSGD) algorithm, which integrates average consensus with local-SGD steps and outperforms the standard centralized SGD methods. Due to the presence of inconsistency under non-i.i.d. setting, some variants of
D-PSGD~\citep{tang2018d,xin2021improved} are studied to handle the data heterogeneity issue, e.g., $D^2$ method~\citep{tang2018d} by storing
previous status, and GT-DSGD method~\citep{xin2021hybrid} by using gradient tracking technique~\citep{xu2015augmented}. Subsequently, \cite{sun2020improving,pan2020d} proposed some accelerated
decentralized SGD algorithms (i.e., D-GET and D-SPIDER-SFO) by using variance reduced
gradient estimator of SARAH/SPIDER~\citep{nguyen2017sarah,fang2018spider}, which obtain a near-optimal sample complexity of $O(\epsilon^{-3})$ for finding the stationary solution of stochastic optimization problems. To reduce large batch-size at each iteration, \cite{zhang2021gt,xin2021hybrid} proposed a class of efficient momentum-based decentralized SGD algorithms (i.e., GT-HSGD and GT-STORM) based on momentum-based variance reduced gradient estimator of ProxHSGD/STORM~\citep{cutkosky2019momentum,tran2022hybrid}, which also obtain a near-optimal sample complexity of $\tilde{O}(\epsilon^{-3})$.

Meanwhile, some decentralized stochastic gradient-based algorithms have been developed
to solve the above finite-sum  Problem~(\ref{eq:2}).
\cite{sun2020improving,xin2022fast} proposed a class of efficient decentralized algorithms for nonconvex finite-sum optimization based on variance reduced gradient estimator of SARAH~\citep{nguyen2017sarah}.
Subsequently, \cite{zhan2022efficient} presented a fast decentralized algorithm for nonconvex finite-sum optimization based on variance reduced gradient estimator of ZeroSARAH~\citep{li2021zerosarah}
without computing multiple full gradients.

It well known that the adaptive gradient methods show superior performances
n many machine learning tasks such as training DNNs.
More recently, \cite{nazari2022dadam,sun2022adaptive,chen2023convergence} proposed some adaptive decentralized algorithms for stochastic optimization based on the existing Adam algorithm~\citep{kingma2014adam} or its variants. However, these adaptive decentralized algorithms still suffer high sample and communication complexities in finding the stationary solution of Problem~(\ref{eq:1}) (Please see Table~\ref{tab:1}).
Naturally, there still exists an open question:
\begin{center}
\begin{tcolorbox}
\emph{Could we design adaptive decentralized algorithms with lower sample and communication complexities to solve Problems~(\ref{eq:1}) and~(\ref{eq:2}) ? }
\end{tcolorbox}
\end{center}
In the paper, to fill this gap, we affirmatively answer to this question, and
 propose a class of faster adaptive decentralized algorithms to solve Problems~(\ref{eq:1}) and~(\ref{eq:2}), respectively, based on the momentum-based variance-reduced and gradient tracking techniques.
In particular, our methods use a unified adaptive matrix to flexibly incorporate various adaptive learning rates.
Our main contributions are several folds:
\begin{itemize}
 \vspace*{-6pt}
\item[(1)] We propose a class of efficient adaptive decentralized optimization algorithms (i.e., AdaMDOS and AdaMDOF) to solve Problems~(\ref{eq:1}) and~(\ref{eq:2}), respectively, based on the momentum-based variance-reduced and gradient tracking techniques simultaneously. Moreover, we provide a convergence analysis framework for our methods.
\item[(2)] We prove that our AdaMDOS algorithm reaches the near optimal sample complexity of $\tilde{O}(\epsilon^{-3})$ for finding an $\epsilon$-stationary solution of Problem~(\ref{eq:1}), which matches the lower bound of smooth nonconvex stochastic optimization~\citep{arjevani2023lower}.
\item[(3)] We prove that our AdaMDOF algorithm reaches the near optimal sample complexity of $O(\sqrt{n}\epsilon^{-2})$ for finding an $\epsilon$-stationary solution of Problem~(\ref{eq:2}), which matches the lower bound of smooth nonconvex finite-sum optimization~\citep{fang2018spider}.
\item[(4)] We conduct some numerical experiments on training nonconvex machine learning tasks to verify the efficiency of our proposed  algorithms.
 \vspace*{-6pt}
\end{itemize}
Since our algorithms use a unified adaptive matrix including various adaptive learning rates,
our convergence analysis does not consider some specific cases such as sparse stochastic gradients.
Despite this, our adaptive algorithms still obtain lower sample and communication complexities compared to the existing adaptive decentralized algorithms.
 \vspace*{-8pt}
\section{Related Works}
 \vspace*{-4pt}
In this section, we overview some representative decentralized optimization algorithms and
adaptive gradient algorithms, respectively.
 \vspace*{-8pt}
\subsection{ Decentralized Optimization }
 \vspace*{-4pt}
Decentralized optimization is an efficient framework to collaboratively solve distributed problems by multiple worker nodes, where a worker node only needs to communicate with its neighbors at each iteration. The traditional decentralized optimization methods include some popular
algorithms such as Alternating Direction Method of Multipliers (ADMM)~\citep{boyd2011distributed}, Dual Averaging~\citep{duchi2011dual}. Subsequently, some efficient decentralized optimization algorithms have been developed, e.g., Extra~\citep{shi2015extra}, Next~\citep{di2016next}, Prox-PDA~\citep{hong2017prox}. Meanwhile, \cite{lian2017can} proposed an efficient decentralized stochastic gradient descent algorithm (i.e., D-PSGD), which shows that the decentralized SGD can outperform the parameter server-based SGD algorithms relying on high communication cost. From this, decentralized algorithms have begun to shine
in machine learning such as training DNNs.
Subsequently, \cite{tang2018d} proposed an accelerated D-PSGD (i.e., $D^2$) by using
previous status. Meanwhile, \citep{xin2021hybrid} further proposed an efficient decentralized SGD algorithm (i.e., GT-DSGD) by using gradient tracking technique~\citep{xu2015augmented}.
By using the variance-reduced techniques, some other accelerated decentralized SGD algorithms~\cite{sun2020improving,pan2020d,cutkosky2019momentum,tran2022hybrid} have been proposed, include
D-SPIDER-SFO~\citep{pan2020d} and GT-HSGD~\citep{xin2021hybrid}.
Meanwhile, \cite{nazari2022dadam} studied the decentralized version of AMSGrad~\citep{reddi2019convergence}
for online optimization.
Moreover, \cite{sun2022adaptive,chen2023convergence} developed adaptive decentralized algorithms for stochastic optimization by using local adaptive learning rates.
\vspace*{-8pt}
\subsection{ Adaptive Gradient Methods }
\vspace*{-4pt}
Adaptive gradient methods~\citep{duchi2011adaptive,kingma2014adam,loshchilov2017decoupled} recently have been successfully applied in machine learning tasks such as training DNNs.
Adam \citep{kingma2014adam} is one of popular adaptive gradient methods by using a coordinate-wise adaptive learning rate and momentum technique to accelerate algorithm, which is the default optimization tool
 for training attention models~\citep{zhang2020adaptive}.
Subsequently, some variants of Adam \citep{reddi2019convergence,chen2018convergence,guo2021novel}
have been presented to obtain a convergence guarantee under the nonconvex setting.
Due to using the coordinate-wise type of
adaptive learning rates, Adam frequently shows a bad generalization performance in training DNNs.
To improve the generalization performances, recently some adaptive gradient methods such as AdamW ~\citep{loshchilov2017decoupled}, AdaGrad~\citep{li2019convergence} and AdaBelief~\citep{zhuang2020adabelief} have been proposed.
More recently, some accelerated adaptive gradient methods \citep{cutkosky2019momentum,huang2021super,levy2021storm+,kavis2022adaptive} have been proposed based on the variance-reduced techniques.
 \vspace*{-6pt}
\section{Preliminaries}
 \vspace*{-4pt}
\subsection{Notations}
 \vspace*{-4pt}
$[m]$ denotes the set $\{1,2,\cdots,m\}$.
$\|\cdot\|$ denotes the $\ell_2$ norm for vectors and spectral norm for matrices.
$\langle x,y\rangle$ denotes the inner product of two vectors $x$ and $y$. For vectors $x$ and $y$, $x^r \ (r>0)$ denotes the element-wise
power operation, $x/y$ denotes the element-wise division and $\max(x,y)$ denotes the element-wise maximum. $I_{d}$ denotes a $d$-dimensional identity matrix.
$a_t=O(b_t)$ denotes that $a_t \leq c b_t$ for some constant $c>0$. The notation $\tilde{O}(\cdot)$ hides logarithmic terms. $\mbox{ones}(d,1)$ denotes an all-one d-dimensional vector.
 \vspace*{-6pt}
\subsection{Assumptions}
\vspace*{-4pt}
In this subsection, we give some mild assumptions on the Problems~\eqref{eq:1} and~\eqref{eq:2}.
\begin{assumption} \label{ass:1}
(\textbf{Smoothness})
For any $i \in [m]$, each component loss function $f^i(x;\xi^i)$ is $L$-smooth, such that
for all $x_1,x_2 \in \mathbb{R}^d$
\begin{align}
   \|\nabla f^i(x_1;\xi^i)-\nabla f^i(x_2;\xi^i)\| \leq L\|x_1-x_2\|.
\end{align}
\end{assumption}
Clearly, based on Assumptions \ref{ass:1}, we have
\begin{align}
 & \|\nabla F(x_1)- \nabla F(x_2)\| \nonumber \\ & = \|\frac{1}{m}\sum_{i=1}^m\big(\mathbb{E}_{\xi^i}[\nabla f^i(x_1;\xi^i)]-\mathbb{E}_{\xi^i}[\nabla f^i(x_2;\xi^i)]\big)\| \nonumber \\
 & \leq \frac{1}{m}\sum_{i=1}^m\mathbb{E}_{\xi^i}\|\nabla f^i(x_1;\xi^i)-\nabla f^i(x_2;\xi^i)\| \leq L\|x_1-x_2\|, \nonumber
\end{align}
i.e., the global function $F(x)$ is $L$-smooth as well.
\begin{assumption} \label{ass:2}
(\textbf{Sampling Oracle})
Stochastic function $f^i(x;\xi^i)$ has an unbiased stochastic gradient
with bounded variance for any $i\in [m]$, i.e.,
\begin{align}
\mathbb{E}[\nabla f^i(x;\xi^i)] = \nabla f^i(x), \ \mathbb{E}\|\nabla f^i(x;\xi^i) - \nabla f^i(x) \|^2 \leq \sigma^2. \nonumber
\end{align}
\end{assumption}
\begin{assumption} \label{ass:3}
(\textbf{Lower Bounded}) The objective function $F(x)$ is lower bounded, i.e.,  $F^* = \inf_{x\in
\mathbb{R}^d} F(x)$.
\end{assumption}
\begin{assumption} \label{ass:4}
(\textbf{Network Protocol})
The graph $G=(V,E)$ is connected and undirected, which
can be represented by a mixing matrix $W\in \mathbb{R}^{m\times m}$: 1) $W_{i,j}>0$ if
$W_{i,j}\in E$ and $W_{i,j}=0$ otherwise; 2)
$W$ is doubly stochastic such that $W=W^T$, $\sum_{i=1}^mW_{i,j}=1$ and $\sum_{j=1}^mW_{i,j}=1$; 3)
the eigenvalues of $W$ satisfy $\lambda_m \leq \cdots \leq \lambda_2 < \lambda_1=1$ and
$\nu=\max(|\lambda_2|, |\lambda_m|)<1$.
\end{assumption}
\begin{assumption} \label{ass:5}
In our algorithms, the local adaptive matrices $A^i_t\succeq \rho I_d \succ 0$ for all $i\in [m]$, $t\geq 1$ for updating the variables $x$, where $\rho>0$ is an appropriate positive number.
\end{assumption}

Assumptions~\ref{ass:1} and~\ref{ass:2} are commonly used in stochastic smooth nonconvex optimization~\citep{sun2020improving,pan2020d,cutkosky2019momentum,tran2022hybrid}.
Assumption~\ref{ass:3} ensures the feasibility of Problems~(\ref{eq:1}) and~(\ref{eq:2}).
Assumption~\ref{ass:4} shows the protocol properties of network $G=(V,E)$, which is very common
in the decentralized distributed optimization~\citep{lian2017can,xin2021hybrid}.
Assumption \ref{ass:5} imposes that each local adaptive matrix is positive definite, which is commonly used in many  adaptive gradient methods for non-distributed optimization~\cite{huang2021super,yun2021adaptive}.

\begin{algorithm}[t]
\caption{ Adaptive Momentum-Based Decentralized Optimization (\textbf{AdaMDOS}) Algorithm for \textbf{Stochastic} Optimization}
\label{alg:1}
\begin{algorithmic}[1]
\STATE {\bfseries Input:} $T>0$, tuning parameters $\{\gamma, \eta_t, \beta_t \}$, initial inputs $x^i_1\in \mathbb{R}^d$ for all $i \in [m]$; \\
\STATE {\bfseries initialize:} Set $x^i_0=\tilde{x}^i_0$ for $i \in [m]$, and draw one sample $\xi_{0}^i$
and then compute $u^i_0 = \nabla f^i(x^i_0;\xi^i_{0})$ and
$w^i_0 = \sum_{j\in \mathcal{N}_i} W_{i,j}u^j_0$ for all $i \in [m]$. \\
\FOR{$t=0$ \textbf{to} $T-1$}
\FOR{$i=1,\cdots,m$ (\textbf{in parallel}) }
\STATE Generate the adaptive matrix $A^i_t \in \mathbb{R}^{d \times d}$; \\
\textcolor{blue}{\textbf{One example} of $A^i_t$ by using update rule ($a^i_0 = 0$, $0 < \varrho < 1$, $\rho>0$.) } \\
\textcolor{blue}{ Compute $a^i_t = \varrho a^i_{t-1} + (1 - \varrho)(\nabla f^i(x^i_t;\xi^i_t))^2$, $A^i_t = \mbox{diag}(\sqrt{a^i_t} + \rho I_d)$}; \\
\STATE $\tilde{x}^i_{t+1} = \sum_{j\in \mathcal{N}_i} W_{i,j}x^j_t - \gamma (A^i_t)^{-1}w^i_t$; \\
\STATE $x^i_{t+1} =  x^i_t + \eta_t(\tilde{x}^i_{t+1}-x^i_t) $;  \\
\STATE Randomly draw a sample $\xi^i_{t+1}\sim \mathcal{D}^i$;
\STATE \textcolor{red}{$u^i_{t+1} = \nabla f^i(x^i_{t+1};\xi^i_{t+1}) + (1-\beta_{t+1})(u^i_t -\nabla f^i(x^i_t;\xi^i_{t+1}))$; } \\
\STATE $w^i_{t+1} =  \sum_{j\in \mathcal{N}_i}W_{i,j}\big(w^j_t + u^j_{t+1}-u^j_t \big)$; \\
\ENDFOR
\ENDFOR
\STATE {\bfseries Output:} Chosen uniformly random from $\{x^i_{t\geq1}\}_{i=1}^m$.
\end{algorithmic}
\end{algorithm}

\begin{algorithm}[t]
\caption{ Adaptive Momentum-Based Decentralized Optimization (\textbf{AdaMDOF}) Algorithm for \textbf{Finite-Sum} Optimization}
\label{alg:2}
\begin{algorithmic}[1]
\STATE {\bfseries Input:} $T>0$, tuning parameters $\{\gamma, \eta_t, \beta_t \}$, initial inputs $x^i_1\in \mathbb{R}^d$ for all $i \in [m]$; \\
\STATE {\bfseries initialize:} Set $x^i_0=x^i_1=\tilde{x}^i_1$, $z^i_{1,0}=z^i_{2,0}=\cdots=z^i_{n,0}=0$ and $u^i_0=w^i_0=0$ for any $i \in [m]$. \\
\FOR{$t=1$ \textbf{to} $T$}
\FOR{$i=1,\cdots,m$ (\textbf{in parallel}) }
\STATE  Randomly draw a minibatch samples $\mathcal{I}^i_t$ with $|\mathcal{I}^i_t|=b$;
\STATE \textcolor{red}{$u^i_{t} = \frac{1}{b}\sum_{k\in \mathcal{I}^i_t}\big(\nabla f^i_k(x^i_t)-\nabla f^i_k(x^i_{t-1})\big) + (1-\beta_t) u^i_{t-1} + \beta_t\Big(\frac{1}{b}\sum_{k\in \mathcal{I}^i_t}\big(\nabla f^i_k(x^i_{t-1})-z^i_{k,t-1}\big) + \frac{1}{n}\sum_{j=1}^nz^i_{j,t-1}\Big)$; } \\
\STATE $w^i_t =  \sum_{j\in \mathcal{N}_i}W_{i,j}\big(w^j_{t-1} + u^j_t-u^j_{t-1} \big)$;  \\
\STATE Generate the adaptive matrix $A^i_t \in \mathbb{R}^{d \times d}$; \\
\textcolor{blue}{\textbf{One example} of $A^i_t$ by using update rule ($a^i_0 = 0$, $0 < \varrho < 1$, $\rho>0$.) } \\
\textcolor{blue}{ Compute $a^i_t = \varrho a^i_{t-1} + (1 - \varrho)\big(\frac{1}{b}\sum_{k\in \mathcal{I}^i_t}\nabla f^i_k(x^i_t)\big)^2$, $A^i_t = \mbox{diag}(\sqrt{a^i_t} + \rho I_d )$}; \\
\STATE $\tilde{x}^i_{t+1} = \sum_{j\in \mathcal{N}_i} W_{i,j}x^j_t - \gamma (A^i_t)^{-1}w^i_t$; \\
\STATE $x^i_{t+1} =  x^i_t + \eta_t(\tilde{x}^i_{t+1}-x^i_t) $;  \\
\STATE $z^i_{k,t} = \nabla f^i_k(x_t)$ for $k\in \mathcal{I}^i_t$ and $z^i_{k,t} = z^i_{k,t-1}$ for $k \notin \mathcal{I}^i_t$.
\ENDFOR
\ENDFOR
\STATE {\bfseries Output:} Chosen uniformly random from $\{x^i_{t\geq 1}\}_{i=1}^m$.
\end{algorithmic}
\end{algorithm}

\section{ Adaptive Momentum-Based Decentralized Algorithms }
In this section, we propose a class of efficient adaptive momentum-based decentralized algorithms to
solve Problems~\eqref{eq:1} and ~\eqref{eq:2}, respectively, which build  on the momentum-based and gradient tracking techniques.
\subsection{ AdaMDOS Algorithm for Stochastic Optimization}
In this subsection, we propose a faster adaptive momentum-based decentralized (AdaMDOS) algorithm for the
stochastic Problem~(\ref{eq:1}) over a network, which builds on
the variance-reduced momentum technique of STORM~\citep{cutkosky2019momentum,tran2022hybrid} and gradient tracking technique~\citep{xu2015augmented}. In particular, our AdaMDOS algorithm also uses the momentum iteration and unified adaptive learning rate to update variable.
Algorithm \ref{alg:1} provides the algorithmic framework of our AdaMDOS algorithm.

At the line 5 of Algorithm~\ref{alg:1}, we generate an adaptive matrix based on the historical stochastic gradients $\big\{\nabla f^i(x^i_l;\xi^i_l)\big\}_{1\leq l\leq t}$.
And we give an Adam-like adaptive learning rate, defined as
\begin{align}
& a^i_t  = \varrho a^i_{t-1} + (1 - \varrho)(\nabla f^i(x^i_t;\xi^i_t))^2 \nonumber \\
& \quad = \sum_{l=1}^t (1-\varrho)\varrho^{t-l}(\nabla f^i(x^i_l;\xi^i_l))^2, \nonumber \\
& A_t^i = \mbox{diag}(\sqrt{a^i_t} + \rho I_d), \label{eq:exp1}
\end{align}
clearly, we have $A_t^i \succeq \rho I_d$, which satisfies Assumption~\ref{ass:5}.
Besides one example~(\ref{eq:exp1}), we can also generate many adaptive matrices satisfying the above Assumption~\ref{ass:5}.
e.g., the Barzilai-Borwein-like adaptive matrix, defined as
\begin{align} \label{eq:a3}
  & a_t^i = \frac{|\langle x^i_t-x^i_{t-1}, \nabla f^i(x^i_t;\xi^i_t)-\nabla f^i(x^i_{t-1};\xi^i_t)\rangle|}{\|x^i_t-x^i_{t-1}\|^2}+ \rho, \nonumber \\
  & A_t^i = a_t^i I_d\succeq \rho I_d.
\end{align}
At the line 6 of Algorithm~\ref{alg:1}, each client updates the local variable $x^i$ based on adaptive matrix $A^i_t$ and momentum-based gradient estimator $w^i_t$:
\begin{align}
 \tilde{x}^i_{t+1} = \sum_{j\in \mathcal{N}_i} W_{i,j}x^j_t - \gamma (A^i_t)^{-1}w^i_t,
\end{align}
where the constant $\gamma>0$. Here $\mathcal{N}_i=\{j\in V \ | \ (i,j)\in E, j=i\}$
denotes the neighborhood of the $i$-th client. Here each client communicates with its neighbors
to update the variable $x$.
Then we further use the momentum iteration to update the variable $x$ at the line 7 of Algorithm~\ref{alg:1}:
\begin{align}
 x^i_{t+1} =  x^i_t + \eta_t(\tilde{x}^i_{t+1}-x^i_t),
\end{align}
where $\eta_t\in (0,1)$.

At lines 8-9 of Algorithm \ref{alg:1}, each client
uses the variance-reduced momentum-based technique~\citep{tran2022hybrid,cutkosky2019momentum}
to update the stochastic gradients by using local data $\xi^i_{t+1}$: for $i\in [m]$
\begin{align}
u^i_{t+1} & = \nabla f^i(x^i_{t+1};\xi^i_{t+1}) \nonumber \\
& \quad + (1-\beta_{t+1})(u^i_t -\nabla f^i(x^i_t;\xi^i_{t+1})),
\end{align}
where $\beta_{t+1}\in (0,1)$.
At the line 10 of our Algorithm \ref{alg:1}, then each client communicates with its neighbors
to compute gradient estimators $w^i_{t+1}$, defined as
\begin{align}
w^i_{t+1} =  \sum_{j\in \mathcal{N}_i}W_{i,j}\big(w^j_t + u^j_{t+1}-u^j_t \big),
\end{align}
which uses gradient tracking technique~\citep{xu2015augmented,di2016next} to reduce the consensus error.
Thus, the local stochastic gradient estimator $w^i_{t+1}$ can track the directions of global gradients.

\subsection{ AdaMDOF Algorithm for Finite-Sum Optimization}
In this subsection, we propose a faster adaptive momentum-based decentralized (AdaMDOF) algorithm for distributed
finite-sum problem~(\ref{eq:2}) over a network, which builds on
the variance-reduced momentum technique of ZeroSARAH~\citep{li2021zerosarah} and gradient tracking technique.
Algorithm \ref{alg:2} provides the algorithmic framework of our AdaMDOF algorithm.

Algorithm~\ref{alg:2} is fundamentally similar to Algorithm~\ref{alg:1}, differing primarily in its application of the variance-reduced momentum technique from ZeroSARAH~\citep{li2021zerosarah}
to update the stochastic gradients by using local data: for $i\in [m]$
\begin{align}
u^i_{t} & = \frac{1}{b}\sum_{k\in \mathcal{I}^i_t}\big(\nabla f^i_k(x^i_t)-\nabla f^i_k(x^i_{t-1})\big) + (1-\beta_t) u^i_{t-1} \nonumber \\
& \quad + \beta_t\Big(\frac{1}{b}\sum_{k\in \mathcal{I}^i_t}\big(\nabla f^i_k(x^i_{t-1})-z^i_{k,t-1}\big) + \frac{1}{n}\sum_{j=1}^nz^i_{j,t-1}\Big), \nonumber
\end{align}
where $\beta_{t+1}\in (0,1)$. Here the ZeroSARAH technique can be seen as the combination of SARAH~\citep{nguyen2017sarah} and SAGA~\citep{defazio2014saga} techniques.

\section{Convergence Analysis}
In this section, under some mild assumptions, we provide the convergence properties of our AdaMDOS and
AdaMDOF algorithms for Problems~(\ref{eq:1}) and~(\ref{eq:2}), respectively.
All related proofs are provided in the following Appendix.
For notational simplicity, let $\bar{x}_t=\frac{1}{m}\sum_{i=1}^m x^i_t$ for all $t\geq 1$.

\subsection{ Convergence Properties of our AdaMDOS Algorithm }
For AdaMDOS algorithm, we define a useful Lyapunov function, for any $t\geq 1$
\begin{align}
\Omega_t & = \mathbb{E} \Big[ F(\bar{x}_t) + (\lambda_{t-1}-\frac{9\gamma\eta}{2\rho})\|\bar{u}_{t-1} - \overline{\nabla f(x_{t-1})}\|^2 \nonumber \\
& \ + \chi_{t-1}\frac{1}{m}\sum_{i=1}^m \|u^i_{t-1} - \nabla f^i(x^i_{t-1})\|^2 \nonumber \\
& \ + (\theta_{t-1}-\frac{19\gamma\eta L^2}{4\rho})\frac{1}{m}\sum_{i=1}^m\|x^i_{t-1}-\bar{x}_{t-1}\|^2 \nonumber \\
& \ +(\vartheta_{t-1}-\frac{\gamma\eta}{4\rho})\frac{1}{m}\sum_{i=1}^m\|w^i_{t-1}-\bar{w}_{t-1}\|^2  \nonumber \\
& \ + \frac{\rho\gamma\eta}{6} \frac{1}{m}\sum_{i=1}^m\|g^i_{t-1}\|^2\Big], \nonumber
\end{align}
where $g^i_t = (A^i_t)^{-1}w^i_t$, $\overline{\nabla f(x_t)}= \frac{1}{m}\sum_{i=1}^m \nabla f^i(x^i_t)$, $\chi_t\geq 0$, $\lambda_t\geq \frac{9\gamma\eta}{2\rho}$,
$\theta_t\geq \frac{29\gamma\eta L^2}{6\rho}$, $\vartheta_t\geq \frac{\gamma\eta}{4\rho}$ and $\eta_t=\eta$ for all $t\geq0$.

\begin{theorem} \label{th:1}
 Suppose the sequences $\big\{\{x^i_t\}_{i=1}^m\big\}_{t=1}^T$ be generated from Algorithm~\ref{alg:1}.
 Under the above Assumptions~\ref{ass:1}-\ref{ass:5}, and let $\eta_t=\eta$, $0<\beta_t\leq1$ for all $t\geq 0$, $\gamma\leq \min\big(\frac{\rho(1-\nu^2)}{48\theta_t},\frac{3\rho(1-\nu^2)\theta_t}{58L^2}\big)$, $\eta\leq \min\big(\frac{\rho\sqrt{1-\nu^2}}{4L\gamma\sqrt{3(1+\nu^2)}\sqrt{H_t}},\frac{\sqrt{\rho(1-\nu^2)\theta_t}}{2 L\sqrt{\gamma(3+\nu^2)}\sqrt{H_t}}\big)$ with $H_t=\frac{9}{2\beta_t}+\frac{8\nu^2}{(1-\nu)^2}$ for all $t\geq1$, we have
\begin{align}
& \frac{1}{T}\sum_{t=1}^T \mathbb{E}\|\nabla F(\bar{x}_t)\| \\
& \leq \frac{1}{T}\sum_{t=1}^T\frac{1}{m}\sum_{i=1}^m\mathbb{E}[ \|\nabla F(x^i_t)\| + L\|\bar{x}_t-x^i_t\|] \nonumber \\
 &  \leq \Big(\frac{6\sqrt{G}}{\sqrt{T}}\!+\!\frac{12\sigma}{\rho}\sqrt{\frac{1}{T}\sum_{t=1}^TH_t\beta^2_t}\Big)
 \sqrt{\frac{1}{T}\sum_{t=1}^T\frac{1}{m}\sum_{i=1}^m\mathbb{E}\|A^i_t\|^2},  \nonumber
\end{align}
where $G= \frac{F(\bar{x}_1)-F^*}{\rho\gamma\eta}+\big( \frac{4\nu^2}{\rho^2(1-\nu)}
 +\frac{9}{2\rho^2\beta_0}-\frac{9}{2\rho^2}\big)\sigma^2 $.
\end{theorem}

\begin{remark}
Based on Assumption~\ref{ass:1},
if using Barzilai-Borwein-like adaptive matrix~(\ref{eq:a3}), we have
$\sqrt{\frac{1}{T}\sum_{t=1}^T\frac{1}{m}\sum_{i=1}^m\mathbb{E}\|A^i_t\|^2}\leq L+\rho$.

Let $\beta_t = \frac{1}{T^{2/3}}$ for all $t\geq1$, we have
\begin{align}
 H_t = \frac{9}{2\beta_t}+\frac{8\nu^2}{(1-\nu)^2}=O(T^{2/3}),
\end{align}
and then we can obtain $\sqrt{\frac{1}{T}\sum_{t=1}^TH_t\beta^2_t}=O(\frac{1}{T^{1/3}})$.
We set $\theta_t=\theta \geq \frac{29\gamma\eta L^2}{6\rho}$ for all $t\geq1$.
Meanwhile, we can set $\rho=O(1)$, $\gamma=O(1)$ and $\eta=O(\frac{1}{T^{1/3}})$.
 Then we obtain $G=O(T^{1/3})$ and
\begin{align}
  & \frac{1}{T}\sum_{t=1}^T \mathbb{E}\|\nabla F(\bar{x}_t)\| \nonumber \\
  & \leq O(\frac{1}{T^{1/3}}+\frac{\sigma}{T^{1/3}})
  \sqrt{\frac{1}{T}\sum_{t=1}^T\frac{1}{m}\sum_{i=1}^m\mathbb{E}\|A^i_t\|^2}.
\end{align}
Since $\sqrt{\frac{1}{T}\sum_{t=1}^T\frac{1}{m}\sum_{i=1}^m
  \mathbb{E}\|A^i_t\|^2}=O(1)$, let
\begin{align}
 \frac{1}{T}\sum_{t=1}^T \mathbb{E}\|\nabla F(\bar{x}_t)\| \leq O(\frac{1}{T^{1/3}}+\frac{\sigma}{T^{1/3}}) \leq \epsilon,
\end{align}
we have $T= O(\epsilon^{-3})$. Since our
AdaMDOS algorithm only use one sample at each iteration, it obtains a near-optimal sample complexity of $1\cdot T=O(\epsilon^{-3})$ for finding an $\epsilon$-stationary point of Problem~(\ref{eq:1}), which matches the lower bound of smooth nonconvex stochastic optimization~\citep{arjevani2023lower}.
\end{remark}

\begin{assumption} \label{ass:6}
(\textbf{Lipschitz Continuity})
For any $i \in [m]$, each component loss function $f^i(x;\xi^i)$ is $M$-Lipschitz continuity, such that
for all $x\in \mathbb{R}^d$
\begin{align}
   \|\nabla f^i(x;\xi^i)\| \leq M.
\end{align}
\end{assumption}
Assumption~\ref{ass:6} is commonly used in the adaptive gradient algorithms~\citep{reddi2019convergence,chen2018convergence,guo2021novel,sun2022adaptive,chen2023convergence}.
\begin{remark}
Based on Assumption~\ref{ass:6},
if using Adam-like adaptive matrix given in Algorithm~\ref{alg:1}, we have
$\sqrt{\frac{1}{T}\sum_{t=1}^T\frac{1}{m}\sum_{i=1}^m\mathbb{E}\|A^i_t\|^2}\leq M+\rho$.

Based on the above Theorem~\ref{th:1}, our
AdaMDOS algorithm still obtains a near-optimal sample (gradient) complexity of $O(\epsilon^{-3})$ for finding an $\epsilon$-stationary point of Problem~(\ref{eq:1}).
\end{remark}

\subsection{ Convergence Properties of our AdaMDOF Algorithm }
For AdaMDOF algorithm,  we first define a useful Lyapunov function for any $t\geq 1$
\begin{align}
& \Phi_t = \mathbb{E}\Big[ F(\bar{x}_t) + (\lambda_{t-1}-\frac{9\gamma\eta}{2\rho})\|\bar{u}_{t-1} - \overline{\nabla f(x_{t-1})}\|^2 \nonumber \\
& + \chi_{t-1}\frac{1}{m}\sum_{i=1}^m \|u^i_{t-1}\! -\! \nabla f^i(x^i_{t-1})\|^2 \!+\! \frac{\rho\gamma\eta}{6} \frac{1}{m}\sum_{i=1}^m\|g^i_{t-1}\|^2 \nonumber \\
& + \alpha_{t-1}\frac{1}{m}\sum_{i=1}^m\frac{1}{n}\sum_{k=1}^n\|\nabla f^i_k(x^i_{t-1})-z^i_{k,t-1}\|^2 \nonumber \\
& + (\theta_{t-1}-\frac{19\gamma\eta L^2}{4\rho})\frac{1}{m}\sum_{i=1}^m\|x^i_{t-1}-\bar{x}_{t-1}\|^2 \nonumber \\
& +(\vartheta_{t-1}-\frac{3\gamma\eta}{4\rho})\frac{1}{m}\sum_{i=1}^m\|w^i_{t-1}-\bar{w}_{t-1}\|^2\Big], \nonumber
\end{align}
where $\alpha_t\geq 0$, $\chi_t\geq 0$, $\lambda_t\geq\frac{9\gamma\eta}{2\rho}$, $\theta_t\geq \frac{29\gamma\eta L^2}{6\rho}$,$\vartheta_t\geq\frac{3\gamma\eta}{4\rho}$ and $\eta_t=\eta$ for all $t\geq0$.

\begin{theorem}  \label{th:2}
Suppose the sequences $\big\{\{x^i_t\}_{i=1}^m\big\}_{t=1}^T$ be generated from Algorithm~\ref{alg:2}.
 Under the above Assumptions~\ref{ass:1}-\ref{ass:5}, and let $\eta_t=\eta$, $0<\beta_t\leq1$ for all $t\geq 0$, $\gamma\leq \min\big(\frac{\rho(1-\nu^2)}{48\theta_t},\frac{3\rho(1-\nu^2)\theta_t}{58L^2}\big)$, $\eta\leq \min\big(\frac{\rho\sqrt{1-\nu^2}}{2L\gamma\sqrt{6(1+\nu^2)}\sqrt{H_t}},\frac{\sqrt{\rho(1-\nu^2)\theta_t}}{2 L\sqrt{\gamma(3+\nu^2)}\sqrt{H_t}}\big)$ with $H_t=\frac{9}{b\beta_t}+\frac{6\nu^2}{b(1-\nu)^2}+\frac{4n^2\beta^2_t}{b^3}\big(\frac{9}{\beta_t}+
\frac{9\nu^2}{(1-\nu)^2}\big)+\frac{3\nu^2}{(1-\nu)^2}$ for all $t\geq 0$, we have
\begin{align}
 & \frac{1}{T}\sum_{t=1}^T \mathbb{E}\|\nabla F(\bar{x}_t)\| \\
  & \leq \frac{1}{T}\sum_{t=1}^T\frac{1}{m}\sum_{i=1}^m\mathbb{E}[\|\nabla F(x^i_t)\| + L\|\bar{x}_t-x^i_t\|] \nonumber \\
  & \leq \frac{6\sqrt{G}}{\sqrt{T}}\sqrt{\frac{1}{T}\sum_{t=1}^T\frac{1}{m}\sum_{i=1}^m\mathbb{E}\|A^i_t\|^2}, \nonumber
\end{align}
where $G=\frac{F(\bar{x}_1)-F^*}{\rho\gamma\eta}+\big(\frac{18\beta_0}{\rho^2}+\frac{18\beta^2_0\nu^2}{\rho^2(1-\nu)^2}
 +\frac{3\nu^2}{\rho^2(1-\nu)^2}
 +\frac{9}{2\rho^2\beta_0}-\frac{9}{2\rho^2}\big)
 \frac{1}{m}\sum_{i=1}^m\frac{1}{n}\sum_{k=1}^n\|\nabla f^i_k(x^i_{0})\|^2$ is independent on
$T$, $b$ and $n$.
\end{theorem}

\begin{remark}
Based on Assumption~\ref{ass:1},
if using Barzilai-Borwein-like adaptive matrix~(\ref{eq:a3}), we have
$\sqrt{\frac{1}{T}\sum_{t=1}^T\frac{1}{m}\sum_{i=1}^m\mathbb{E}\|A^i_t\|^2}\leq L+\rho$.
Based on Assumption~\ref{ass:6}, we have
$\sqrt{\frac{1}{T}\sum_{t=1}^T\frac{1}{m}\sum_{i=1}^m\mathbb{E}\|A^i_t\|^2}\leq M+\rho$.

Let $\theta_t=\theta\geq \frac{29\gamma\eta L^2}{6\rho}$, $b=\sqrt{n}$ and $\beta_t = \frac{b}{n}$ for all $t\geq1$, we have
\begin{align}
 H_t & = \frac{9}{b\beta_t}\!+\!\frac{6\nu^2}{b(1-\nu)^2}\!+\!\frac{4n^2\beta^2_t}{b^3}\big(\frac{9}{\beta_t}\!+\!
\frac{9\nu^2}{(1-\nu)^2}\big)\!+\!\frac{3\nu^2}{(1-\nu)^2} \nonumber \\
& \leq 45 +\frac{45\nu^2}{(1-\nu)^2}.
\end{align}
Then we have $\frac{1}{\sqrt{H_t}}\geq \frac{1}{\sqrt{45 +\frac{45\nu^2}{(1-\nu)^2}}}$
and
\begin{align}
& \min\big(\frac{\rho\sqrt{1-\nu^2}}{2L\gamma\sqrt{6(1+\nu^2)}\sqrt{H_t}},\frac{\sqrt{\rho(1-\nu^2)\theta}}{2 L\sqrt{\gamma(3+\nu^2)}\sqrt{H_t}}\big) \nonumber \\
& \geq \min\big(\frac{\rho\sqrt{1-\nu^2}}{2L\gamma\sqrt{6(1+\nu^2)}},\frac{\sqrt{\rho(1-\nu^2)\theta}}{2 L\sqrt{\gamma(3+\nu^2)}}\big)\sqrt{45\!+\!\frac{45\nu^2}{(1-\nu)^2}}. \nonumber
\end{align}
Thus we can let $\eta= \min\big(\frac{\rho\sqrt{1-\nu^2}}{2L\gamma\sqrt{6(1+\nu^2)}},\frac{\sqrt{\rho(1-\nu^2)\theta}}{2 L\sqrt{\gamma(3+\nu^2)}}\big)\sqrt{45 +\frac{45\nu^2}{(1-\nu)^2}}$ and
$\gamma=\min\big(\frac{\rho(1-\nu^2)}{48\theta},\frac{3\rho(1-\nu^2)\theta}{58L^2}\big)$.
\textbf{Note that} we set $\beta_t=\frac{b}{n}$ for all $t\geq1$,
while we can set $\beta_0\in (0,1)$, which is independent on
$T$, $b$ and $n$. Let $\rho=O(1)$, $\eta=O(1)$ and $\gamma=O(1)$,
 we have $G= \frac{F(\bar{x}_1)-F^*}{\rho\gamma\eta}+\big(\frac{18\beta_0}{\rho^2}+\frac{18\beta^2_0\nu^2}{\rho^2(1-\nu)^2}
 +\frac{3\nu^2}{\rho^2(1-\nu)^2}
 +\frac{9}{2\rho^2\beta_0}-\frac{9}{2\rho^2}\big)
 \frac{1}{m}\sum_{i=1}^m\frac{1}{n}\sum_{k=1}^n\|\nabla f^i_k(x^i_{0})\|^2 =O(1)$ is independent on
$T$, $b$ and $n$.
Since $\sqrt{\frac{1}{T}\sum_{t=1}^T\frac{1}{m}\sum_{i=1}^m
  \mathbb{E}\|A^i_t\|^2}=O(1)$, set
\begin{align}
  \frac{1}{T}\sum_{t=1}^T \mathbb{E}\|\nabla F(\bar{x}_t)\|
  \leq O(\frac{1}{\sqrt{T}}) \leq \epsilon, \nonumber
\end{align}
we have $T= O(\epsilon^{-2})$. Since our AdaMDOF algorithm requires $b$ samples, we can obtain a near-optimal sample complexity of $T\cdot b=O(\sqrt{n}\epsilon^{-2})$, which matches the lower bound of smooth nonconvex finite-sum optimization~\citep{fang2018spider}.
\end{remark}

\begin{figure}[ht]
\centering
 \subfloat{\includegraphics[width=0.24\textwidth]{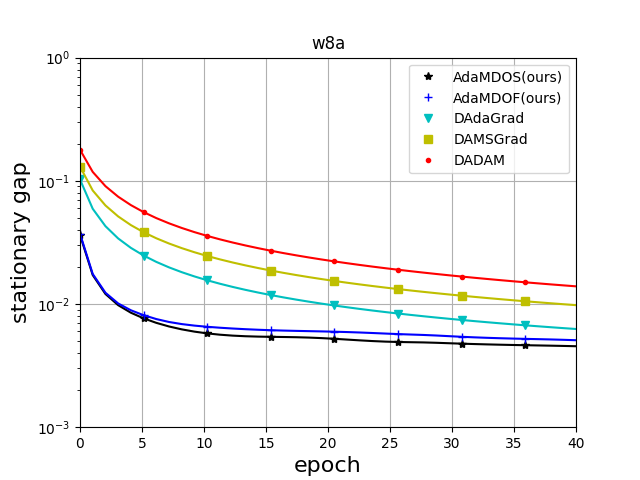}}
  \hfill
 \subfloat{\includegraphics[width=0.24\textwidth]{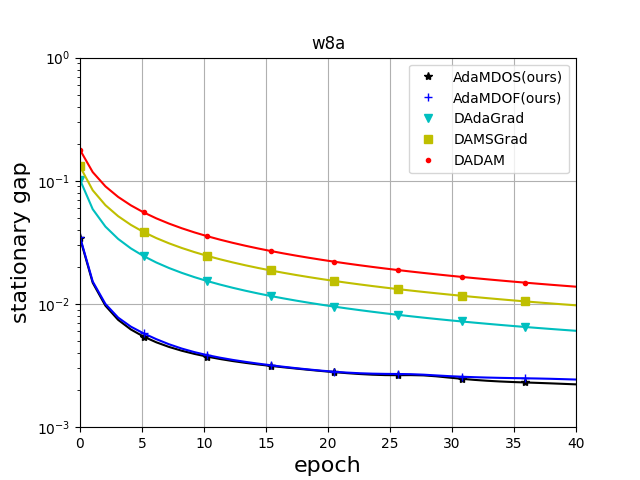}}
  \hfill
\caption{ Stationary gap vs epoch at \textbf{w8a} dataset under the ring network (Left) and the 3-regular network (Right).}
\label{fig:LL-w8a}
\end{figure}

\begin{figure}[ht]
\centering
 \subfloat{\includegraphics[width=0.24\textwidth]{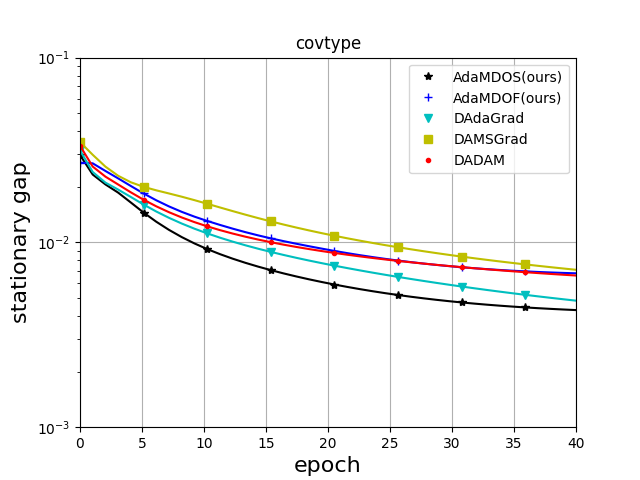}}
  \hfill
 \subfloat{\includegraphics[width=0.24\textwidth]{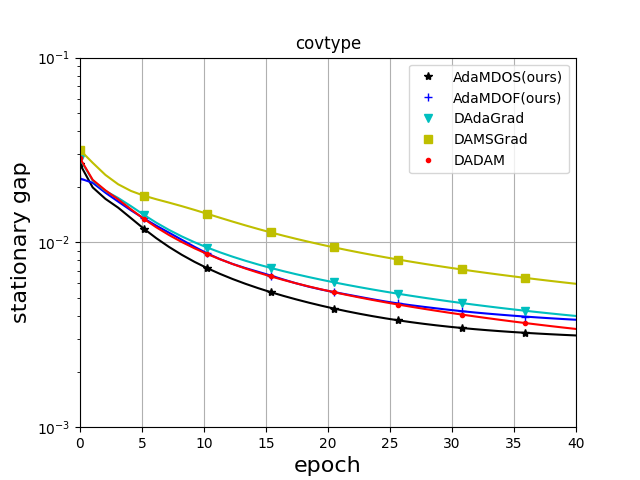}}
  \hfill
\caption{ Stationary gap vs epoch at \textbf{covertype} dataset under the ring network (Left) and the 3-regular network (Right).}
\label{fig:LL-covtype}
\end{figure}

\begin{figure*}[ht]
\centering
 \subfloat{\includegraphics[width=0.33\textwidth]{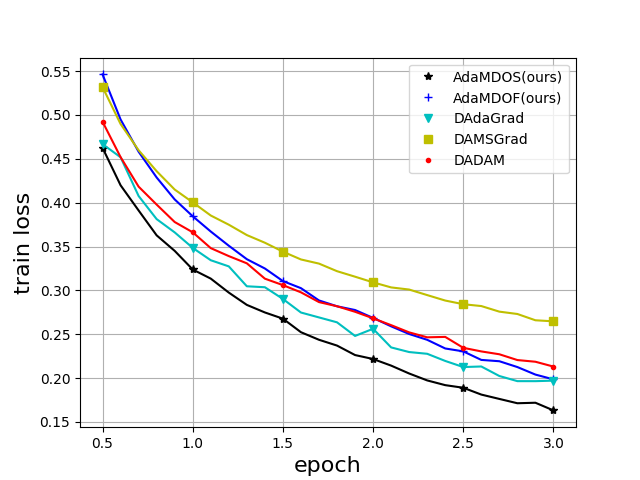}}
  \hfill
 \subfloat{\includegraphics[width=0.33\textwidth]{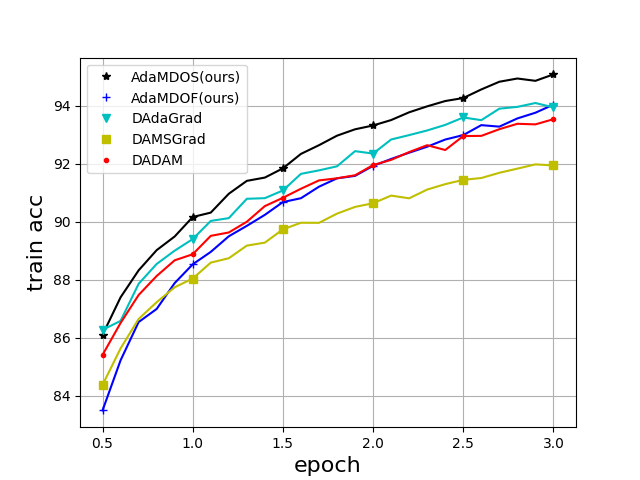}}
  \hfill
 \subfloat{\includegraphics[width=0.33\textwidth]{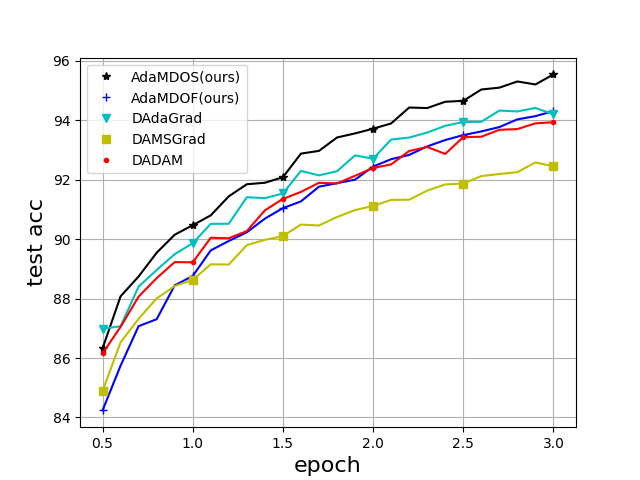}}
  \hfill
\caption{ Training CNN on MNIST dataset: training loss vs epoch (Left), training accuracy (\%) vs epoch (Middle), and test accuracy (\%) vs epoch (Right) under the \emph{ring} network.}
\label{fig:CNN-ring}
\end{figure*}

\begin{figure*}[ht]
\centering
 \subfloat{\includegraphics[width=0.33\textwidth]{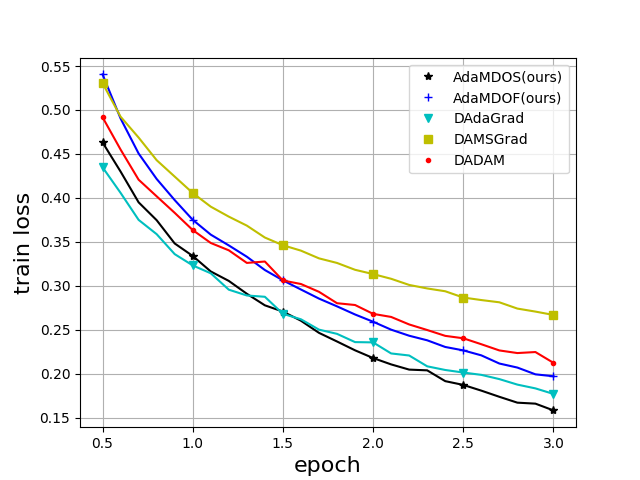}}
  \hfill
 \subfloat{\includegraphics[width=0.33\textwidth]{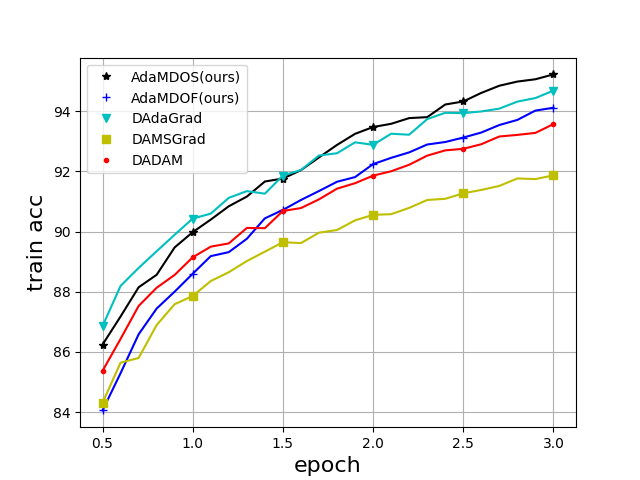}}
  \hfill
 \subfloat{\includegraphics[width=0.33\textwidth]{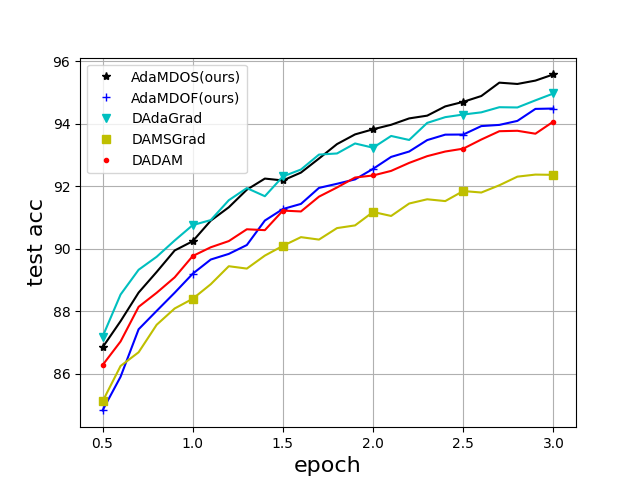}}
  \hfill
\caption{ Training CNN on MNIST dataset: training loss vs epoch (Left), training accuracy (\%) vs epoch (Middle), and test accuracy (\%) vs epoch (Right) under the \emph{3-regular} network.}
\label{fig:CNN-3}
\end{figure*}

\section{Numerical Experiments}
In this section, we apply some numerical experiments to demonstrate efficiency of
our AdaMDOS and AdaMDOF algorithms. Since the AdaRWGD~\citep{sun2022adaptive} relies on the random walk
instead of parallel framework used in other algorithms, our adaptive decentralized algorithms
only compare to the existing adaptive decentralized methods (i.e, DADAM~\citep{nazari2022dadam}, DAMSGrad~\citep{chen2023convergence}, DAdaGrad~\citep{chen2023convergence}) given in Tabel~\ref{tab:1}.
In decentralized algorithms, we consider two classical undirected networks that
connect all clients, i.e., the \emph{ring} and \emph{3-regular} expander networks~\citep{hoory2006expander},
described in the following Appendix~\ref{app:add}.

\subsection{ Training Logistic Model }
In this subsection, we consider learning a non-convex logistic model~\citep{allen2016variance} for binary
classification over a decentralized network of $m$ nodes with $n$ data samples at each node:
\begin{align}
  \min_{x\in \mathbb{R}^d} \frac{1}{m}\sum_{i=1}^m\Big(\frac{1}{n}\sum_{k=1}^n f^i(x;\xi^i_k) + \lambda\|x\|^2\Big),
\end{align}
where $\lambda>0$ and $f^i(x;\xi^i_k) = \frac{1}{1+ \exp(l_k^i\langle a_k^i, x\rangle)}$ is a nonconvex sigmoid loss function. Here $\xi^i_k=(a^i_k,l^i_k)$ denotes the $k$-th sample at $i$-th node, where $a^i_k \in \mathbb{R}^d$ denotes the features and $l^i_k\in \{-1,1\}$ is a label. In the experiment, we set the regularization parameter $\lambda=10^{-5}$, and use the same initial solution $x_0 = x_0^i=0.01\cdot\mbox{ones}(d,1)$ for all $i\in [m]$ for all algorithms. We use public w8a and covertype datasets\footnote{available at https://www.openml.org/}.
The w8a dataset includes 60,000 training examples, where we partitioned into 5
clients each containing 12,000 training examples. The covertype dataset includes 100,000 training examples, where we partitioned into 5 clients each containing 20,000 training examples.

In the experiment, we characterize performance
of the algorithms in comparison in terms of the decrease of stationary
gap versus epochs, where the \emph{stationary gap} is defined as $\|\nabla F(\bar{x}_t)\|+\frac{1}{m}\sum_{i=1}^m\|\bar{x}_t-x^i_t\|$,
where $x_t^i$ is the estimate of the stationary solution at the $i$-th node and $\bar{x}_t=\frac{1}{m}\sum_{i=1}^mx_t^i$,
and each epoch represents $n$ component gradient computations at each node.
In the experiment, for fair comparison, we use the batch size $b=10$ in all algorithms, and
set $\beta_1 = \beta_2 = 0.9$ in the DADAM~\citep{nazari2022dadam} and
DAMSGrad~\citep{chen2023convergence}, and set $\beta_1= 0.9$ in the DAdaGrad~\citep{chen2023convergence}, and
set $\varrho=\beta_t=\eta_t=0.9$ for all $t\geq 1$ in our algorithms.

Figures~\ref{fig:LL-w8a} and~\ref{fig:LL-covtype} show that our AdaMDOS method outperforms all comparisons,
while due to requiring large batch-size, our AdaMDOF is comparable with the existing adaptive methods.
\subsection{ Training Convolutional Neural Network }
In this subsection, we consider training a Convolutional Neural Network (CNN) for MNIST
classification over a decentralized network. Here we use the same CNN architecture as in~\cite{mcmahan2017communication}.
This CNN includes two $5\times 5$ convolution layers
(the first with 32 channels, the second with 64, each followed with $2\times 2$
max pooling), a fully connected layer with 512 units and
ReLu activation, and a final softmax output layer (1,663,370
total parameters). The MNIST dataset~\cite{lecun2010mnist} consists of 10 classes of $28\times 28$ grayscale images,
which includes 60,000 training examples and 10,000 testing examples, which we partitioned into 5
clients each containing 12000 training and 2000 testing examples.

In the experiment, we characterize performance
of the algorithms in comparison in terms of the decrease of training loss versus epochs,
where the loss denotes the objective function value in training CNN.
Meanwhile, we also use the training accuracy and test accuracy, where the accuracy denotes the classification accuracy.
For fair comparison, we use the batch size $b=10$ in all algorithms, and
set $\beta_1 = \beta_2 = 0.9$ in the DADAM~\citep{nazari2022dadam} and
DAMSGrad~\citep{chen2023convergence}, and set $\beta_1= 0.9$ in the DAdaGrad~\citep{chen2023convergence}, and
set $\varrho=\beta_t=\eta_t=0.9$ for all $t\geq 1$ in our algorithms.
Figures~\ref{fig:CNN-ring} and~\ref{fig:CNN-3} also show that our AdaMDOS method outperforms all comparisons, while due to requiring large batch size, our AdaMDOF is comparable with the existing adaptive methods.

\subsection{ Training Residual Network }
In this subsection, we consider training a residual neural network for Tiny-ImageNet classification over a decentralized network. Here we use the ResNet-18 as in~\citep{he2016deep}, which includes a $3\times3$ convolution layer followed with batch-norm and ReLU activation, eight residual blocks (start with 64 channels, channel number doubled at third, fifth and seventh block, end with 512 channels), a $4\times4$ max pooling, a fully connected layer with 512 units and ReLU activation, and a final softmax output layer. Each residual block contains a shortcut and two $3\times3$ convolution layers, the first followed with batch-norm and ReLU activation,
the second followed with batch-norm.

The Tiny-ImageNet dataset~\cite{Le2015TinyIV} consists of 200 classes of 64 x 64 RGB images, which includes 100,000 training examples and 10,000 testing examples, respectively. Here we partitioned into 5 clients, where each client contains 20,000 training and 2000 testing examples, respectively.

\begin{figure}[ht]
\centering
\subfloat{\includegraphics[width=0.24\textwidth]{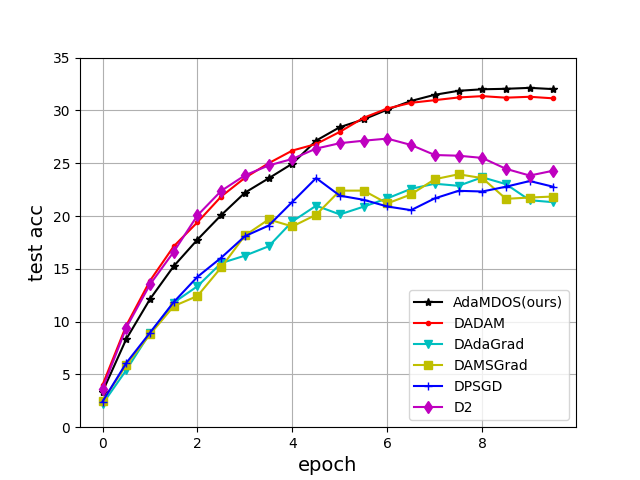}}
  \hfill
 \subfloat{\includegraphics[width=0.24\textwidth]{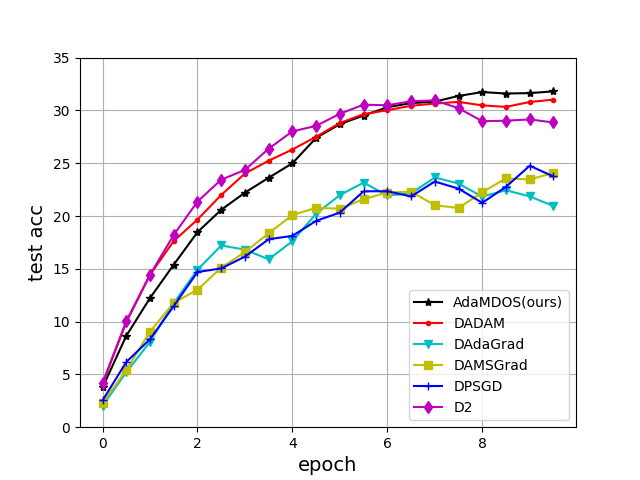}}
  \hfill
\caption{ Training ResNet-18 on Tiny-ImageNet dataset: test accuracy (\%) vs epoch under the \emph{ring} network (Left) and the \emph{3-regular} network (Right).}
\label{fig:ResNet}
\end{figure}

For fair comparison, we use the batch size $b=10$ in all algorithms, and set $\beta_1=\beta_2=0.9$ in the DADAM and DAMSGrad, and set $\beta_1=0.9$ in the DAdaGrad, and set $\varrho=\beta_t=\eta_t=0.9$ for all $t\geq 1$ in our algorithms. In this experiment, we add two basic non-adaptive decentralized algorithms, i.e., D-PSGD~\cite{lian2017can} and $D^2$~\cite{tang2018d}, as the comparisons.
From Figure~\ref{fig:ResNet}, we find that although DADAM and $D^2$ methods outperform our AdaMDOS method at the beginning of the iteration, while our AdaMDOS then outperforms all comparisons.

\section{Conclusion}
In the paper, we studied the distributed nonconvex stochastic and finite-sum optimization problems over a network.
Moreover, we proposed a faster adaptive momentum-based decentralized optimization algorithm (i.e., AdaMDOS) to solve
the stochastic problems, which reaches a near-optimal sample complexity of $\tilde{O}(\epsilon^{-3})$
for nonconvex stochastic optimization. Meanwhile, we proposed a faster adaptive momentum-based decentralized optimization algorithm (i.e., AdaMDOF) to solve
the finite-sum problems, which obtains a near-optimal sample complexity of $O(\sqrt{n}\epsilon^{-2})$
for nonconvex finite-sum optimization. In particular, our methods use a unified adaptive matrix including various types of adaptive learning rate.

% Acknowledgements should only appear in the accepted version.
\section*{Acknowledgements}
We thank the anonymous reviewers for their helpful comments. This paper was partially supported by NSFC under Grant No.
62376125. It was also partially supported by the Fundamental Research Funds for the Central Universities NO.NJ2023032.

\section*{Impact Statement}
This paper presents work whose goal is to advance the field of Machine Learning. There are many potential societal consequences of our work, none which we feel must be specifically highlighted here.

% In the unusual situation where you want a paper to appear in the
% references without citing it in the main text, use \nocite
\nocite{langley00}

\bibliography{AdaMDO}
\bibliographystyle{icml2024}

%%%%%%%%%%%%%%%%%%%%%%%%%%%%%%%%%%%%%%%%%%%%%%%%%%%%%%%%%%%%%%%%%%%%%%%%%%%%%%%
%%%%%%%%%%%%%%%%%%%%%%%%%%%%%%%%%%%%%%%%%%%%%%%%%%%%%%%%%%%%%%%%%%%%%%%%%%%%%%%
% APPENDIX
%%%%%%%%%%%%%%%%%%%%%%%%%%%%%%%%%%%%%%%%%%%%%%%%%%%%%%%%%%%%%%%%%%%%%%%%%%%%%%%
%%%%%%%%%%%%%%%%%%%%%%%%%%%%%%%%%%%%%%%%%%%%%%%%%%%%%%%%%%%%%%%%%%%%%%%%%%%%%%%
\newpage
\appendix
\onecolumn

\section{Appendix}
In this section, we provide the detailed convergence analysis of our algorithms.
We first give some notations and review some useful lemmas.

For the \textbf{stochastic} problem~(\ref{eq:1}). Let $\mathbb{E}_t = \mathbb{E}_{\xi_t,\xi_{t-1},\cdots,\xi_1}$ with $\xi_t \in \{\xi_t^i\}_{i=1}^m$.
 Let $\bar{x}_t=\frac{1}{m}\sum_{i=1}^m x^i_t$,
$\bar{w}_t=\frac{1}{m}\sum_{i=1}^m w^i_t$, $\nabla f^i(x^i_t)=\mathbb{E}[\nabla f^i(x^i_t;\xi^i_t)]$ for all $i\in [m], \ t\geq 1$ and
\begin{align}
\overline{\nabla f(x_t)}= \frac{1}{m}\sum_{i=1}^m \nabla f^i(x^i_t), \ \nabla F(\bar{x}_t) = \frac{1}{m}\sum_{i=1}^m \nabla f^i(\bar{x}_t).
\end{align}

For the \textbf{finite-sum} problem~(\ref{eq:2}). Let $\mathbb{E}_t = \mathbb{E}_{\mathcal{I}_t,\mathcal{I}_{t-1},\cdots,\mathcal{I}_1}$ with $\mathcal{I}_t \in \{\mathcal{I}_t^i\}_{i=1}^m$.Let $\bar{x}_t=\frac{1}{m}\sum_{i=1}^m x^i_t$,
$\bar{w}_t=\frac{1}{m}\sum_{i=1}^m w^i_t$, $\nabla f^i(x^i_t)=\frac{1}{n}\sum_{k=1}^n\nabla f^i_k(x^i_t)$ for all $i\in [m]$ and
\begin{align}
\overline{\nabla f(x_t)}=\frac{1}{m}\sum_{i=1}^m (\frac{1}{n}\sum_{k=1}^n\nabla f^i_k(x^i_t))=
\frac{1}{m}\sum_{i=1}^m \nabla f^i(x^i_t), \
\nabla F(\bar{x}_t) = \frac{1}{m}\sum_{i=1}^m \nabla f^i(\bar{x}_t).
\end{align}

\begin{lemma} \label{lem:A1}
Given $m$ vectors $\{u^i\}_{i=1}^m$, the following inequalities satisfy: $||u^i + u^j||^2 \leq (1+c)||u^i||^2 + (1+\frac{1}{c})||u^j||^2$ for any $c > 0$, and $||\frac{1}{m}\sum_{i=1}^m u^i||^2 \le \frac{1}{m}\sum_{i=1}^{m} ||u^i||^2$.
\end{lemma}

\begin{lemma} \label{lem:A2}
Given a finite sequence $\{u^i\}_{i=1}^m$, and $\bar{u} = \frac{1}{m}\sum_{i=1}^m u^i$,
the following inequality satisfies $\sum_{i=1}^m \|u^i - \bar{u}\|^2 \leq \sum_{i=1}^m \|u^i\|^2$.
\end{lemma}

\begin{lemma} \label{lem:B1}
 The sequences $\{u^i_{t\geq 1},w^i_{t\geq 1}\}_{i=1}^m$ be generated from our Algorithm~\ref{alg:1} or~\ref{alg:2}, we have for all $t\geq 1$,
 \begin{align}
 \frac{1}{m}\sum_{i=1}^mu^i_t = \bar{u}_t= \bar{w}_t=\frac{1}{m}\sum_{i=1}^mw^i_t.
 \end{align}
\end{lemma}

\begin{proof}
We proceed by induction.
From our Algorithm~\ref{alg:1}, since $w^i_1 = \sum_{j\in \mathcal{N}_i} W_{i,j}u^j_1$, we have
\begin{align}
 \bar{w}_1 = \frac{1}{m}\sum_{i=1}^m w^i_1 = \frac{1}{m}\sum_{i=1}^m\sum_{j\in \mathcal{N}_i} W_{i,j}u^j_1
 =\frac{1}{m}\sum_{j=1}^m u^j_1\sum_{i=1}^m W_{i,j} = \frac{1}{m}\sum_{j=1}^m u^j_1 = \bar{u}_1,
\end{align}
where the second last equality is due to $\sum_{i=1}^m W_{i,j}=1$ from Assumption \ref{ass:4}.

From the line 10 of Algorithm~\ref{alg:1}, we have for all $t\geq 1$
\begin{align}
  \bar{w}_{t+1} & = \frac{1}{m}\sum_{i=1}^m w^i_{t+1} = \frac{1}{m}\sum_{i=1}^m\sum_{j\in \mathcal{N}_i} W_{i,j}\big(w^j_t
  +u^j_{t+1}-u^j_t\big) \nonumber \\
  & = \frac{1}{m}\sum_{j=1}^m\big(w^j_t
  +u^j_{t+1}-u^j_t\big)\sum_{i=1}^m W_{i,j}= \frac{1}{m}\sum_{j=1}^m\big(w^j_t
  +u^j_{t+1}-u^j_t\big) = \bar{w}_t
  +\bar{u}_{t+1}-\bar{u}_t = \bar{u}_{t+1},
\end{align}
where the second last equality is due to $\mathcal{N}_i=\{j\in V \ | \ (i,j)\in E, j=i\}$ and $\sum_{i=1}^m W_{i,j}=1$, and the last equality holds by
the inductive hypothesis, i.e., $\bar{w}_t=\bar{u}_t$.

\end{proof}

\begin{lemma} \label{lem:B2}
 Suppose the sequence $\big\{x_{t\geq1}^i, \tilde{x}^i_{t\geq1}\big\}_{i=1}^m$ be generated from Algorithm \ref{alg:1} or  \ref{alg:2}.
 Let $ 0< \gamma \leq \frac{\rho}{4L\eta_t}$ for all $t\geq 0$,
 then we have
 \begin{align}
 F(\bar{x}_{t+1}) \leq F(\bar{x}_t)+\frac{2\gamma\eta_t}{\rho}\|\nabla F(\bar{x}_t)-\bar{u}_t\|^2+\frac{\gamma\eta_t}{2\rho}\frac{1}{m}\sum_{i=1}^m\|w^i_t -\bar{w}_t\|^2-\frac{\rho\gamma\eta_t}{4} \frac{1}{m}\sum_{i=1}^m\|g^i_t\|^2,
 \end{align}
 where $g^i_t = (A^i_t)^{-1}w^i_t$ for any $i\in [m]$.
\end{lemma}

\begin{proof}
Let $g^i_t = (A^i_t)^{-1}w^i_t$, we have $\tilde{x}^i_{t+1} = \sum_{j\in \mathcal{N}_i} W_{i,j}x^j_t - \gamma (A^i_t)^{-1}w^i_t$. Then we have
\begin{align}
\bar{\tilde{x}}_{t+1} & = \frac{1}{m}\sum_{i=1}^m\tilde{x}^i_{t+1} =\frac{1}{m}\sum_{i=1}^m\sum_{j\in \mathcal{N}_i} W_{i,j}x^j_t - \gamma \frac{1}{m}\sum_{i=1}^m(A^i_t)^{-1}w^i_t \nonumber \\
& = \frac{1}{m}\sum_{j\in \mathcal{N}_i}x^j_t\sum_{i=1}^m W_{i,j} - \gamma \frac{1}{m}\sum_{i=1}^mg^i_t =
\bar{x}_t - \gamma \bar{g}_t,
\end{align}
where the last equality is due to $\sum_{i=1}^m W_{i,j}=1$.
Since $A^i_t\succeq \rho I_d$ for all $i\in [m]$, we have
\begin{align}
 \rho \|g^i_t\|^2 & \leq \langle A^i_t g^i_t, g^i_t\rangle = \langle w^i_t, g^i_t\rangle \mathop{=}^{(i)} \langle w^i_t -\bar{w}_t , g^i_t\rangle + \langle\bar{u}_t , g^i_t\rangle \nonumber \\
 & \leq \frac{1}{2\rho}\|w^i_t -\bar{w}_t\|^2 + \frac{\rho}{2}\|g^i_t\|^2 + \langle\bar{u}_t , g^i_t\rangle,
\end{align}
where the equality (i) is due to $\bar{u}_t=\bar{w}_t$.
So we can obtain
\begin{align}
 0 \leq  \frac{1}{2\rho}\|w^i_t -\bar{w}_t\|^2 - \frac{\rho}{2}\|g^i_t\|^2 + \langle\bar{u}_t , g^i_t\rangle,
\end{align}
Then we have for any $i\in[m]$
\begin{align}
 0 \leq  \frac{\gamma\eta_t}{2\rho}\|w^i_t -\bar{w}_t\|^2 - \frac{\rho\gamma\eta_t}{2}\|g^i_t\|^2 + \gamma\eta_t\langle\bar{u}_t , g^i_t\rangle.
\end{align}
Then taking an average over $i$ from 1 to $m$ yields that
\begin{align} \label{eq:B1}
 0 & \leq  \frac{\gamma\eta_t}{2\rho}\frac{1}{m}\sum_{i=1}^m\|w^i_t -\bar{w}_t\|^2 - \frac{\rho\gamma\eta_t}{2}\frac{1}{m}\sum_{i=1}^m\|g^i_t\|^2 + \gamma\eta_t\frac{1}{m}\sum_{i=1}^m\langle\bar{u}_t , g^i_t\rangle \nonumber \\
 & = \frac{\gamma\eta_t}{2\rho}\frac{1}{m}\sum_{i=1}^m\|w^i_t -\bar{w}_t\|^2 - \frac{\rho\gamma\eta_t}{2}\frac{1}{m}\sum_{i=1}^m\|g^i_t\|^2 + \gamma\eta_t\langle\bar{u}_t , \bar{g}_t\rangle.
\end{align}

According to Assumption \ref{ass:1}, i.e., the function $F(x)$ is $L$-smooth,
we have
 \begin{align} \label{eq:B2}
  F(\bar{x}_{t+1}) & \leq F(\bar{x}_t) + \langle\nabla F(\bar{x}_t), \bar{x}_{t+1}-\bar{x}_t\rangle + \frac{L}{2}\|\bar{x}_{t+1}-\bar{x}_t\|^2 \nonumber\\
  & = F(\bar{x}_t)+ \eta_t\langle \nabla F(\bar{x}_t),\bar{\tilde{x}}_{t+1}-\bar{x}_t\rangle + \frac{L\eta_t^2}{2}\|\bar{\tilde{x}}_{t+1}-\bar{x}_t\|^2 \nonumber \\
   & = F(\bar{x}_t)+ \eta_t\langle \nabla F(\bar{x}_t)-\bar{u}_t+\bar{u}_t,-\gamma\bar{g}_t\rangle + \frac{L\gamma^2\eta_t^2}{2}\|\bar{g}_t\|^2 \nonumber \\
  & \leq F(\bar{x}_t)+\frac{2\gamma\eta_t}{\rho}\|\nabla F(\bar{x}_t)-\bar{u}_t\|^2+\frac{\rho\gamma\eta_t}{8}\|\bar{g}_t\|^2 -\gamma \eta_t\langle \bar{u}_t,\bar{g}_t\rangle + \frac{L\gamma^2\eta_t^2}{2}\|\bar{g}_t\|^2 \nonumber \\
  & \leq  F(\bar{x}_t)+\frac{2\gamma\eta_t}{\rho}\|\nabla F(\bar{x}_t)-\bar{u}_t\|^2+\frac{\rho\gamma\eta_t}{8}\frac{1}{m}\sum_{i=1}^m\|g^i_t\|^2 -\gamma \eta_t\langle \bar{u}_t,\bar{g}_t\rangle + \frac{L\gamma^2\eta_t^2}{2}\frac{1}{m}\sum_{i=1}^m\|g^i_t\|^2,
 \end{align}
where the second equality is due to $\bar{x}_{t+1}=\bar{x}_t + \eta_t(\bar{\tilde{x}}_{t+1}-\bar{x}_t)$.

By summing the above inequalities (\ref{eq:B1}) with (\ref{eq:B2}), we can obtain
 \begin{align}
   F(\bar{x}_{t+1}) & \leq  F(\bar{x}_t)+\frac{2\gamma\eta_t}{\rho}\|\nabla F(\bar{x}_t)-\bar{u}_t\|^2+\frac{\gamma\eta_t}{2\rho}\frac{1}{m}\sum_{i=1}^m\|w^i_t -\bar{w}_t\|^2-(\frac{3\rho\gamma\eta_t}{8} - \frac{L\gamma^2\eta_t^2}{2})\frac{1}{m}\sum_{i=1}^m\|g^i_t\|^2 \nonumber \\
   & \leq F(\bar{x}_t)+\frac{2\gamma\eta_t}{\rho}\|\nabla F(\bar{x}_t)-\bar{u}_t\|^2+\frac{\gamma\eta_t}{2\rho}\frac{1}{m}\sum_{i=1}^m\|w^i_t -\bar{w}_t\|^2-\frac{\rho\gamma\eta_t}{4} \frac{1}{m}\sum_{i=1}^m\|g^i_t\|^2,
 \end{align}
where the last inequality is due to $\gamma \leq \frac{\rho}{4L\eta_t}$ for all $t\geq 1$.

\end{proof}

\subsection{ Convergence Analysis of AdaMDOS Algorithm}
In this subsection, we provide the convergence analysis of our AdaMDOS Algorithm for \textbf{stochastic} optimization.

\begin{lemma} \label{lem:D1}
Under the above assumptions, and assume the stochastic gradient estimators $\big\{u^i_{t\geq 1}\big\}_{i=1}^m$ be generated from Algorithm \ref{alg:1}, we have
 \begin{align}
 & \mathbb{E}\|u^i_{t+1} - \nabla f^i(x_{t+1})\|^2
  \leq (1-\beta_{t+1})\mathbb{E} \|u^i_t - \nabla f^i(x_t)\|^2 + 2\beta_{t+1}^2\sigma^2+ 2L^2\eta_t^2\mathbb{E}\|\tilde{x}^i_{t+1} - x^i_t\|^2, \ \forall i\in[m] \\
 & \mathbb{E}\|\bar{u}_{t+1} - \overline{\nabla f(x_{t+1})}\|^2
  \leq (1-\beta_{t+1})\mathbb{E} \|\bar{u}_t -\overline{\nabla f(x_t)}\|^2 + \frac{2\beta_{t+1}^2\sigma^2}{m} + \frac{2L^2\eta_t^2}{m^2}\sum_{i=1}^m\mathbb{E}\|\tilde{x}^i_{t+1} - x^i_t\|^2,
 \end{align}
where $\overline{\nabla f(x_t)} = \frac{1}{m}\sum_{i=1}^m\nabla f^i(x^i_t)$.
\end{lemma}

\begin{proof}
Since $u^i_{t+1} = \nabla f^i(x^i_{t+1};\xi^i_{t+1}) + (1-\beta_{t+1})(u^i_t-\nabla f^i(x^i_t;\xi^i_{t+1}))$ for any $i\in [m]$, we have
\begin{align}
\bar{u}_{t+1} & = \frac{1}{m}\sum_{i=1}^m\big( \nabla f^i(x^i_{t+1};\xi^i_{t+1}) + (1-\beta_{t+1})(u^i_t-\nabla f^i(x^i_t;\xi^i_{t+1}))\big) \nonumber \\
& =  \overline{\nabla f(x_{t+1};\xi_{t+1})} + (1-\beta_{t+1})(\bar{u}_t - \overline{\nabla f(x_t;\xi_{t+1})}).
\end{align}

Then we have
\begin{align}
 &\mathbb{E}\|\bar{u}_{t+1} - \overline{\nabla f(x_{t+1})}\|^2  \nonumber \\
 & = \mathbb{E}\|\frac{1}{m}\sum_{i=1}^m\big(u^i_{t+1} - \nabla f^i(x^i_{t+1})\big)\|^2 \nonumber \\
 & = \mathbb{E}\|\frac{1}{m}\sum_{i=1}^m\big(\nabla f^i(x^i_{t+1};\xi^i_{t+1}) + (1-\beta_{t+1})\big(u^i_t
 - \nabla f^i(x^i_t;\xi^i_{t+1})\big) - \nabla f^i(x^i_{t+1})\big)\|^2 \nonumber \\
 & = \mathbb{E}\big\|\frac{1}{m}\sum_{i=1}^m\Big( \nabla f^i(x^i_{t+1};\xi^i_{t+1})-\nabla f^i(x^i_{t+1}) - (1-\beta_{t+1})\big( \nabla f^i(x^i_t;\xi^i_{t+1}) - \nabla f^i(x^i_t) \big) \Big) \nonumber \\
 & \quad + (1-\beta_{t+1})\frac{1}{m}\sum_{i=1}^m\big( u^i_t - \nabla f^i(x^i_t) \big) \big\|^2 \nonumber \\
 & = \frac{1}{m^2}\sum_{i=1}^m\mathbb{E}\big\|\nabla f^i(x^i_{t+1};\xi^i_{t+1})-\nabla f^i(x^i_{t+1}) - (1-\beta_{t+1})\big( \nabla f^i(x^i_t;\xi^i_{t+1}) - \nabla f^i(x^i_t) \big) \big\|^2 \nonumber \\
 & \quad + (1-\beta_{t+1})^2\mathbb{E}\| \bar{u}_t - \overline{\nabla f(x_t)} \big\|^2 \nonumber \\
 & \leq \frac{2(1-\beta_{t+1})^2}{m^2}\sum_{i=1}^m\mathbb{E}\|\nabla f^i(x^i_{t+1};\xi^i_{t+1})-\nabla f^i(x^i_{t+1}) - \nabla f^i(x^i_t;\xi^i_{t+1}) + \nabla f^i(x^i_t) \big) \|^2 \nonumber \\
 & \quad + \frac{2\beta^2_{t+1}}{m^2}\sum_{i=1}^m\mathbb{E}\|\nabla f^i(x^i_{t+1};\xi^i_{t+1})-\nabla f^i(x^i_{t+1}) \|^2 + (1-\beta_{t+1})^2\mathbb{E}\|\bar{u}_t - \overline{\nabla f(x_t)} \|^2 \nonumber \\
 & \leq \frac{2(1-\beta_{t+1})^2}{m^2}\sum_{i=1}^m\mathbb{E}\|\nabla f^i(x^i_{t+1};\xi^i_{t+1})- \nabla f^i(x^i_t;\xi^i_{t+1}) \|^2 + \frac{2\beta^2_{t+1}\sigma^2}{m}
  + (1-\beta_{t+1})^2\mathbb{E}\|\bar{u}_t - \overline{\nabla f(x_t)}\|^2 \nonumber \\
 & \leq (1-\beta_{t+1})^2 \mathbb{E}\|\bar{u}_t - \overline{\nabla f(x_t)}\|^2 + \frac{2\beta^2_{t+1}\sigma^2}{m} + \frac{2(1-\beta_{t+1})^2L^2}{m^2}\sum_{i=1}^m\mathbb{E}
  \|x^i_{t+1}-x^i_t\|^2  \nonumber \\
 & \leq (1-\beta_{t+1}) \mathbb{E}\|\bar{u}_t - \overline{\nabla f(x_t)} \|^2 + \frac{2\beta^2_{t+1}\sigma^2}{m} + \frac{2L^2\eta^2_t}{m^2}\sum_{i=1}^m\mathbb{E}\|\tilde{x}^i_{t+1}-x^i_t\|^2 \nonumber,
\end{align}
where the forth equality holds by the following fact: for any $i\in [m]$,
\begin{align}
 \mathbb{E}_{\xi^i_{t+1}} \big[\nabla f^i(x^i_{t+1};\xi^i_{t+1})\big]=\nabla f^i(x^i_{t+1}), \
 \mathbb{E}_{\xi^i_{t+1}} \big[\nabla f^i(x^i_t;\xi^i_{t+1}))\big]= \nabla f^i(x^i_t), \nonumber
\end{align}
and for any $i \neq j\in [m]$, $\xi^i_{t+1}$ and $\xi^j_{t+1}$ are independent;
the second inequality holds by the inequality $\mathbb{E}\|\zeta-\mathbb{E}[\zeta]\|^2 \leq \mathbb{E}\|\zeta\|^2$ and Assumption \ref{ass:4};
the second last inequality is due to Assumption \ref{ass:2}; the last inequality holds by $0<\beta_{t+1} \leq 1$ and $x^i_{t+1}=x^i_t+\eta_t(\tilde{x}^i_{t+1}-x^i_t)$.

Similarly, we have
\begin{align}
 \mathbb{E}\|u^i_{t+1} - \nabla f^i(x_{t+1})\|^2
  \leq (1-\beta_{t+1})\mathbb{E} \|u^i_t - \nabla f^i(x_t)\|^2 + 2\beta_{t+1}^2\sigma^2+ 2L^2\eta_t^2\mathbb{E}\|\tilde{x}^i_{t+1} - x^i_t\|^2, \ \forall i\in[m].
 \end{align}

\end{proof}

\begin{lemma} \label{lem:D2}
Given the sequence $\big\{x^i_{t\geq1},w^i_{t\geq1}\big\}_{i=1}^m$ be generated from Algorithm \ref{alg:1}. We have
\begin{align}
\sum_{i=1}^m\|x^i_{t+1} - \bar{x}_{t+1}\|^2 & \leq (1-\frac{(1-\nu^2)\eta_t}{2})\sum_{i=1}^m\|x^i_t-\bar{x}_t\|^2
+ \frac{2\eta_t\gamma^2}{1-\nu^2}\sum_{i=1}^m\| g^i_t-\bar{g}_t \|^2,
\nonumber \\
\sum_{i=1}^m\|\tilde{x}^i_{t+1}-x^i_t\|^2
 & \leq (3+\nu^2)\sum_{i=1}^m\|x^i_t -\bar{x}_t\|^2 + \frac{2(1+\nu^2)}{1-\nu^2}\gamma^2
 \sum_{i=1}^m\|g^i_t\|^2 \nonumber \\
\sum_{i=1}^m\|w^i_{t+1}-\bar{w}_{t+1}\|^2
    & \leq \nu\sum_{i=1}^m\|w^i_t-\bar{w}_t\|^2 + \frac{\nu^2}{1-\nu}\big(4\beta_{t+1}^2\sum_{i=1}^m\|u^i_t-\nabla f^i(x^i_t)\|^2 \nonumber \\
    & \quad + 4\beta^2_{t+1}m\sigma^2 + 8\eta^2_tL^2\sum_{i=1}^m\|\tilde{x}^i_{t+1}-x^i_t\|^2\big).  \nonumber
\end{align}

\end{lemma}

\begin{proof}
For notational simplicity, let $x_t=[(x^1_t)^T,\cdots,(x^m_t)^T]^T\in \mathbb{R}^{md}$, $\tilde{x}_t=[(\tilde{x}^1_t)^T,\cdots,(\tilde{x}^m_t)^T]^T\in \mathbb{R}^{md}$ and $g_t=[(g^1_t)^T,\cdots,(g^m_t)^T]^T\in \mathbb{R}^{md}$ for all $t\geq1$.
By using Assumption \ref{ass:4}, since $W\textbf{1}=\textbf{1}$ and $\tilde{W}=W\otimes I_d$, we have $\tilde{W}(\textbf{1}\otimes \bar{x}_t)=\textbf{1}\otimes\bar{x}_t$.
Meanwhile, we have $\textbf{1}^T(x_t - \textbf{1}\otimes\bar{x})=0$ and $\tilde{W}\textbf{1}=\textbf{1}$.
Thus, we have
\begin{align} \label{eq:D1}
  \|\tilde{W}x_t - \textbf{1}\otimes\bar{x}_t\|^2 = \|\tilde{W}(x_t-\textbf{1}\otimes\bar{x}_t)\|^2 \leq \nu^2\|x_t-\textbf{1}\otimes\bar{x}_t\|^2,
\end{align}
where the above inequality holds by $x_t - \textbf{1}\otimes\bar{x}_t$ is orthogonal to $\textbf{1}$ that is the eigenvector corresponding to the largest eigenvalue of $\tilde{W}$, and $\nu$ denotes the second largest eigenvalue of $\tilde{W}$.

Since $\tilde{x}^i_{t+1} = \sum_{j\in \mathcal{N}_i} W_{i,j}x^j_t - \gamma (A^i_t)^{-1}w^i_t= \sum_{j\in \mathcal{N}_i} W_{i,j}x^j_t - \gamma g^i_t$ for all $i\in [m]$, we have $\tilde{x}_{t+1} = \tilde{W}x_t - \gamma g_t$ and $\bar{\tilde{x}}_{t+1} = \bar{x}_t - \gamma \bar{g}_t$.
Since $x_{t+1}=x_t + \eta_t(\tilde{x}_{t+1}-x_t)$ and $\bar{x}_{t+1}=\bar{x}_t + \eta_t(\bar{\tilde{x}}_{t+1}-\bar{x}_t)$, we have
\begin{align}
  \sum_{i=1}^m\|x^i_{t+1} - \bar{x}_{t+1}\|^2 & = \big\| x_{t+1} - \textbf{1}\otimes\bar{x}_{t+1}\big\|^2 \\
  & = \big\|x_t + \eta_t(\tilde{x}_{t+1}-x_t) - \textbf{1}\otimes\big((\bar{x}_t + \eta_t(\bar{\tilde{x}}_{t+1}-\bar{x}_t)\big) \big\|^2 \nonumber \\
  & \leq (1+\alpha_1)(1-\eta_t)^2\|x_t-\textbf{1}\otimes\bar{x}_t\|^2+(1+\frac{1}{\alpha_1})\eta^2_t
  \|\tilde{x}_{t+1}-\textbf{1}\otimes\bar{\tilde{x}}_{t+1} \|^2 \nonumber \\
  & \mathop{=}^{(i)} (1-\eta_t)\|x_t-\textbf{1}\otimes\bar{x}_t\|^2+\eta_t
  \|\tilde{x}_{t+1}-\textbf{1}\otimes\bar{\tilde{x}}_{t+1} \|^2 \nonumber \\
  & = (1-\eta_t)\|x_t-\textbf{1}\otimes\bar{x}_t\|^2+\eta_t
  \|\tilde{W}x_t - \gamma g_t-\textbf{1}\otimes\big( \bar{x}_t - \gamma \bar{g}_t\big) \|^2 \nonumber \\
  & \leq (1-\eta_t)\|x_t-\textbf{1}\otimes\bar{x}_t\|^2+(1+\alpha_2)\eta_t
  \|\tilde{W}x_t - \textbf{1}\otimes\bar{x}_t\|^2 + (1+\frac{1}{\alpha_2})\eta_t\gamma^2\| g_t-\textbf{1}\otimes\bar{g}_t \|^2 \nonumber \\
  & \mathop{\leq}^{(ii)} (1-\eta_t)\|x_t-\textbf{1}\otimes\bar{x}_t\|^2+\frac{(1+\nu^2)\eta_t}{2}
  \|x_t - \textbf{1}\otimes\bar{x}_t\|^2 + \frac{\eta_t\gamma^2(1+\nu^2)}{1-\nu^2}\| g_t-\textbf{1}\otimes\bar{g}_t \|^2 \nonumber \\
  & \mathop{\leq}^{(iii)} (1-\frac{(1-\nu^2)\eta_t}{2})\|x_t-\textbf{1}\otimes\bar{x}_t\|^2 + \frac{2\eta_t\gamma^2}{1-\nu^2}\| g_t-\textbf{1}\otimes\bar{g}_t \|^2 \nonumber \\
  & = (1-\frac{(1-\nu^2)\eta_t}{2})\sum_{i=1}^m\|x^i_t-\bar{x}_t\|^2 + \frac{2\eta_t\gamma^2}{1-\nu^2}\sum_{i=1}^m\| g^i_t-\bar{g}_t \|^2,
\end{align}
where the above equality $(i)$ is due to $\alpha_1=\frac{\eta_t}{1-\eta_t}$, and the second inequality $(ii)$ holds by $\alpha_2=\frac{1-\nu^2}{2\nu^2}$ and $\|\tilde{W}x_t - \textbf{1}\otimes\bar{x}_t\|^2 \leq
\nu^2\|x_t - \textbf{1}\otimes\bar{x}_t\|^2$, and
 the above inequality $(ii)$ is due to $0<\nu<1$.
Meanwhile, we have
\begin{align}
 \sum_{i=1}^m\|\tilde{x}^i_{t+1}-\bar{x}_t\|^2 & =\|\tilde{x}_{t+1}-\textbf{1}\otimes\bar{x}_t\|^2
 \nonumber \\
 & = \|\tilde{W}x_t - \gamma g_t - \textbf{1}\otimes\bar{x}_t\|^2 \nonumber \\
 & \leq (1+\alpha_2)\nu^2\|x_t -\textbf{1}\otimes\bar{x}_t\|^2 + (1+\frac{1}{\alpha_2})\gamma^2\|g_t\|^2 \nonumber \\
 & \mathop{=}^{(i)} \frac{1+\nu^2}{2}\sum_{i=1}^m\|x^i_t -\bar{x}_t\|^2 + \frac{1+\nu^2}{1-\nu^2}\gamma^2\sum_{i=1}^m\|g^i_t\|^2,
\end{align}
where the last equality $(i)$ holds by $\alpha_2=\frac{1-\nu^2}{2\nu^2}$.
Then we have
\begin{align}
 \sum_{i=1}^m\|\tilde{x}^i_{t+1}-x^i_t\|^2 & =\|\tilde{x}_{t+1}-x_t\|^2 \nonumber \\
 & = \|\tilde{x}_{t+1}- \textbf{1}\otimes\bar{x}_t + \textbf{1}\otimes\bar{x}_t - x_t\|^2 \nonumber \\
 & \leq 2\|\tilde{x}_{t+1} - \textbf{1}\otimes\bar{x}_t\|^2 + 2\|x_t - \textbf{1}\otimes\bar{x}_t\|^2 \nonumber \\
 & = (3+\nu^2)\sum_{i=1}^m\|x^i_t -\bar{x}_t\|^2 + \frac{2(1+\nu^2)}{1-\nu^2}\gamma^2\sum_{i=1}^m\|g^i_t\|^2.
\end{align}

Let $w_t=[(w^1_t)^T,(w^2_t)^T,\cdots,(w^m_t)^T]^T$,
$u_t=[(u^1_t)^T,(u^2_t)^T,\cdots,(u^m_t)^T]^T$ and $\bar{w}_t = \frac{1}{m}\sum_{i=1}^mw^i_t$ and
$\bar{u}_t = \frac{1}{m}\sum_{i=1}^mu^i_t$. Then we have for any $t\geq 1$,
\begin{align}
  w_{t+1} = \tilde{W}\big(w_t + u_{t+1} - u_t\big). \nonumber
\end{align}
According to the above proof of Lemma~\ref{lem:B1}, we have $\bar{w}_{t+1}=\bar{w}_t + \bar{u}_{t+1} - \bar{u}_t$ for all $t\geq1$.
Thus we have
\begin{align} \label{eq:D3}
     \sum_{i=1}^m\|w^i_{t+1}-\bar{w}_{t+1}\|^2 & =
     \|w_{t+1}-\textbf{1}\otimes\bar{w}_{t+1}\|^2 \nonumber \\
    & = \big\| \tilde{W}\big(w_t + u_{t+1} - u_t\big) -\textbf{1}\otimes\big(\bar{w}_t + \bar{u}_{t+1} - \bar{u}_t \big)\big\|^2 \nonumber \\
    & \leq (1+c)\|\tilde{W}w_t-\textbf{1}\otimes\bar{w}_t\|^2 + (1+\frac{1}{c})\big\|\tilde{W}\big( u_{t+1} - u_t\big)-\textbf{1}\otimes\big(\bar{u}_{t+1} - \bar{u}_t\big)\big\|^2 \nonumber \\
    & \leq (1+c)\nu^2\|w_t-\textbf{1}\otimes\bar{w}_t\|^2 + (1+\frac{1}{c})\nu^2\big\|u_{t+1} - u_t-\textbf{1}\otimes\big(\bar{u}_{t+1} - \bar{u}_t\big)\big\|^2 \nonumber \\
    & \leq (1+c)\nu^2\|w_t-\textbf{1}\otimes\bar{w}_t\|^2 + (1+\frac{1}{c})\nu^2\big\|u_{t+1} - u_t\big\|^2,
\end{align}
where the last inequality holds by Lemma~\ref{lem:A2}.

Since $u^i_{t+1} = \nabla f^i(x^i_{t+1};\xi^i_{t+1}) + (1-\beta_{t+1})(u^i_t-\nabla f^i(x^i_t;\xi^i_{t+1}))$ for any $i\in [m]$ and $t\geq 1$, we have
\begin{align} \label{eq:D4}
 \big\|u_{t+1} - u_t\big\|^2 & = \sum_{i=1}^m\|u^i_{t+1} - u^i_t\|^2 \nonumber \\
 &= \sum_{i=1}^m\|\nabla f^i(x^i_{t+1};\xi^i_{t+1}) - \beta_{t+1}u^i_t - (1-\beta_{t+1})\nabla f^i(x^i_t;\xi^i_{t+1})\|^2 \nonumber \\
 &= \sum_{i=1}^m\|\beta_{t+1}(\nabla f^i(x^i_{t+1};\xi^i_{t+1})-\nabla f^i(x^i_{t+1})) + \beta_{t+1}(\nabla f^i(x^i_{t+1})-\nabla f^i(x^i_t)) + \beta_{t+1}(\nabla f^i(x^i_t)-u^i_t) \nonumber \\
 & \qquad + (1-\beta_{t+1})(\nabla f^i(x^i_{t+1};\xi^i_{t+1})-\nabla f^i(x^i_t;\xi^i_{t+1}))\|^2 \nonumber \\
 &= 4\beta^2_{t+1}\sum_{i=1}^m\|\nabla f^i(x^i_{t+1};\xi^i_{t+1})-\nabla f^i(x^i_{t+1})\|^2 +4\beta^2_{t+1}\sum_{i=1}^m\|\nabla f^i(x^i_{t+1})-\nabla f^i(x^i_t)\|^2 \nonumber \\
 & \quad + 4\beta^2_{t+1}\sum_{i=1}^m\|\nabla f^i(x^i_t)- u^i_t \|^2 + 4(1-\beta_{t+1})^2\sum_{i=1}^m\|\nabla f^i(x^i_{t+1};\xi^i_{t+1})-\nabla f^i(x^i_t;\xi^i_{t+1})\|^2 \nonumber \\
 & \leq 4\beta_{t+1}^2\sum_{i=1}^m\|u^i_t-\nabla f^i(x^i_t)\|^2+ 4\beta^2_{t+1}m\sigma^2 + 4\beta_{t+1}^2L^2\sum_{i=1}^m\|x^i_{t+1}-x^i_t\|^2 \nonumber \\
 & \quad + 4(1-\beta_{t+1})^2L^2\sum_{i=1}^m\|x^i_{t+1}-x^i_t\|^2 \nonumber \\
 & \leq 4\beta_{t+1}^2\sum_{i=1}^m\|u^i_t-\nabla f^i(x^i_t)\|^2+ 4\beta^2_{t+1}m\sigma^2 + 8\eta^2_tL^2\sum_{i=1}^m\|\tilde{x}^i_{t+1}-x^i_t\|^2,
\end{align}
where the last inequality holds by $0<\beta_t<1$ and $x^i_{t+1}=x^i_t+\eta_t(\tilde{x}^i_{t+1}-x^i_t)$.

Plugging the above inequalities (\ref{eq:D4}) into (\ref{eq:D3}), we have
\begin{align}
    \sum_{i=1}^m\|w^i_{t+1}-\bar{w}_{t+1}\|^2
    & \leq (1+c)\nu^2\|w_t-\textbf{1}\otimes\bar{w}_t\|^2 + (1+\frac{1}{c})\nu^2\big\|u_{t+1} - u_t\big\|^2 \nonumber \\
    & \leq (1+c)\nu^2\|w_t-\textbf{1}\otimes\bar{w}_t\|^2 + (1+\frac{1}{c})\nu^2\big(4\beta_{t+1}^2\sum_{i=1}^m\|u^i_t-\nabla f^i(x^i_t)\|^2+ 4\beta^2_{t+1}m\sigma^2 \nonumber \\
    & \quad + 8\eta^2_tL^2\sum_{i=1}^m\|\tilde{x}^i_{t+1}-x^i_t\|^2\big).
\end{align}
Let $c=\frac{1}{\nu}-1$, we have
\begin{align}
   \sum_{i=1}^m\|w^i_{t+1}-\bar{w}_{t+1}\|^2
    & \leq \nu\sum_{i=1}^m\|w^i_t-\bar{w}_t\|^2 + \frac{\nu^2}{1-\nu}\big(4\beta_{t+1}^2\sum_{i=1}^m\|u^i_t-\nabla f^i(x^i_t)\|^2 \nonumber \\
    & \quad + 4\beta^2_{t+1}m\sigma^2 + 8\eta^2_tL^2\sum_{i=1}^m\|\tilde{x}^i_{t+1}-x^i_t\|^2\big).
\end{align}

\end{proof}

\begin{theorem}  \label{th:A1}
(Restatement of Theorem~\ref{th:1})
 Suppose the sequences $\big\{\{x^i_t\}_{i=1}^m\big\}_{t=1}^T$ be generated from Algorithm~\ref{alg:1}.
 Under the above Assumptions~\ref{ass:1}-\ref{ass:5}, and let $\eta_t=\eta$, $0<\beta_t\leq1$ for all $t\geq 0$, $\gamma\leq \min\big(\frac{\rho(1-\nu^2)}{48\theta_t},\frac{3\rho(1-\nu^2)\theta_t}{58L^2}\big)$, $\eta\leq \min\big(\frac{\rho\sqrt{1-\nu^2}}{4L\gamma\sqrt{3(1+\nu^2)}\sqrt{H_t}},\frac{\sqrt{\rho(1-\nu^2)\theta_t}}{2 L\sqrt{\gamma(3+\nu^2)}\sqrt{H_t}}\big)$ with $H_t=\frac{9}{2\beta_t}+\frac{8\nu^2}{(1-\nu)^2}$ for all $t\geq1$, we have
\begin{align}
\frac{1}{T}\sum_{t=1}^T\mathbb{E} \|\nabla F(\bar{x}_t)\|
  & \leq \frac{1}{T}\sum_{t=1}^T\frac{1}{m}\sum_{i=1}^m\mathbb{E}[ \|\nabla F(x^i_t)\| + L\|\bar{x}_t-x^i_t\|] \nonumber \\
  & \leq \Big(\frac{6\sqrt{G}}{\sqrt{T}}\!+\!\frac{12\sigma}{\rho}\sqrt{\frac{1}{T}\sum_{t=1}^TH_t\beta^2_t}\Big)
  \sqrt{\frac{1}{T}\sum_{t=1}^T\frac{1}{m}\sum_{i=1}^m\mathbb{E}\|A^i_t\|^2}.
\end{align}

where $G= \frac{F(\bar{x}_1)-F^*}{\rho\gamma\eta}+\big( \frac{4\nu^2}{\rho^2(1-\nu)}
 +\frac{9}{2\rho^2\beta_0}-\frac{9}{2\rho^2}\big)\sigma^2 $.
\end{theorem}

\begin{proof}
Without loss of generality, let $\eta=\eta_1=\cdots=\eta_T$.
According to Lemma \ref{lem:B2}, we have
\begin{align} \label{eq:H1}
 F(\bar{x}_{t+1}) \leq F(\bar{x}_t)+\frac{2\gamma\eta}{\rho}\|\nabla F(\bar{x}_t)-\bar{u}_t\|^2+\frac{\gamma\eta}{2\rho}\frac{1}{m}\sum_{i=1}^m\|w^i_t -\bar{w}_t\|^2-\frac{\rho\gamma\eta}{4} \frac{1}{m}\sum_{i=1}^m\|g^i_t\|^2.
\end{align}

According to the Lemma~\ref{lem:D1},
we have
 \begin{align} \label{eq:H2}
  \mathbb{E}\|\bar{u}_t - \overline{\nabla f(x_t)}\|^2
  \leq (1-\beta_t)\mathbb{E} \|\bar{u}_{t-1} -\overline{\nabla f(x_{t-1})}\|^2 + \frac{2\beta_t^2\sigma^2}{m} + \frac{2L^2\eta^2}{m^2}\sum_{i=1}^m\mathbb{E}\|\tilde{x}^i_t - x^i_{t-1}\|^2,
 \end{align}
and
 \begin{align} \label{eq:H3}
 \frac{1}{m}\sum_{i=1}^m \mathbb{E}\|u^i_t - \nabla f^i(x_t)\|^2
  \leq (1-\beta_t)\frac{1}{m}\sum_{i=1}^m \mathbb{E} \|u^i_{t-1} - \nabla f^i(x_{t-1})\|^2 + 2\beta_t^2\sigma^2 + 2L^2\eta^2\frac{1}{m}\sum_{i=1}^m\mathbb{E}\|\tilde{x}^i_t - x^i_{t-1}\|^2.
 \end{align}

According to Lemma \ref{lem:D2}, we have
\begin{align} \label{eq:H4}
\frac{1}{m}\sum_{i=1}^m\mathbb{E}\|w^i_t-\bar{w}_t\|^2
& \leq \nu\frac{1}{m}\sum_{i=1}^m\|w^i_{t-1}-\bar{w}_{t-1}\|^2 + \frac{4\nu^2}{1-\nu}\frac{1}{m}\sum_{i=1}^m\Big(2L^2\eta^2\|\tilde{x}^i_t-x^i_{t-1}\|^2 + \beta_t^2\|\nabla f^i(x^i_{t-1})-u^i_{t-1}\|^2 \nonumber \\
& \quad + \beta^2_tm\sigma^2\Big).
\end{align}
Meanwhile, we also have
\begin{align} \label{eq:H5}
\frac{1}{m}\sum_{i=1}^m\|x^i_t - \bar{x}_t\|^2 & \leq (1-\frac{(1-\nu^2)\eta}{2})\frac{1}{m}\sum_{i=1}^m\|x^i_{t-1}-\bar{x}_{t-1}\|^2
+ \frac{2\eta\gamma^2}{1-\nu^2}\frac{1}{m}\sum_{i=1}^m\| g^i_{t-1}-\bar{g}_{t-1} \|^2 \nonumber \\
& \leq (1-\frac{(1-\nu^2)\eta}{2})\frac{1}{m}\sum_{i=1}^m\|x^i_{t-1}-\bar{x}_{t-1}\|^2
+ \frac{2\eta\gamma^2}{1-\nu^2}\frac{1}{m}\sum_{i=1}^m(\| g^i_{t-1}\|^2 + \|\bar{g}_{t-1} \|^2) \nonumber \\
& \leq (1-\frac{(1-\nu^2)\eta}{2})\frac{1}{m}\sum_{i=1}^m\|x^i_{t-1}-\bar{x}_{t-1}\|^2
+ \frac{4\eta\gamma^2}{1-\nu^2}\frac{1}{m}\sum_{i=1}^m\|g^i_{t-1}\|^2.
\end{align}
and
\begin{align} \label{eq:H6}
\frac{1}{m}\sum_{i=1}^m\|\tilde{x}^i_t-x^i_{t-1}\|^2
 & \leq (3+\nu^2)\frac{1}{m}\sum_{i=1}^m\|x^i_{t-1} -\bar{x}_{t-1}\|^2 + \frac{2(1+\nu^2)}{1-\nu^2}\gamma^2
 \frac{1}{m}\sum_{i=1}^m\|g^i_{t-1}\|^2.
\end{align}

Next considering the term $\|\bar{u}_t-\nabla F(\bar{x}_t)\|^2$, we have
\begin{align}
\|\bar{u}_t-\nabla F(\bar{x}_t)\|^2 & = \|\bar{u}_t- \overline{\nabla f(x_t)} + \overline{\nabla f(x_t)} - \nabla F(\bar{x}_t)\|^2 \nonumber \\
& \leq 2\|\bar{u}_t- \overline{\nabla f(x_t)}\|^2 + 2\|\overline{\nabla f(x_t)} - \nabla F(\bar{x}_t)\|^2 \nonumber \\
& \leq 2\|\bar{u}_t- \overline{\nabla f(x_t)}\|^2 + 2\|\frac{1}{m}\sum_{i=1}^m\nabla f^i(x^i_t) - \frac{1}{m}\sum_{i=1}^m\nabla f^i(\bar{x}_t)\|^2 \nonumber \\
& \leq 2\|\bar{u}_t- \overline{\nabla f(x_t)}\|^2 + \frac{2L^2}{m}\sum_{i=1}^m\|x^i_t - \bar{x}_t\|^2, \nonumber
\end{align}
where the last inequality is due to Assumption~\ref{ass:1}. Then we can obtain
\begin{align} \label{eq:H7}
 - \|\bar{u}_t- \overline{\nabla f(x_t)}\|^2 \leq -\frac{1}{2}\|\bar{u}_t-\nabla F(\bar{x}_t)\|^2 + \frac{L^2}{m}\sum_{i=1}^m\|x^i_t - \bar{x}_t\|^2.
\end{align}
Since $\bar{u}_t=\bar{w}_t$ for all $t\geq1$, we have
\begin{align}
 \frac{1}{m}\sum_{i=1}^m\|w^i_t-\nabla F(x^i_t)\|^2 & = \frac{1}{m}\sum_{i=1}^m\|w^i_t-\bar{w}_t+\bar{u}_t-\nabla F(\bar{x}_t)+\nabla F(\bar{x}_t)-\nabla F(x^i_t)\|^2 \nonumber \\
 & \leq 3\frac{1}{m}\sum_{i=1}^m\|w^i_t-\bar{w}_t\|^2 + 3\|\bar{u}_t-\nabla F(\bar{x}_t)\|^2+3\frac{1}{m}\sum_{i=1}^m\|\nabla F(\bar{x}_t)-\nabla F(x^i_t)\|^2 \nonumber \\
 & \leq 3\frac{1}{m}\sum_{i=1}^m\|w^i_t-\bar{w}_t\|^2 + 3\|\bar{u}_t-\nabla F(\bar{x}_t)\|^2+3L^2\frac{1}{m}\sum_{i=1}^m\|x^i_t-\bar{x}_t\|^2. \nonumber
\end{align}
Then we have
\begin{align} \label{eq:H8}
 -\|\bar{u}_t-\nabla F(\bar{x}_t)\|^2
 & \leq -\frac{1}{3m}\sum_{i=1}^m\|w^i_t-\nabla F(x^i_t)\|^2 + \frac{1}{m}\sum_{i=1}^m\|w^i_t-\bar{w}_t\|^2 +\frac{L^2}{m}\sum_{i=1}^m\|x^i_t-\bar{x}_t\|^2.
\end{align}

We define a useful Lyapunov function (i.e., potential function), for any $t\geq 1$
\begin{align} \label{eq:H10}
\Omega_t & = \mathbb{E}_t \big[ F(\bar{x}_t) + (\lambda_{t-1}-\frac{9\gamma\eta}{2\rho})\|\bar{u}_{t-1} - \overline{\nabla f(x_{t-1})}\|^2 + \chi_{t-1}\frac{1}{m}\sum_{i=1}^m \|u^i_{t-1} - \nabla f^i(x^i_{t-1})\|^2 \\
& \qquad  + (\theta_{t-1}-\frac{19\gamma\eta L^2}{4\rho})\frac{1}{m}\sum_{i=1}^m\|x^i_{t-1}-\bar{x}_{t-1}\|^2  +(\vartheta_{t-1}-\frac{\gamma\eta}{4\rho})\frac{1}{m}\sum_{i=1}^m\|w^i_{t-1}-\bar{w}_{t-1}\|^2 \nonumber \\
& \qquad + \frac{\rho\gamma\eta}{6} \frac{1}{m}\sum_{i=1}^m\|g^i_{t-1}\|^2\big], \nonumber
\end{align}
where $\chi_{t-1}\geq 0$, $\lambda_{t-1}\geq \frac{9\gamma\eta}{2\rho}$,
$\theta_{t-1}\geq \frac{29\gamma\eta L^2}{6\rho}$ and $\vartheta_{t-1}\geq \frac{\gamma\eta}{4\rho}$ for all $t\geq1$.
Then we have
\begin{align} \label{eq:H11}
 \Omega_{t+1} & = \mathbb{E}_{t+1}\big[ F(\bar{x}_{t+1}) + (\lambda_t-\frac{9\gamma\eta}{2\rho})\|\bar{u}_t - \overline{\nabla f(x_t)}\|^2 + \chi_t\frac{1}{m}\sum_{i=1}^m \|u^i_t - \nabla f^i(x^i_t)\|^2  \nonumber \\
 & \qquad + (\theta_t-\frac{29\gamma\eta L^2}{6\rho})\frac{1}{m}\sum_{i=1}^m\|x^i_t-\bar{x}_t\|^2 +(\vartheta_t-\frac{3\gamma\eta}{4\rho})\frac{1}{m}\sum_{i=1}^m\|w^i_t-\bar{w}_t\|^2 + \frac{\rho\gamma\eta}{6} \frac{1}{m}\sum_{i=1}^m\|g^i_t\|^2 \big] \nonumber \\
 & \mathop{\leq}^{(i)} \mathbb{E}_{t+1}\Big[ F(\bar{x}_t) -\frac{\gamma\eta}{4\rho}\|\bar{u}_t - \nabla F(\bar{x}_t)\|^2 - \frac{\rho\gamma\eta}{12} \frac{1}{m}\sum_{i=1}^m\|g^i_t\|^2 + \frac{9\gamma\eta L^2}{2\rho}\frac{1}{m}\sum_{i=1}^m\|x^i_t - \bar{x}_t\|^2 + \lambda_t\|\bar{u}_t-\overline{\nabla f(x_t)}\|^2 \nonumber \\
 & \quad  + \chi_t\frac{1}{m}\sum_{i=1}^m \|u^i_t - \nabla f^i(x^i_t)\|^2 + (\theta_t-\frac{29\gamma\eta L^2}{6\rho})\frac{1}{m}\sum_{i=1}^m\|x^i_t-\bar{x}_t\|^2  +(\vartheta_t-\frac{\gamma\eta}{4\rho})\frac{1}{m}\sum_{i=1}^m\|w^i_t-\bar{w}_t\|^2 \Big] \nonumber \\
 & \mathop{\leq}^{(ii)} \mathbb{E}_{t+1}\Big[ F(\bar{x}_t) -\frac{\gamma\eta}{12\rho}\frac{1}{m}\sum_{i=1}^m\|w^i_t-\nabla F(x^i_t)\|^2 - \frac{\rho\gamma\eta}{12} \frac{1}{m}\sum_{i=1}^m\|g^i_t\|^2 -\frac{\gamma\eta L^2}{12\rho}\frac{1}{m}\sum_{i=1}^m\|x^i_t-\bar{x}_t\|^2 \nonumber \\
 & \quad + \lambda_t(1-\beta_t)\|\bar{u}_{t-1} -\overline{\nabla f(x_{t-1})}\|^2 + \frac{2\lambda_t\beta_t^2\sigma^2}{m} + \frac{2\lambda_tL^2\eta^2}{m^2}\sum_{i=1}^m\|\tilde{x}^i_t - x^i_{t-1}\|^2 \nonumber \\
 & \quad + \chi_t(1-\beta_t)\frac{1}{m}\sum_{i=1}^m \|u^i_{t-1} - \nabla f^i(x_{t-1})\|^2 + 2\chi_t\beta_t^2\sigma^2 + 2\chi_t L^2\eta^2\frac{1}{m}\sum_{i=1}^m \|\tilde{x}^i_t - x^i_{t-1}\|^2 \nonumber \\
 & \quad + \theta_t(1-\frac{(1-\nu^2)\eta}{2})\frac{1}{m}\sum_{i=1}^m\|x^i_{t-1}-\bar{x}_{t-1}\|^2
+ \theta_t\frac{4\eta\gamma^2}{1-\nu^2}\frac{1}{m}\sum_{i=1}^m\|g^i_{t-1}\|^2 \nonumber \\
& \quad + \vartheta_t\nu\frac{1}{m}\sum_{i=1}^m\|w^i_{t-1}-\bar{w}_{t-1}\|^2 + \frac{4\vartheta_t\nu^2}{1-\nu}\frac{1}{m}\sum_{i=1}^m\Big(2L^2\eta^2\|\tilde{x}^i_t-x^i_{t-1}\|^2 + \beta_t^2\|\nabla f^i(x^i_{t-1})-u^i_{t-1}\|^2 \nonumber \\
& \quad + \beta^2_tm\sigma^2\Big) \Big] ,
\end{align}
where the inequality (i) is due to the above inequalities (\ref{eq:H1}) and (\ref{eq:H7}); and
the inequality (ii) holds by the above inequalities (\ref{eq:H2}), (\ref{eq:H3}), (\ref{eq:H5}) and (\ref{eq:H8}).

Then we have
\begin{align} \label{eq:H12}
\Omega_{t+1} & \leq  \mathbb{E}_{t+1}\Big[ F(\bar{x}_t) -\frac{\gamma\eta}{12\rho}\frac{1}{m}\sum_{i=1}^m\|w^i_t-\nabla F(x^i_t)\|^2 - \frac{\rho\gamma\eta}{12} \frac{1}{m}\sum_{i=1}^m\|g^i_t\|^2  -\frac{\gamma\eta L^2}{12\rho}\frac{1}{m}\sum_{i=1}^m\|x^i_t-\bar{x}_t\|^2  \nonumber \\
 & \quad + \lambda_t(1-\beta_t)\|\bar{u}_{t-1}-\overline{\nabla f(x_{t-1})}\|^2  + \big(\chi_t -\chi_t\beta_t + \frac{4\vartheta_t\beta^2_t\nu^2}{1-\nu} \big)\frac{1}{m}\sum_{i=1}^m\|u^i_{t-1}-\nabla f^i(x^i_{t-1})\|^2 \nonumber \\
 & \quad + 2L^2\eta^2\big( \lambda_t + \chi_t + \frac{4\vartheta_t\nu^2}{1-\nu}\big)\frac{1}{m}\sum_{i=1}^m\|\tilde{x}^i_t-x^i_{t-1}\|^2  + \theta_t(1-\frac{(1-\nu^2)\eta}{2})\frac{1}{m}\sum_{i=1}^m\|x^i_{t-1}-\bar{x}_{t-1}\|^2 \nonumber \\
 & \quad
+ \theta_t\frac{4\eta\gamma^2}{1-\nu^2}\frac{1}{m}\sum_{i=1}^m\|g^i_{t-1}\|^2  + \vartheta_t\nu\frac{1}{m}\sum_{i=1}^m\|w^i_{t-1}-\bar{w}_{t-1}\|^2 + \big(\frac{2\lambda_t}{m}+2\chi_t+\frac{4\vartheta_t\nu^2}{1-\nu}\big)\beta_t^2\sigma^2 \Big] \nonumber \\
& \mathop{\leq}^{(i)} \mathbb{E}_{t+1}\Big[ F(\bar{x}_t) -\frac{\gamma\eta}{12\rho}\frac{1}{m}\sum_{i=1}^m\|w^i_t-\nabla F(x^i_t)\|^2 - \frac{\rho\gamma\eta}{12} \frac{1}{m}\sum_{i=1}^m\|g^i_t\|^2  -\frac{\gamma\eta L^2}{12\rho}\frac{1}{m}\sum_{i=1}^m\|x^i_t-\bar{x}_t\|^2  \nonumber \\
 & \quad + \lambda_t(1-\beta_t)\|\bar{u}_{t-1}-\overline{\nabla f(x_{t-1})}\|^2  + \big(\chi_t -\chi_t\beta_t + \frac{4\vartheta_t\beta^2_t\nu^2}{1-\nu} \big)\frac{1}{m}\sum_{i=1}^m\|u^i_{t-1}-\nabla f^i(x^i_{t-1})\|^2 \nonumber \\
 & \quad + 2L^2\eta^2\big( \lambda_t + \chi_t + \frac{4\vartheta_t\nu^2}{1-\nu}\big)\Big( (3+\nu^2)\frac{1}{m}\sum_{i=1}^m\|x^i_{t-1} -\bar{x}_{t-1}\|^2 + \frac{2(1+\nu^2)}{1-\nu^2}\gamma^2
 \frac{1}{m}\sum_{i=1}^m\|g^i_{t-1}\|^2 \Big)  \nonumber \\
& \quad + \theta_t(1-\frac{(1-\nu^2)\eta}{2})\frac{1}{m}\sum_{i=1}^m\|x^i_{t-1}-\bar{x}_{t-1}\|^2 + \theta_t\frac{4\eta\gamma^2}{1-\nu^2}\frac{1}{m}\sum_{i=1}^m\|g^i_{t-1}\|^2\nonumber \\
 & \quad
  + \vartheta_t\nu\frac{1}{m}\sum_{i=1}^m\|w^i_{t-1}-\bar{w}_{t-1}\|^2 + \big(\frac{2\lambda_t}{m}+2\chi_t+\frac{4\vartheta_t\nu^2}{1-\nu}\big)\beta_t^2\sigma^2 \Big],
\end{align}
where the inequality (i) is due to the above inequality (\ref{eq:H6}).

Since $0<\beta_t < 1$ for all $t\geq1$, $\lambda_t = \frac{9\gamma\eta}{2\rho\beta_t}\geq \frac{9\gamma\eta}{2\rho}$ and $\lambda_t \leq \lambda_{t-1}$, then we have $\lambda_t(1-\beta_t)\leq \lambda_{t-1} - \frac{9\gamma\eta}{2\rho}$. Let $\chi_t =\frac{4\vartheta_t\nu^2}{1-\nu}\geq \frac{4\vartheta_t\beta_t\nu^2}{1-\nu}$ and $\chi_t\leq \chi_{t-1}$, we have $\chi_t -\chi_t\beta_t + \frac{4\vartheta_t\beta^2_t\nu^2}{1-\nu} \leq \chi_{t-1}$. Let $\vartheta_t=\frac{\gamma\eta}{\rho(1-\nu)}$ for all $t\geq1$, since $0<\nu<1$, we have $\vartheta_t\nu=\vartheta_t-(1-\nu)\vartheta_t\leq \vartheta_t-\frac{3\gamma\eta}{4\rho}=\vartheta_{t-1}-\frac{3\gamma\eta}{4\rho}$.
Meanwhile, let $\theta_t\leq \theta_{t-1}$ for all $t\geq1$, $\gamma\leq \min\big(\frac{\rho(1-\nu^2)}{48\theta_t},\frac{3\rho(1-\nu^2)\theta_t}{58L^2}\big)$, $\eta\leq \min\big(\frac{\rho\sqrt{1-\nu^2}}{4L\gamma\sqrt{3(1+\nu^2)}\sqrt{H_t}},\frac{\sqrt{\rho(1-\nu^2)\theta_t}}{2 L\sqrt{\gamma(3+\nu^2)}\sqrt{H_t}}\big)$ with $H_t=\frac{9}{2\beta_t}+\frac{8\nu^2}{(1-\nu)^2}$ for all $t\geq1$, we can obtain
\begin{align}
& \theta_t(1-\frac{(1-\nu^2)\eta}{2}) + 2L^2\eta^2(3+\nu^2)\big( \lambda_t + \chi_t + \frac{4\vartheta_t\nu^2}{1-\nu}\big) \leq \theta_{t-1} - \frac{29\gamma\eta L^2}{6\rho}, \\
& \theta_t\frac{4\eta\gamma^2}{1-\nu^2} +\frac{4(1+\nu^2)}{1-\nu^2}L^2\eta^2\gamma^2\big( \lambda_t + \chi_t + \frac{4\vartheta_t\nu^2}{1-\nu}\big)  \leq \frac{\rho\gamma\eta}{6}.
\end{align}

Based on the choice of these parameters and the above inequality (\ref{eq:H12}), we have
\begin{align} \label{eq:H13}
  \Omega_{t+1} & \leq \mathbb{E}_{t+1}\Big[ F(\bar{x}_t) -\frac{\gamma\eta}{12\rho}\frac{1}{m}\sum_{i=1}^m\|w^i_t-\nabla F(x^i_t)\|^2 - \frac{\rho\gamma\eta}{12} \frac{1}{m}\sum_{i=1}^m\|g^i_t\|^2  -\frac{\gamma\eta L^2}{12\rho}\frac{1}{m}\sum_{i=1}^m\|x^i_t-\bar{x}_t\|^2  \nonumber \\
 & \quad + (\lambda_{t-1}-\frac{9\gamma\eta}{2\rho})\|\bar{u}_{t-1}-\overline{\nabla f(x_{t-1})}\|^2  + \chi_{t-1}\frac{1}{m}\sum_{i=1}^m\|u^i_{t-1}-\nabla f^i(x^i_{t-1})\|^2 \nonumber \\
 & \quad  +  (\theta_{t-1} - \frac{29\gamma\eta L^2}{6\rho})\frac{1}{m}\sum_{i=1}^m\|x^i_{t-1}-\bar{x}_{t-1}\|^2 + (\vartheta_{t-1}-\frac{3\gamma\eta}{4\rho})\frac{1}{m}\sum_{i=1}^m\|w^i_{t-1}-\bar{w}_{t-1}\|^2 \nonumber \\
 & \quad + \frac{\rho\gamma\eta}{6}\frac{1}{m}\sum_{i=1}^m\|g^i_{t-1}\|^2 + \frac{2\gamma\eta H_t}{\rho}\beta^2_t\sigma^2  \Big] \nonumber \\
 & = \Omega_t  -\frac{\gamma\eta}{12\rho}\frac{1}{m}\sum_{i=1}^m\|w^i_t-\nabla F(x^i_t)\|^2 - \frac{\rho\gamma\eta}{12} \frac{1}{m}\sum_{i=1}^m\|g^i_t\|^2  -\frac{\gamma\eta L^2}{12\rho}\frac{1}{m}\sum_{i=1}^m\|x^i_t-\bar{x}_t\|^2 + \frac{2\gamma\eta H_t}{\rho}\beta^2_t\sigma^2.
\end{align}
Then we can obtain
\begin{align} \label{eq:H14}
\frac{1}{\rho^2}\frac{1}{m}\sum_{i=1}^m\|w^i_t-\nabla F(x^i_t)\|^2 +\frac{L^2}{\rho^2}\frac{1}{m}\sum_{i=1}^m\|x^i_t-\bar{x}_t\|^2 + \frac{1}{m}\sum_{i=1}^m\|g^i_t\|^2  \leq \frac{12(\Omega_t-\Omega_{t+1})}{\gamma\rho\eta} + \frac{48H_t}{\rho^2}\beta^2_t\sigma^2.
\end{align}

Since $x^1_0=\tilde{x}^1_0=\cdots=x^m_0=\tilde{x}^m_0$, $u^1_0=\cdots=u^m_0$ and
$w^1_0=\cdots=w^m_0=0$, we have
\begin{align} \label{eq:H15}
 \Omega_1 & = \mathbb{E}\big[ F(\bar{x}_1) + (\lambda_{0}-\frac{9\gamma\eta}{2\rho})\|\bar{u}_0 - \overline{\nabla f(x_0)}\|^2
 + \chi_0\frac{1}{m}\sum_{i=1}^m\|u^i_0-\nabla f^i(x^i_0)\|^2 \nonumber \\
 & \leq F(\bar{x}_1)+\big(\chi_0+\lambda_0-\frac{9\gamma\eta}{2\rho}\big)\sigma^2\nonumber \\
 & = F(\bar{x}_1)+\big(\frac{4\gamma\eta\nu^2}{\rho(1-\nu)^2}
 +\frac{9\gamma\eta}{2\rho\beta_0}-\frac{9\gamma\eta}{2\rho}\big)\sigma^2.
\end{align}

Let $\mathcal{M}^i_t = \|g^i_t\|
+ \frac{1}{\rho}\|\nabla F(x^i_t)-w^i_t\| + \frac{L}{\rho}\|\bar{x}_t-x^i_t\|$.
Then we have
\begin{align}
\mathcal{M}^i_t & = \|g^i_t\|
+ \frac{1}{\rho}\|\nabla F(x^i_t)-w^i_t\| + \frac{L}{\rho}\|\bar{x}_t-x^i_t\| \nonumber \\
& \mathop{=}^{(i)} \|(A^i_t)^{-1}w^i_t\|
+ \frac{1}{\rho}\|\nabla F(x^i_t)-w^i_t\| + \frac{L}{\rho}\|\bar{x}_t-x^i_t\| \nonumber \\
& = \frac{1}{\|A^i_t\|}\|A^i_t\|\|(A^i_t)^{-1}w^i_t\|
+ \frac{1}{\rho}\|\nabla F(x^i_t)-w^i_t\| + \frac{L}{\rho}\|\bar{x}_t-x^i_t\| \nonumber \\
& \geq \frac{1}{\|A^i_t\|}\|w^i_t\|
+ \frac{1}{\rho}\|\nabla F(x^i_t)-w^i_t\| + \frac{L}{\rho}\|\bar{x}_t-x^i_t\| \nonumber \\
& \mathop{\geq}^{(ii)} \frac{1}{\|A^i_t\|}\|w^i_t\|
+ \frac{1}{\|A^i_t\|}\|\nabla F(x^i_t)-w^i_t\| + \frac{L}{\|A^i_t\|}\|\bar{x}_t-x^i_t\| \nonumber \\
& \geq \frac{1}{\|A^i_t\|}\big(\|\nabla  F(x^i_t)\| + L\|\bar{x}_t-x^i_t\|\big),
\end{align}
where the equality $(i)$ holds by $g^i_t = (A^i_t)^{-1}w^i_t$, and the inequality $(ii)$ holds by $\|A^i_t\| \geq \rho$ for all $t\geq1$ due to Assumption \ref{ass:4}.
Then we have
 \begin{align} \label{eq:H16}
  \|\nabla F(x^i_t)\| + L\|\bar{x}_t-x^i_t\| \leq \mathcal{M}^i_t \|A^i_t\|.
 \end{align}

According to the above inequality (\ref{eq:H14}), we have
\begin{align} \label{eq:H17}
\frac{1}{T}\sum_{t=1}^T\frac{1}{m}\sum_{i=1}^m\mathbb{E}[\mathcal{M}^i_t]^2  &
\leq \frac{1}{T}\sum_{t=1}^T\big[\frac{3}{\rho^2}\frac{1}{m}\sum_{i=1}^m\|w^i_t-\nabla F(x^i_t)\|^2 +\frac{3L^2}{\rho^2}\frac{1}{m}\sum_{i=1}^m\|x^i_t-\bar{x}_t\|^2 + \frac{3}{m}\sum_{i=1}^m\|g^i_t\|^2 \big] \nonumber \\
& \leq \frac{1}{T}\sum_{t=1}^T\frac{36(\Omega_t-\Omega_{t+1})}{\gamma\rho\eta} \leq \frac{36(\Omega_1-F^*)}{T\gamma\rho\eta} + \frac{1}{T}\sum_{t=1}^T\frac{144H_t}{\rho^2}\beta^2_t\sigma^2.
\end{align}
By using Cauchy-Schwarz inequality to the above inequality~(\ref{eq:H16}), we have
 \begin{align} \label{eq:H18}
  \frac{1}{T}\sum_{t=1}^T\frac{1}{m}\sum_{i=1}^m\mathbb{E}[ \|\nabla F(x^i_t)\| + L\|\bar{x}_t-x^i_t\|] & \leq \frac{1}{T}\sum_{t=1}^T\frac{1}{m}\sum_{i=1}^m\mathbb{E}\big[\mathcal{M}^i_t \|A^i_t\|\big] \nonumber \\
  & \leq \sqrt{\frac{1}{T}\sum_{t=1}^T\frac{1}{m}\sum_{i=1}^m\mathbb{E}[\mathcal{M}^i_t]^2} \sqrt{\frac{1}{T}\sum_{t=1}^T\frac{1}{m}\sum_{i=1}^m\mathbb{E}\|A^i_t\|^2}.
 \end{align}

By plugging the above inequalities \eqref{eq:H18} into \eqref{eq:H17}, we can obtain
\begin{align}
  \frac{1}{T}\sum_{t=1}^T\frac{1}{m}\sum_{i=1}^m\mathbb{E}[ \|\nabla F(x^i_t)\| + L\|\bar{x}_t-x^i_t\|] \leq \big(\frac{6\sqrt{\Omega_1-F^*}}{\sqrt{T\gamma\rho\eta}}+\frac{12\sigma\sqrt{\frac{1}{T}\sum_{t=1}^TH_t\beta^2_t}
  }{\rho}\big)\sqrt{\frac{1}{T}\sum_{t=1}^T\frac{1}{m}\sum_{i=1}^m\mathbb{E}\|A^i_t\|^2}.
\end{align}

Since $F(x)$ is $L$-smooth, we have
\begin{align}
 \|\nabla F(\bar{x}_t)\| = \|\nabla F(\bar{x}_t) - \nabla F(x^i_t) + \nabla F(x^i_t)\| \leq \|\nabla F(x^i_t)\| + L\|x^i_t - \bar{x}_t\|.
\end{align}
Meanwhile, let $G= \frac{F(\bar{x}_1)-F^*}{\rho\gamma\eta}+\big( \frac{4\nu^2}{\rho^2(1-\nu)}
 +\frac{9}{2\rho^2\beta_0}-\frac{9}{2\rho^2}\big)\sigma^2 $. Then we have
\begin{align}
  \frac{1}{T}\sum_{t=1}^T\mathbb{E} \|\nabla F(\bar{x}_t)\|
  & \leq \frac{1}{T}\sum_{t=1}^T\frac{1}{m}\sum_{i=1}^m\mathbb{E}[ \|\nabla F(x^i_t)\| + L\|\bar{x}_t-x^i_t\|] \nonumber \\
  & \leq \big(\frac{6\sqrt{\Omega_1-F^*}}{\sqrt{T\gamma\rho\eta}}+\frac{12\sigma\sqrt{\frac{1}{T}\sum_{t=1}^TH_t\beta^2_t}}{\rho}\big)\sqrt{\frac{1}{T}\sum_{t=1}^T\frac{1}{m}\sum_{i=1}^m\mathbb{E}\|A^i_t\|^2} \nonumber \\
  & = \big(\frac{6\sqrt{G}}{\sqrt{T}}+\frac{12\sigma\sqrt{\frac{1}{T}\sum_{t=1}^TH_t\beta^2_t}}{\rho}\big)\sqrt{\frac{1}{T}\sum_{t=1}^T\frac{1}{m}\sum_{i=1}^m\mathbb{E}\|A^i_t\|^2}.
\end{align}

Let $\beta_t = \frac{1}{T^{2/3}}$ for all $t\geq1$, we have
\begin{align}
 H_t = \frac{9}{2\beta_t}+\frac{8\nu^2}{(1-\nu)^2}=O(T^{2/3}).
\end{align}
We set $\theta_t=\theta\geq \frac{29\gamma\eta L^2}{6\rho}$ for all $t\geq1$.
Then we can set $\gamma=O(1)$ and $\eta=O(\frac{1}{T^{1/3}})$.
Thus we have $G= \frac{F(\bar{x}_1)-F^*}{\rho\gamma\eta}+\big( \frac{4\nu^2}{\rho^2(1-\nu)}
 +\frac{9}{2\rho^2\beta_0}-\frac{9}{2\rho^2}\big)\sigma^2=O(T^{1/3})$, and then we can obtain
\begin{align}
  \frac{1}{T}\sum_{t=1}^T \mathbb{E}\|\nabla F(\bar{x}_t)\|
  & \leq \frac{1}{T}\sum_{t=1}^T\frac{1}{m}\sum_{i=1}^m\mathbb{E}[ \|\nabla F(x^i_t)\| + L\|\bar{x}_t-x^i_t\|] \nonumber \\
  & \leq O(\frac{1}{T^{1/3}}+\frac{\sigma}{T^{1/3}})\sqrt{\frac{1}{T}\sum_{t=1}^T\frac{1}{m}\sum_{i=1}^m\mathbb{E}\|A^i_t\|^2}.
\end{align}
Given $\sqrt{\frac{1}{T}\sum_{t=1}^T\frac{1}{m}\sum_{i=1}^m\mathbb{E}\|A^i_t\|^2}=O(1)$,
set
\begin{align}
  \frac{1}{T}\sum_{t=1}^T \mathbb{E}\|\nabla F(\bar{x}_t)\|
  & \leq O(\frac{1}{T^{1/3}}+\frac{\sigma}{T^{1/3}})\sqrt{\frac{1}{T}\sum_{t=1}^T\frac{1}{m}\sum_{i=1}^m\mathbb{E}\|A^i_t\|^2} \leq \epsilon,
\end{align}
we have $T= O(\epsilon^{-3})$. Since our Algorithm~\ref{alg:1}, it reaches a near-optimal sample complexity of $1\cdot T=O(\epsilon^{-3})$ for finding an $\epsilon$-stationary solution of Problem~(\ref{eq:1}).

\end{proof}

\subsection{ Convergence Analysis of AdaMDOF Algorithm}
In this subsection, we provide the convergence analysis of our AdaMDOF Algorithm for \textbf{finite-sum} optimization.

\begin{lemma} \label{lem:C1}
Under the above Assumptions~\ref{ass:1}-\ref{ass:2}, and assume the gradient estimators
$\big\{u^i_{t\geq 1}\big\}_{i=1}^m$ be generated from Algorithm \ref{alg:2},
we have
 \begin{align}
 \mathbb{E}_t\|u^i_t - \nabla f^i(x^i_t)\|^2
 & \leq  (1-\beta_t)\|u^i_{t-1}-\nabla f^i(x^i_{t-1})\|^2 + \frac{2\beta^2_t}{b}\frac{1}{n}\sum_{k=1}^n\|\nabla f^i_k(x^i_{t-1})-z^i_{k,t-1}\|^2 \nonumber \\
 & \quad + \frac{2L^2\eta^2_{t-1}}{b}\|\tilde{x}^i_t-x^i_{t-1}\|^2, \ \forall i\in [m]  \\
 \mathbb{E}_t\|\bar{u}_t - \overline{\nabla f(x_t)}\|^2 & \leq (1-\beta_t)\|\bar{u}_{t-1}-\overline{\nabla f(x_{t-1})}\|^2 + \frac{2L^2\eta_t^2}{bm}\sum_{i=1}^m\|\tilde{x}^i_t-x^i_{t-1}\|^2 \nonumber \\
 & \quad + \frac{2\beta_t^2}{bm}\sum_{i=1}^m\frac{1}{n}\sum_{k=1}^n\|\nabla f^i_k(x^i_{t-1}) -z^i_{k,t-1} \|^2,
 \end{align}
 where $\beta_t\in(0,1)$ for all $t\geq1$.
\end{lemma}

\begin{proof}
According the line 6 of Algorithm~\ref{alg:2}, i.e,
$u^i_t = \frac{1}{b}\sum_{k\in \mathcal{I}^i_t}\big(\nabla f^i_k(x^i_t)-\nabla f^i_k(x^i_{t-1})\big) + (1-\beta_t) u^i_{t-1} + \beta_t\Big(\frac{1}{b}\sum_{k\in \mathcal{I}^i_t}\big(\nabla f^i_k(x^i_{t-1})-z^i_{k,t-1}\big) + \frac{1}{n}\sum_{j=1}^nz^i_{j,t-1}\Big)$,
we have
\begin{align}
 &\mathbb{E}_t\|u^i_t - \nabla f^i(x^i_t)\|^2  \\
 & = \mathbb{E}_t\big\|\frac{1}{b}\sum_{k\in \mathcal{I}^i_t}\big(\nabla f^i_k(x^i_t)-\nabla f^i_k(x^i_{t-1})\big) + (1-\beta_t) u^i_{t-1} + \beta_t\Big(\frac{1}{b}\sum_{k\in \mathcal{I}^i_t}\big(\nabla f^i_k(x^i_{t-1})-z^i_{k,t-1}\big) + \frac{1}{n}\sum_{j=1}^nz^i_{j,t-1}\Big)- \nabla f^i(x^i_t)\big\|^2 \nonumber \\
 & = \mathbb{E}_t\big\|\frac{1}{b}\sum_{k\in \mathcal{I}^i_t}\big(\nabla f^i_k(x^i_t)-\nabla f^i_k(x^i_{t-1})\big) - \nabla f^i(x^i_t)+ \nabla f^i(x^i_{t-1}) + (1-\beta_t)(u^i_{t-1}-\nabla f^i(x^i_{t-1}))\nonumber \\
 & \quad + \beta_t\big( \frac{1}{b}\sum_{k\in \mathcal{I}^i_t}\big(\nabla f^i_k(x^i_{t-1})-z^i_{k,t-1}\big) + \frac{1}{n}\sum_{j=1}^nz^i_{j,t-1} - \nabla f^i(x^i_{t-1}) \big) \big\|^2 \nonumber \\
 & = \mathbb{E}_t\big\|\frac{1}{b}\sum_{k\in \mathcal{I}^i_t}\big(\nabla f^i_k(x^i_t)-\nabla f^i_k(x^i_{t-1})\big) - \nabla f^i(x^i_t)+ \nabla f^i(x^i_{t-1}) + \beta_t\big( \frac{1}{b}\sum_{k\in \mathcal{I}^i_t}\big(\nabla f^i_k(x^i_{t-1})-z^i_{k,t-1}\big) \nonumber \\
 & \quad  + \frac{1}{n}\sum_{j=1}^nz^i_{j,t-1} - \nabla f^i(x^i_{t-1}) \big) \big\|^2 + (1-\beta_t)^2\|u^i_{t-1}-\nabla f^i(x^i_{t-1})\|^2 \nonumber \\
 & \leq 2\mathbb{E}_t\big\|\frac{1}{b}\sum_{k\in \mathcal{I}^i_t}\big(\nabla f^i_k(x^i_t)-\nabla f^i_k(x^i_{t-1})\big) - \nabla f^i(x^i_t)+ \nabla f^i(x^i_{t-1})\big\|^2 \nonumber \\
 & \quad + 2\beta^2_t\mathbb{E}_t\big\|\frac{1}{b}\sum_{k\in \mathcal{I}^i_t}\big(\nabla f^i_k(x^i_{t-1})-z^i_{k,t-1}\big) + \frac{1}{n}\sum_{j=1}^nz^i_{j,t-1} - \nabla f^i(x^i_{t-1}) \big\|^2 + (1-\beta_t)^2\|u^i_{t-1}-\nabla f^i(x^i_{t-1})\|^2 \nonumber \\
 & \leq \frac{2L^2}{b}\|x^i_t-x^i_{t-1}\|^2 + \frac{2\beta^2_t}{b}\frac{1}{n}\sum_{k=1}^n\|z^i_{k,t-1}-\nabla f^i_k(x^i_{t-1})\|^2+ (1-\beta_t)^2\|u^i_{t-1}-\nabla f^i(x^i_{t-1})\|^2 \nonumber \\
 & \leq \frac{2L^2\eta^2_{t-1}}{b}\|\tilde{x}^i_t-x^i_{t-1}\|^2 + \frac{2\beta^2_t}{b}\frac{1}{n}\sum_{k=1}^n\|z^i_{k,t-1}-\nabla f^i_k(x^i_{t-1})\|^2+ (1-\beta_t)\|u^i_{t-1}-\nabla f^i(x^i_{t-1})\|^2,
\end{align}
where the third equality holds by
\begin{align}
& \mathbb{E}_{t} \big[\frac{1}{b}\sum_{k\in \mathcal{I}^i_t}\big(\nabla f^i_k(x^i_t)-\nabla f^i_k(x^i_{t-1})\big) \big] = \nabla f^i(x^i_t)- \nabla f^i(x^i_{t-1}), \nonumber \\
& \mathbb{E}_{t} \big[ \frac{1}{b}\sum_{k\in \mathcal{I}^i_t}\big(\nabla f^i_k(x^i_{t-1})-z^i_{k,t-1}\big) \big] =\nabla f^i(x^i_{t-1}) - \frac{1}{n}\sum_{j=1}^nz^j_{t-1}, \nonumber
\end{align}
and the second last inequality holds by the inequality $\mathbb{E}\|\zeta-\mathbb{E}[\zeta]\|^2 \leq \mathbb{E}\|\zeta\|^2$ and Assumption \ref{ass:1};
the last inequality holds by $0<\beta_t \leq 1$ and $x^i_t=x^i_{t-1}+\eta_{t-1}(\tilde{x}_t-x^i_{t-1})$.

Let $\bar{u}_t=\frac{1}{m}\sum_{i=1}^m u^i_t$, $\nabla f^i(x^i_t)=\frac{1}{n}\sum_{k=1}^n\nabla f^i_k(x^i_t)$ for all $i\in [m]$ and $\overline{\nabla f(x_t)}=\frac{1}{m}\sum_{i=1}^m (\frac{1}{n}\sum_{k=1}^n\nabla f^i_k(x^i_t))=
\frac{1}{m}\sum_{i=1}^m \nabla f^i(x^i_t)$.
We can obtain
\begin{align}
 & \mathbb{E}_t\|\bar{u}_t - \overline{\nabla f(x_t)}\|^2 =  \mathbb{E}_t\big\|\frac{1}{m}\sum_{i=1}^m(u^i_t - \nabla f^i(x^i_t))\big\|^2 \nonumber \\
 & = \mathbb{E}_t\big\|\frac{1}{m}\sum_{i=1}^m\Big(\frac{1}{b}\sum_{k\in \mathcal{I}^i_t}\big(\nabla f^i_k(x^i_t)-\nabla f^i_k(x^i_{t-1})\big) - \nabla f^i(x^i_t)+ \nabla f^i(x^i_{t-1}) + (1-\beta_t)(u^i_{t-1}-\nabla f^i(x^i_{t-1}))\nonumber \\
 & \quad + \beta_t\big( \frac{1}{b}\sum_{k\in \mathcal{I}^i_t}\big(\nabla f^i_k(x^i_{t-1})-z^i_{k,t-1}\big) + \frac{1}{n}\sum_{j=1}^nz^i_{j,t-1} - \nabla f^i(x^i_{t-1}) \big) \Big)\big\|^2 \nonumber \\
 & \mathop{=}^{(i)} (1-\beta_t)^2\|\bar{u}_{t-1}-\overline{\nabla f(x_{t-1})}\|^2 + \mathbb{E}_t\big\|\frac{1}{m}\sum_{i=1}^m\big(\frac{1}{b}\sum_{k\in \mathcal{I}^i_t}\big(\nabla f^i_k(x^i_t)-\nabla f^i_k(x^i_{t-1})\big) - \nabla f^i(x^i_t)+ \nabla f^i(x^i_{t-1}) \nonumber \\
 & \quad + \beta_t\big( \frac{1}{b}\sum_{k\in \mathcal{I}^i_t}\big(\nabla f^i_k(x^i_{t-1})-z^i_{k,t-1}\big) + \frac{1}{n}\sum_{j=1}^nz^i_{j,t-1} - \nabla f^i(x^i_{t-1}) \big)\big)\big\|^2 \nonumber \\
 & \leq (1-\beta_t)^2\|\bar{u}_{t-1}-\overline{\nabla f(x_{t-1})}\|^2 + \frac{2}{m}\sum_{i=1}^m\mathbb{E}_t\big\|\frac{1}{b}\sum_{k\in \mathcal{I}^i_t}\big(\nabla f^i_k(x^i_t)-\nabla f^i_k(x^i_{t-1})\big) - \nabla f^i(x^i_t)+ \nabla f^i(x^i_{t-1}) \big\|^2 \nonumber \\
 & \quad + \frac{2\beta_t^2}{m}\sum_{i=1}^m\mathbb{E}_t\big\| \frac{1}{b}\sum_{k\in \mathcal{I}^i_t}\big(\nabla f^i_k(x^i_{t-1})-z^i_{k,t-1}\big) + \frac{1}{n}\sum_{j=1}^nz^i_{j,t-1} - \nabla f^i(x^i_{t-1}) \big\|^2
 \nonumber \\
 & \leq (1-\beta_t)^2\|\bar{u}_{t-1}-\overline{\nabla f(x_{t-1})}\|^2 + \frac{2L^2}{mb}\sum_{i=1}^m\|x^i_t-x^i_{t-1}\|^2 + \frac{2\beta_t^2}{mb}\sum_{i=1}^m\frac{1}{n}\sum_{k=1}^n\|\nabla f^i_k(x^i_{t-1}) -z^i_{k,t-1} \|^2 \nonumber \\
 & \mathop{\leq}^{(ii)} (1-\beta_t)\|\bar{u}_{t-1}-\overline{\nabla f(x_{t-1})}\|^2 + \frac{2L^2\eta_t^2}{bm}\sum_{i=1}^m\|\tilde{x}^i_t-x^i_{t-1}\|^2 + \frac{2\beta_t^2}{bm}\sum_{i=1}^m\frac{1}{n}\sum_{k=1}^n\|\nabla f^i_k(x^i_{t-1}) -z^i_{k,t-1} \|^2,
\end{align}
where the above equality $(i)$ is due to for all $i\in [m]$
\begin{align}
& \mathbb{E}_{t} \big[\frac{1}{b}\sum_{k\in \mathcal{I}^i_t}\big(\nabla f^i_k(x^i_t)-\nabla f^i_k(x^i_{t-1})\big) \big] = \nabla f^i(x^i_t)- \nabla f^i(x^i_{t-1}), \nonumber \\
& \mathbb{E}_{t} \big[ \frac{1}{b}\sum_{k\in \mathcal{I}^i_t}\big(\nabla f^i_k(x^i_{t-1})-z^i_{k,t-1}\big) \big] =\nabla f^i(x^i_{t-1}) - \frac{1}{n}\sum_{j=1}^nz^j_{t-1}, \nonumber
\end{align}
and $\{\mathcal{I}^i_t\}_{i=1}^m$ are independent, and the above inequality $(ii)$ holds by
$0<\beta_t<1$ and $x^i_t=x^i_{t-1}+\eta_t(\tilde{x}^i_t-x^i_{t-1})$.

\end{proof}

\begin{lemma} \label{lem:C2}
Under Assumption~\ref{ass:1}, the sequence $\{z^i_{k,t}\}$ is defined in Line 11 of Algorithm~\ref{alg:2}, then we have
\begin{align}
 \mathbb{E}_t\big[\frac{1}{n}\sum_{k=1}^n\|\nabla f^i_k(x^i_t)-z^i_{k,t}\|^2\big] & \leq (1-\frac{b}{2n})\frac{1}{n}\sum_{k=1}^n\|\nabla f^i_k(x^i_{t-1})-z^i_{k,t-1}\|^2 \nonumber \\
 & \quad + (\frac{2n}{b}-\frac{b}{n}-1)L^2\eta^2_{t-1}\|\tilde{x}^i_t-x^i_{t-1}\|^2, \ \forall i \in [m].
\end{align}
\end{lemma}

\begin{proof}
According to the Line 11 of Algorithm~\ref{alg:2}, we have
\begin{align}
 \mathbb{E}_t\big[\frac{1}{n}\sum_{k=1}^n\|\nabla f^i_k(x^i_t)-z^i_{k,t}\|^2\big] & =
 (1-\frac{b}{n})\frac{1}{n}\sum_{k=1}^n \|\nabla f^i_k(x^i_t)-z^i_{k,t-1}\|^2 \\
 & = (1-\frac{b}{n})\frac{1}{n}\sum_{k=1}^n \|\nabla f^i_k(x^i_t) - \nabla f^i_k(x^i_{t-1}) + \nabla f^i_k(x^i_{t-1})-z^i_{k,t-1}\|^2 \nonumber \\
 & \leq (1+\frac{1}{\alpha})(1-\frac{b}{n})\frac{1}{n}\sum_{k=1}^n \|\nabla f^i_k(x^i_t) -\nabla f^i_k(x^i_{t-1})\|^2 \nonumber \\
 & \quad + (1+\alpha)(1-\frac{b}{n})\frac{1}{n}\sum_{k=1}^n\|\nabla f^i_k(x^i_{t-1})-z^i_{k,t-1}\|^2
 \nonumber \\
 & \leq (1+\frac{1}{\alpha})(1-\frac{b}{n})L^2\|x^i_t-x^i_{t-1}\|^2+ (1+\alpha)(1-\frac{b}{n})\frac{1}{n}\sum_{k=1}^n\|\nabla f^i_k(x^i_{t-1})-z^i_{k,t-1}\|^2, \nonumber
\end{align}
where the last inequality is due to Assumption~\ref{ass:1}.

Let $\alpha=\frac{b}{2n}$, then we have
\begin{align}
 \mathbb{E}_t\big[\frac{1}{n}\sum_{i=1}^n\|\nabla f^i(x^i_t)-z^i_t\|^2\big] & \leq
  (1-\frac{b}{2n})\frac{1}{n}\sum_{i=1}^n\|\nabla f^i(x^i_{t-1})-z^i_{t-1}\|^2 + (\frac{2n}{b}-\frac{b}{n}-1)L^2\|x^i_t-x^i_{t-1}\|^2  \\
 & = (1-\frac{b}{2n})\frac{1}{n}\sum_{i=1}^n\|\nabla f^i(x^i_{t-1})-z^i_{t-1}\|^2 + (\frac{2n}{b}-\frac{b}{n}-1)L^2\eta^2_{t-1}\|\tilde{x}_t-x^i_{t-1}\|^2, \nonumber
\end{align}
where the above equality holds by $x^i_t-x^i_{t-1} = \eta_{t-1}(\tilde{x}_t-x^i_{t-1})$.

\end{proof}

\begin{lemma} \label{lem:C3}
Given the sequence $\big\{x^i_{t\geq1},\tilde{x}^i_{t\geq1},w^i_{t\geq1}\big\}_{i=1}^m$ be generated from Algorithm \ref{alg:2}. We have
\begin{align}
\sum_{i=1}^m\|x^i_{t+1} - \bar{x}_{t+1}\|^2 & \leq (1-\frac{(1-\nu^2)\eta_t}{2})\sum_{i=1}^m\|x^i_t-\bar{x}_t\|^2
+ \frac{2\eta_t\gamma^2}{1-\nu^2}\sum_{i=1}^m\| g^i_t-\bar{g}_t \|^2,
\nonumber \\
\sum_{i=1}^m\|\tilde{x}^i_{t+1}-x^i_t\|^2
 & \leq (3+\nu^2)\sum_{i=1}^m\|x^i_t -\bar{x}_t\|^2 + \frac{2(1+\nu^2)}{1-\nu^2}\gamma^2
 \sum_{i=1}^m\|g^i_t\|^2, \nonumber \\
\sum_{i=1}^m\mathbb{E}_{t+1}\|w^i_{t+1}-\bar{w}_{t+1}\|^2
& \leq \nu\sum_{i=1}^m\|w^i_t-\bar{w}_t\|^2 + \frac{3\nu^2}{1-\nu}\sum_{i=1}^m\big(L^2\eta^2_t\|\tilde{x}^i_{t+1}-x^i_t\|^2 + \beta_{t+1}^2\|\nabla f^i(x^i_t)-u^i_t\|^2 \nonumber \\
& \quad + \frac{\beta_{t+1}^2}{bn}\sum_{k=1}^n\|z^i_{k,t}-\nabla f^i(x^i_t)\|^2\big).  \nonumber
\end{align}

\end{lemma}

\begin{proof}
For notational simplicity, let $x_t=[(x^1_t)^T,\cdots,(x^m_t)^T]^T\in \mathbb{R}^{md}$, $\tilde{x}_t=[(\tilde{x}^1_t)^T,\cdots,(\tilde{x}^m_t)^T]^T\in \mathbb{R}^{md}$ and $g_t=[(g^1_t)^T,\cdots,(g^m_t)^T]^T\in \mathbb{R}^{md}$ for all $t\geq1$.
By using Assumption \ref{ass:4}, since $W\textbf{1}=\textbf{1}$ and $\tilde{W}=W\otimes I_d$, we have $\tilde{W}(\textbf{1}\otimes \bar{x}_t)=\textbf{1}\otimes\bar{x}_t$.
Meanwhile, we have $\textbf{1}^T(x_t - \textbf{1}\otimes\bar{x})=0$ and $\tilde{W}\textbf{1}=\textbf{1}$.
Thus, we have
\begin{align} \label{eq:C1}
  \|\tilde{W}x_t - \textbf{1}\otimes\bar{x}_t\|^2 = \|\tilde{W}(x_t-\textbf{1}\otimes\bar{x}_t)\|^2 \leq \nu^2\|x_t-\textbf{1}\otimes\bar{x}_t\|^2,
\end{align}
where the above inequality holds by $x_t - \textbf{1}\otimes\bar{x}_t$ is orthogonal to $\textbf{1}$ that is the eigenvector corresponding to the largest eigenvalue of $\tilde{W}$, and $\nu$ denotes the second largest eigenvalue of $\tilde{W}$.

Since $\tilde{x}^i_{t+1} = \sum_{j\in \mathcal{N}_i} W_{i,j}x^j_t - \gamma (A^i_t)^{-1}w^i_t= \sum_{j\in \mathcal{N}_i} W_{i,j}x^j_t - \gamma g^i_t$ for all $i\in [m]$, we have $\tilde{x}_{t+1} = \tilde{W}x_t - \gamma g_t$ and $\bar{\tilde{x}}_{t+1} = \bar{x}_t - \gamma \bar{g}_t$.
Since $x_{t+1}=x_t + \eta_t(\tilde{x}_{t+1}-x_t)$ and $\bar{x}_{t+1}=\bar{x}_t + \eta_t(\bar{\tilde{x}}_{t+1}-\bar{x}_t)$, we have
\begin{align}
  \sum_{i=1}^m\|x^i_{t+1} - \bar{x}_{t+1}\|^2 & = \big\| x_{t+1} - \textbf{1}\otimes\bar{x}_{t+1}\big\|^2 \\
  & = \big\|x_t + \eta_t(\tilde{x}_{t+1}-x_t) - \textbf{1}\otimes\big((\bar{x}_t + \eta_t(\bar{\tilde{x}}_{t+1}-\bar{x}_t)\big) \big\|^2 \nonumber \\
  & \leq (1+\alpha_1)(1-\eta_t)^2\|x_t-\textbf{1}\otimes\bar{x}_t\|^2+(1+\frac{1}{\alpha_1})\eta^2_t
  \|\tilde{x}_{t+1}-\textbf{1}\otimes\bar{\tilde{x}}_{t+1} \|^2 \nonumber \\
  & \mathop{=}^{(i)} (1-\eta_t)\|x_t-\textbf{1}\otimes\bar{x}_t\|^2+\eta_t
  \|\tilde{x}_{t+1}-\textbf{1}\otimes\bar{\tilde{x}}_{t+1} \|^2 \nonumber \\
  & = (1-\eta_t)\|x_t-\textbf{1}\otimes\bar{x}_t\|^2+\eta_t
  \|\tilde{W}x_t - \gamma g_t-\textbf{1}\otimes\big( \bar{x}_t - \gamma \bar{g}_t\big) \|^2 \nonumber \\
  & \leq (1-\eta_t)\|x_t-\textbf{1}\otimes\bar{x}_t\|^2+(1+\alpha_2)\eta_t
  \|\tilde{W}x_t - \textbf{1}\otimes\bar{x}_t\|^2 + (1+\frac{1}{\alpha_2})\eta_t\gamma^2\| g_t-\textbf{1}\otimes\bar{g}_t \|^2 \nonumber \\
  & \mathop{\leq}^{(ii)} (1-\eta_t)\|x_t-\textbf{1}\otimes\bar{x}_t\|^2+\frac{(1+\nu^2)\eta_t}{2}
  \|x_t - \textbf{1}\otimes\bar{x}_t\|^2 + \frac{\eta_t\gamma^2(1+\nu^2)}{1-\nu^2}\| g_t-\textbf{1}\otimes\bar{g}_t \|^2 \nonumber \\
  & \mathop{\leq}^{(iii)} (1-\frac{(1-\nu^2)\eta_t}{2})\|x_t-\textbf{1}\otimes\bar{x}_t\|^2 + \frac{2\eta_t\gamma^2}{1-\nu^2}\| g_t-\textbf{1}\otimes\bar{g}_t \|^2 \nonumber \\
  & = (1-\frac{(1-\nu^2)\eta_t}{2})\sum_{i=1}^m\|x^i_t-\bar{x}_t\|^2 + \frac{2\eta_t\gamma^2}{1-\nu^2}\sum_{i=1}^m\| g^i_t-\bar{g}_t \|^2,
\end{align}
where the above equality $(i)$ is due to $\alpha_1=\frac{\eta_t}{1-\eta_t}$, and the second inequality $(ii)$ holds by $\alpha_2=\frac{1-\nu^2}{2\nu^2}$ and $\|\tilde{W}x_t - \textbf{1}\otimes\bar{x}_t\|^2 \leq
\nu^2\|x_t - \textbf{1}\otimes\bar{x}_t\|^2$, and
 the above inequality $(ii)$ is due to $0<\nu<1$.
Meanwhile, we have
\begin{align}
 \sum_{i=1}^m\|\tilde{x}^i_{t+1}-\bar{x}_t\|^2 & =\|\tilde{x}_{t+1}-\textbf{1}\otimes\bar{x}_t\|^2 \nonumber \\
 & = \|\tilde{W}x_t - \gamma g_t - \textbf{1}\otimes\bar{x}_t\|^2 \nonumber \\
 & \leq (1+\alpha_2)\nu^2\|x_t -\textbf{1}\otimes\bar{x}_t\|^2 + (1+\frac{1}{\alpha_2})\gamma^2\|g_t\|^2
 \nonumber \\
 & \mathop{=}^{(i)} \frac{1+\nu^2}{2}\sum_{i=1}^m\|x^i_t -\bar{x}_t\|^2 + \frac{1+\nu^2}{1-\nu^2}\gamma^2\sum_{i=1}^m\|g^i_t\|^2,
\end{align}
where the last equality $(i)$ holds by $\alpha_2=\frac{1-\nu^2}{2\nu^2}$.
Then we have
\begin{align}
 \sum_{i=1}^m\|\tilde{x}^i_{t+1}-x^i_t\|^2 & =\|\tilde{x}_{t+1}-x_t\|^2 \nonumber \\
 & = \|\tilde{x}_{t+1}- \textbf{1}\otimes\bar{x}_t + \textbf{1}\otimes\bar{x}_t - x_t\|^2 \nonumber \\
 & \leq 2\|\tilde{x}_{t+1} - \textbf{1}\otimes\bar{x}_t\|^2 + 2\|x_t - \textbf{1}\otimes\bar{x}_t\|^2 \nonumber \\
 & = (3+\nu^2)\sum_{i=1}^m\|x^i_t -\bar{x}_t\|^2 + \frac{2(1+\nu^2)}{1-\nu^2}\gamma^2\sum_{i=1}^m\|g^i_t\|^2.
\end{align}

Let $w_t=[(w^1_t)^T,(w^2_t)^T,\cdots,(w^m_t)^T]^T$,
$u_t=[(u^1_t)^T,(u^2_t)^T,\cdots,(u^m_t)^T]^T$ and $\bar{w}_t = \frac{1}{m}\sum_{i=1}^mw^i_t$ and
$\bar{u}_t = \frac{1}{m}\sum_{i=1}^mu^i_t$. Then we have for any $t\geq 1$,
\begin{align}
  w_{t+1} = \tilde{W}\big(w_t + u_{t+1} - u_t\big). \nonumber
\end{align}
According to the above proof of Lemma~\ref{lem:B1}, we have $\bar{w}_{t+1}=\bar{w}_t + \bar{u}_{t+1} - \bar{u}_t$ for all $t\geq1$.
Thus we have
\begin{align} \label{eq:C3}
     \sum_{i=1}^m\|w^i_{t+1}-\bar{w}_{t+1}\|^2 & =
     \|w_{t+1}-\textbf{1}\otimes\bar{w}_{t+1}\|^2 \nonumber \\
    & = \big\| \tilde{W}\big(w_t + u_{t+1} - u_t\big) -\textbf{1}\otimes\big(\bar{w}_t + \bar{u}_{t+1} - \bar{u}_t \big)\big\|^2 \nonumber \\
    & \leq (1+c)\|\tilde{W}w_t-\textbf{1}\otimes\bar{w}_t\|^2 + (1+\frac{1}{c})\big\|\tilde{W}\big( u_{t+1} - u_t\big)-\textbf{1}\otimes\big(\bar{u}_{t+1} - \bar{u}_t\big)\big\|^2 \nonumber \\
    & \leq (1+c)\nu^2\|w_t-\textbf{1}\otimes\bar{w}_t\|^2 + (1+\frac{1}{c})\nu^2\big\|u_{t+1} - u_t-\textbf{1}\otimes\big(\bar{u}_{t+1} - \bar{u}_t\big)\big\|^2 \nonumber \\
    & \leq (1+c)\nu^2\|w_t-\textbf{1}\otimes\bar{w}_t\|^2 + (1+\frac{1}{c})\nu^2\big\|u_{t+1} - u_t\big\|^2,
\end{align}
where the last inequality holds by Lemma~\ref{lem:A2}.

Since $u^i_{t+1} = \frac{1}{b}\sum_{k\in \mathcal{I}^i_{t+1}}\big(\nabla f^i_k(x_{t+1})-\nabla f^i_k(x_t)\big) + (1-\beta_{t+1}) u^i_t + \beta_{t+1}\Big(\frac{1}{b}\sum_{k\in \mathcal{I}^i_{t+1}}\big(\nabla f^i_k(x_t)-z^i_{k,t}\big) + \frac{1}{n}\sum_{j=1}^nz^i_{j,t}\Big)$ for any $i\in [m]$ and $t\geq 1$, we have
\begin{align} \label{eq:C4}
 & \mathbb{E}_{t+1}\big\|u_{t+1} - u_t\big\|^2 \nonumber \\
 & = \sum_{i=1}^m\mathbb{E}_{t+1}\|u^i_{t+1} - u^i_t\|^2 \nonumber \\
 &= \sum_{i=1}^m\mathbb{E}_{t+1}\big\|\frac{1}{b}\sum_{k\in \mathcal{I}^i_{t+1}}\big(\nabla f^i_k(x^i_{t+1})-\nabla f^i_k(x^i_t)\big) -\beta_{t+1} u^i_t + \beta_{t+1}\Big(\frac{1}{b}\sum_{k\in \mathcal{I}^i_{t+1}}\big(\nabla f^i_k(x^i_t)-z^i_{k,t}\big) + \frac{1}{n}\sum_{j=1}^nz^i_{j,t}\Big) \big\|^2 \nonumber \\
 &= \sum_{i=1}^m\mathbb{E}_{t+1}\big\|\frac{1}{b}\sum_{k\in \mathcal{I}^i_{t+1}}\big(\nabla f^i_k(x^i_{t+1})-\nabla f^i_k(x^i_t)\big) -\beta_{t+1} u^i_t + \beta_{t+1}\nabla f^i(x^i_t)  \nonumber \\
 & \quad + \beta_{t+1}\Big(\frac{1}{b}\sum_{k\in \mathcal{I}^i_{t+1}}\big(\nabla f^i_k(x^i_t)-z^i_{k,t}\big) + \frac{1}{n}\sum_{j=1}^nz^i_{j,t}- \nabla f^i(x^i_t) \Big) \big\|^2 \nonumber \\
 & \leq 3L^2\sum_{i=1}^m\|x^i_{t+1}-x^i_t\|^2 + 3\beta_{t+1}^2\sum_{i=1}^m\|\nabla f^i(x^i_t)-u^i_t\|^2 + \frac{3\beta_{t+1}^2}{bn}\sum_{i=1}^m\sum_{k=1}^n\|z^i_{k,t}-\nabla f^i(x^i_t)\|^2 \nonumber \\
 & = 3L^2\eta^2_t\sum_{i=1}^m\|\tilde{x}^i_{t+1}-x^i_t\|^2 + 3\beta_{t+1}^2\sum_{i=1}^m\|\nabla f^i(x^i_t)-u^i_t\|^2 + \frac{3\beta_{t+1}^2}{bn}\sum_{i=1}^m\sum_{k=1}^n\|z^i_{k,t}-\nabla f^i(x^i_t)\|^2,
\end{align}
where the second last inequality holds by Assumption~\ref{ass:1} and the last equality is due to $x^i_{t+1}= x^i_t \eta_t(\tilde{x}^i_{t+1}-x^i_t)$.

Plugging the above inequalities (\ref{eq:C4}) into (\ref{eq:C3}), we have
\begin{align}
    \sum_{i=1}^m\|w^i_{t+1}-\bar{w}_{t+1}\|^2
    & \leq (1+c)\nu^2\|w_t-\textbf{1}\otimes\bar{w}_t\|^2 + (1+\frac{1}{c})\nu^2\big\|u_{t+1} - u_t\big\|^2 \nonumber \\
    & \leq (1+c)\nu^2\|w_t-\textbf{1}\otimes\bar{w}_t\|^2 + 3(1+\frac{1}{c})\nu^2\sum_{i=1}^m\big(L^2\eta^2_t\|\tilde{x}^i_{t+1}-x^i_t\|^2 + \beta_{t+1}^2\|\nabla f^i(x^i_t)-u^i_t\|^2 \nonumber \\
    & \quad + \frac{\beta_{t+1}^2}{bn}\sum_{k=1}^n\|z^i_{k,t}-\nabla f^i(x^i_t)\|^2\big).
\end{align}
Let $c=\frac{1}{\nu}-1$, we have
\begin{align}
   \sum_{i=1}^m\mathbb{E}_{t+1}\|w^i_{t+1}-\bar{w}_{t+1}\|^2
    & \leq \nu\sum_{i=1}^m\|w^i_t-\bar{w}_t\|^2 + \frac{3\nu^2}{1-\nu}\sum_{i=1}^m\big(L^2\eta^2_t\|\tilde{x}^i_{t+1}-x^i_t\|^2 + \beta_{t+1}^2\|\nabla f^i(x^i_t)-u^i_t\|^2 \nonumber \\
    & \quad + \frac{\beta_{t+1}^2}{bn}\sum_{k=1}^n\|z^i_{k,t}-\nabla f^i(x^i_t)\|^2\big).
\end{align}

\end{proof}

\begin{theorem}  \label{th:A2}
(Restatement of Theorem~\ref{th:2})
 Suppose the sequences $\big\{\{x^i_t\}_{i=1}^m\big\}_{t=1}^T$ be generated from Algorithm~\ref{alg:2}.
 Under the above Assumptions~\ref{ass:1}-\ref{ass:5}, and let $\eta_t=\eta$, $0<\beta_t\leq1$ for all $t\geq 0$, $\gamma\leq \min\big(\frac{\rho(1-\nu^2)}{48\theta_t},\frac{3\rho(1-\nu^2)\theta_t}{58L^2}\big)$, $\eta\leq \min\big(\frac{\rho\sqrt{1-\nu^2}}{2L\gamma\sqrt{6(1+\nu^2)}\sqrt{H_t}},\frac{\sqrt{\rho(1-\nu^2)\theta_t}}{2 L\sqrt{\gamma(3+\nu^2)}\sqrt{H_t}}\big)$ with $H_t=\frac{9}{b\beta_t}+\frac{6\nu^2}{b(1-\nu)^2}+\frac{4n^2\beta^2_t}{b^3}\big(\frac{9}{\beta_t}+
\frac{9\nu^2}{(1-\nu)^2}\big)+\frac{3\nu^2}{(1-\nu)^2}$ for all $t\geq 0$, we have
\begin{align}
 \frac{1}{T}\sum_{t=1}^T \mathbb{E}\|\nabla F(\bar{x}_t)\|
  \leq \frac{1}{T}\sum_{t=1}^T\frac{1}{m}\sum_{i=1}^m\mathbb{E}[\|\nabla F(x^i_t)\| + L\|\bar{x}_t-x^i_t\|] \leq \frac{6\sqrt{G}}{\sqrt{T}}\sqrt{\frac{1}{T}\sum_{t=1}^T\frac{1}{m}\sum_{i=1}^m\mathbb{E}\|A^i_t\|^2},
\end{align}
where $G=\frac{F(\bar{x}_1)-F^*}{\rho\gamma\eta}+\big(\frac{18\beta_0}{\rho^2}+\frac{18\beta^2_0\nu^2}{\rho^2(1-\nu)^2}
 +\frac{3\nu^2}{\rho^2(1-\nu)^2}
 +\frac{9}{2\rho^2\beta_0}-\frac{9}{2\rho^2}\big)
 \frac{1}{m}\sum_{i=1}^m\frac{1}{n}\sum_{k=1}^n\|\nabla f^i_k(x^i_{0})\|^2$ is independent on
$T$, $b$ and $n$.
\end{theorem}

\begin{proof}
Without loss of generality, let $\eta=\eta_1=\cdots=\eta_T$.
According to Lemma \ref{lem:B2}, we have
\begin{align} \label{eq:G1}
 F(\bar{x}_{t+1}) \leq F(\bar{x}_t)+\frac{2\gamma\eta}{\rho}\|\nabla F(\bar{x}_t)-\bar{u}_t\|^2+\frac{\gamma\eta}{2\rho}\frac{1}{m}\sum_{i=1}^m\|w^i_t -\bar{w}_t\|^2-\frac{\rho\gamma\eta}{4} \frac{1}{m}\sum_{i=1}^m\|g^i_t\|^2.
\end{align}

According to the Lemma~\ref{lem:C1},
we have
 \begin{align} \label{eq:G2}
  \mathbb{E}_t\|\bar{u}_t - \overline{\nabla f(x_t)}\|^2 & \leq (1-\beta_t)\|\bar{u}_{t-1}-\overline{\nabla f(x_{t-1})}\|^2 + \frac{2L^2\eta^2}{bm}\sum_{i=1}^m\|\tilde{x}^i_t-x^i_{t-1}\|^2 \nonumber \\
 & \quad + \frac{2\beta_t^2}{bm}\sum_{i=1}^m\frac{1}{n}\sum_{k=1}^n\|\nabla f^i_k(x^i_{t-1}) -z^i_{k,t-1} \|^2,
 \end{align}
and
 \begin{align} \label{eq:G3}
 \frac{1}{m}\sum_{i=1}^m \mathbb{E}_t\|u^i_t - \nabla f^i(x^i_t)\|^2
 & \leq  (1-\beta_t)\frac{1}{m}\sum_{i=1}^m\|u^i_{t-1}-\nabla f^i(x^i_{t-1})\|^2
 + \frac{2L^2\eta^2}{bm}\sum_{i=1}^m \|\tilde{x}^i_t-x^i_{t-1}\|^2 \nonumber \\
 & \quad + \frac{2\beta^2_t}{b}\frac{1}{m}\sum_{i=1}^m\frac{1}{n}\sum_{k=1}^n\|\nabla f^i_k(x^i_{t-1})-z^i_{k,t-1}\|^2.
 \end{align}

According to Lemma \ref{lem:C2}, we have for any $i\in [m]$
\begin{align} \label{eq:G4}
 \mathbb{E}_t\big[\frac{1}{n}\sum_{k=1}^n\|\nabla f^i_k(x_t)-z^i_{k,t}\|^2\big]
 & \leq (1-\frac{b}{2n})\frac{1}{n}\sum_{k=1}^n\|\nabla f^i_k(x^i_{t-1})-z^i_{k,t-1}\|^2 + (\frac{2n}{b}-\frac{b}{n}-1)L^2\eta^2\|\tilde{x}^i_t-x^i_{t-1}\|^2 \nonumber \\
 & \leq (1-\frac{b}{2n})\frac{1}{n}\sum_{k=1}^n\|\nabla f^i_k(x^i_{t-1})-z^i_{k,t-1}\|^2 + \frac{2n}{b}L^2\eta^2\|\tilde{x}^i_t-x^i_{t-1}\|^2.
\end{align}

According to Lemma \ref{lem:C3}, we have
\begin{align} \label{eq:G5}
\frac{1}{m}\sum_{i=1}^m\mathbb{E}_t\|w^i_t-\bar{w}_t\|^2
& \leq \nu\frac{1}{m}\sum_{i=1}^m\|w^i_{t-1}-\bar{w}_{t-1}\|^2 + \frac{3\nu^2}{1-\nu}\frac{1}{m}\sum_{i=1}^m\Big(L^2\eta^2\|\tilde{x}^i_t-x^i_{t-1}\|^2 + \beta_t^2\|\nabla f^i(x^i_{t-1})-u^i_{t-1}\|^2 \nonumber \\
& \quad + \frac{\beta_t^2}{bn}\sum_{k=1}^n\|z^i_{k,t-1}-\nabla f^i(x^i_{t-1})\|^2\Big).
\end{align}
Meanwhile, we also have
\begin{align} \label{eq:G6}
\frac{1}{m}\sum_{i=1}^m\|x^i_t - \bar{x}_t\|^2 & \leq (1-\frac{(1-\nu^2)\eta}{2})\frac{1}{m}\sum_{i=1}^m\|x^i_{t-1}-\bar{x}_{t-1}\|^2
+ \frac{2\eta\gamma^2}{1-\nu^2}\frac{1}{m}\sum_{i=1}^m\| g^i_{t-1}-\bar{g}_{t-1} \|^2 \nonumber \\
& \leq (1-\frac{(1-\nu^2)\eta}{2})\frac{1}{m}\sum_{i=1}^m\|x^i_{t-1}-\bar{x}_{t-1}\|^2
+ \frac{2\eta\gamma^2}{1-\nu^2}\frac{1}{m}\sum_{i=1}^m(\| g^i_{t-1}\|^2 + \|\bar{g}_{t-1} \|^2) \nonumber \\
& \leq (1-\frac{(1-\nu^2)\eta}{2})\frac{1}{m}\sum_{i=1}^m\|x^i_{t-1}-\bar{x}_{t-1}\|^2
+ \frac{4\eta\gamma^2}{1-\nu^2}\frac{1}{m}\sum_{i=1}^m\|g^i_{t-1}\|^2.
\end{align}
and
\begin{align} \label{eq:G7}
\frac{1}{m}\sum_{i=1}^m\|\tilde{x}^i_t-x^i_{t-1}\|^2
 & \leq (3+\nu^2)\frac{1}{m}\sum_{i=1}^m\|x^i_{t-1} -\bar{x}_{t-1}\|^2 + \frac{2(1+\nu^2)}{1-\nu^2}\gamma^2
 \frac{1}{m}\sum_{i=1}^m\|g^i_{t-1}\|^2.
\end{align}
Next considering the term $\|\bar{u}_t-\nabla F(\bar{x}_t)\|^2$, we have
\begin{align}
\|\bar{u}_t-\nabla F(\bar{x}_t)\|^2 & = \|\bar{u}_t- \overline{\nabla f(x_t)} + \overline{\nabla f(x_t)} - \nabla F(\bar{x}_t)\|^2 \nonumber \\
& \leq 2\|\bar{u}_t- \overline{\nabla f(x_t)}\|^2 + 2\|\overline{\nabla f(x_t)} - \nabla F(\bar{x}_t)\|^2 \nonumber \\
& \leq 2\|\bar{u}_t- \overline{\nabla f(x_t)}\|^2 + 2\|\frac{1}{m}\sum_{i=1}^m\nabla f^i(x^i_t) - \frac{1}{m}\sum_{i=1}^m\nabla f^i(\bar{x}_t)\|^2 \nonumber \\
& \leq 2\|\bar{u}_t- \overline{\nabla f(x_t)}\|^2 + \frac{2L^2}{m}\sum_{i=1}^m\|x^i_t - \bar{x}_t\|^2, \nonumber
\end{align}
where the last inequality is due to Assumption~\ref{ass:1}. Then we can obtain
\begin{align} \label{eq:G8}
 - \|\bar{u}_t- \overline{\nabla f(x_t)}\|^2 \leq -\frac{1}{2}\|\bar{u}_t-\nabla F(\bar{x}_t)\|^2 + \frac{L^2}{m}\sum_{i=1}^m\|x^i_t - \bar{x}_t\|^2.
\end{align}
Since $\bar{u}_t=\bar{w}_t$ for all $t\geq1$, we have
\begin{align}
 \frac{1}{m}\sum_{i=1}^m\|w^i_t-\nabla F(x^i_t)\|^2 & = \frac{1}{m}\sum_{i=1}^m\|w^i_t-\bar{w}_t+\bar{u}_t-\nabla F(\bar{x}_t)+\nabla F(\bar{x}_t)-\nabla F(x^i_t)\|^2 \nonumber \\
 & \leq 3\frac{1}{m}\sum_{i=1}^m\|w^i_t-\bar{w}_t\|^2 + 3\|\bar{u}_t-\nabla F(\bar{x}_t)\|^2+3\frac{1}{m}\sum_{i=1}^m\|\nabla F(\bar{x}_t)-\nabla F(x^i_t)\|^2 \nonumber \\
 & \leq 3\frac{1}{m}\sum_{i=1}^m\|w^i_t-\bar{w}_t\|^2 + 3\|\bar{u}_t-\nabla F(\bar{x}_t)\|^2+3L^2\frac{1}{m}\sum_{i=1}^m\|x^i_t-\bar{x}_t\|^2. \nonumber
\end{align}
Then we have
\begin{align} \label{eq:G9}
 -\|\bar{u}_t-\nabla F(\bar{x}_t)\|^2
 & \leq -\frac{1}{3m}\sum_{i=1}^m\|w^i_t-\nabla F(x^i_t)\|^2 + \frac{1}{m}\sum_{i=1}^m\|w^i_t-\bar{w}_t\|^2 +\frac{L^2}{m}\sum_{i=1}^m\|x^i_t-\bar{x}_t\|^2.
\end{align}

We define a useful Lyapunov function (i.e., potential function), for any $t\geq 1$
\begin{align} \label{eq:G10}
\Phi_t & = \mathbb{E}_t\big[ F(\bar{x}_t) + (\lambda_{t-1}-\frac{9\gamma\eta}{2\rho})\|\bar{u}_{t-1} - \overline{\nabla f(x_{t-1})}\|^2 + \chi_{t-1}\frac{1}{m}\sum_{i=1}^m \|u^i_{t-1} - \nabla f^i(x^i_{t-1})\|^2 \\
& \qquad + \alpha_{t-1}\frac{1}{m}\sum_{i=1}^m\frac{1}{n}\sum_{k=1}^n\|\nabla f^i_k(x^i_{t-1})-z^i_{k,t-1}\|^2 + (\theta_{t-1}-\frac{19\gamma\eta L^2}{4\rho})\frac{1}{m}\sum_{i=1}^m\|x^i_{t-1}-\bar{x}_{t-1}\|^2 \nonumber \\
& \qquad +(\vartheta_{t-1}-\frac{3\gamma\eta}{4\rho})\frac{1}{m}\sum_{i=1}^m\|w^i_{t-1}-\bar{w}_{t-1}\|^2 + \frac{\rho\gamma\eta}{6} \frac{1}{m}\sum_{i=1}^m\|g^i_{t-1}\|^2\big], \nonumber
\end{align}
where $\alpha_{t-1}\geq 0$, $\chi_{t-1}\geq 0$, $\lambda_{t-1}\geq \frac{9\gamma\eta}{2\rho}$,
$\theta_{t-1}\geq \frac{29\gamma\eta L^2}{6\rho}$ and $\vartheta_{t-1}\geq \frac{3\gamma\eta}{4\rho}$ for all $t\geq1$.
Then we have
\begin{align} \label{eq:G11}
 \Phi_{t+1} & = \mathbb{E}_{t+1}\Big[ F(\bar{x}_{t+1}) + (\lambda_t-\frac{9\gamma\eta}{2\rho})\|\bar{u}_t - \overline{\nabla f(x_t)}\|^2 + \chi_t\frac{1}{m}\sum_{i=1}^m \|u^i_t - \nabla f^i(x^i_t)\|^2 \nonumber \\
 & \qquad  + \alpha_t\frac{1}{m}\sum_{i=1}^m\frac{1}{n}\sum_{k=1}^n\|\nabla f^i_k(x^i_t)-z^i_{k,t}\|^2 + (\theta_t-\frac{29\gamma\eta L^2}{6\rho})\frac{1}{m}\sum_{i=1}^m\|x^i_t-\bar{x}_t\|^2 \nonumber \\
 & \qquad+(\vartheta_t-\frac{3\gamma\eta}{4\rho})\frac{1}{m}\sum_{i=1}^m\|w^i_t-\bar{w}_t\|^2 + \frac{\rho\gamma\eta}{6} \frac{1}{m}\sum_{i=1}^m\|g^i_t\|^2 \Big] \nonumber \\
 & \mathop{\leq}^{(i)} \mathbb{E}_{t+1}\Big[ F(\bar{x}_t) -\frac{\gamma\eta}{4\rho}\|\bar{u}_t - \nabla F(\bar{x}_t)\|^2 - \frac{\rho\gamma\eta}{12} \frac{1}{m}\sum_{i=1}^m\|g^i_t\|^2 + \frac{9\gamma\eta L^2}{2\rho}\frac{1}{m}\sum_{i=1}^m\|x^i_t - \bar{x}_t\|^2 + \lambda_t\|\bar{u}_t-\overline{\nabla f(x_t)}\|^2 \nonumber \\
 & \qquad  + \chi_t\frac{1}{m}\sum_{i=1}^m \|u^i_t - \nabla f^i(x^i_t)\|^2 + \alpha_t(1-\frac{b}{2n})\frac{1}{m}\sum_{i=1}^m\frac{1}{n}\sum_{k=1}^n\|\nabla f^i_k(x^i_{t-1})-z^i_{k,t-1}\|^2\nonumber \\
 & \qquad + \frac{2\alpha_t n}{b}L^2\eta^2\frac{1}{m}\sum_{i=1}^m\|\tilde{x}^i_t-x^i_{t-1}\|^2 + (\theta_t-\frac{29\gamma\eta L^2}{6\rho})\frac{1}{m}\sum_{i=1}^m\|x^i_t-\bar{x}_t\|^2 \nonumber \\
 & \qquad   +(\vartheta_t-\frac{\gamma\eta}{4\rho})\frac{1}{m}\sum_{i=1}^m\|w^i_t-\bar{w}_t\|^2 \Big] \nonumber \\
 & \mathop{\leq}^{(ii)} \mathbb{E}_{t+1}\Big[F(\bar{x}_t) -\frac{\gamma\eta}{12\rho}\frac{1}{m}\sum_{i=1}^m\|w^i_t-\nabla F(x^i_t)\|^2 - \frac{\rho\gamma\eta}{12} \frac{1}{m}\sum_{i=1}^m\|g^i_t\|^2 -\frac{\gamma\eta L^2}{12\rho}\frac{1}{m}\sum_{i=1}^m\|x^i_t-\bar{x}_t\|^2 \nonumber \\
 & \quad + \lambda_t(1-\beta_t)\|\bar{u}_{t-1}-\overline{\nabla f(x_{t-1})}\|^2 + \frac{2\lambda_tL^2\eta^2}{bm}\sum_{i=1}^m\|\tilde{x}^i_t-x^i_{t-1}\|^2 + \frac{2\lambda_t\beta_t^2}{bm}\sum_{i=1}^m\frac{1}{n}\sum_{k=1}^n\|\nabla f^i_k(x^i_{t-1}) -z^i_{k,t-1} \|^2 \nonumber \\
 & \quad + \chi_t(1-\beta_t)\frac{1}{m}\sum_{i=1}^m\|u^i_{t-1}-\nabla f^i(x^i_{t-1})\|^2
 + \frac{2\chi_tL^2\eta^2}{bm}\sum_{i=1}^m\|\tilde{x}^i_t-x^i_{t-1}\|^2 \nonumber \\
 & \quad + \frac{2\chi_t\beta^2_t}{b}\frac{1}{m}\sum_{i=1}^m\frac{1}{n}\sum_{k=1}^n\|\nabla f^i_k(x^i_{t-1})-z^i_{k,t-1}\|^2  + \alpha_t(1-\frac{b}{2n})\frac{1}{m}\sum_{i=1}^m\frac{1}{n}\sum_{k=1}^n\|\nabla f^i_k(x^i_{t-1})-z^i_{k,t-1}\|^2 \nonumber \\
 & \quad + \frac{2\alpha_t n}{b}L^2\eta^2\frac{1}{m}\sum_{i=1}^m\|\tilde{x}^i_t-x^i_{t-1}\|^2 + \theta_t(1-\frac{(1-\nu^2)\eta}{2})\frac{1}{m}\sum_{i=1}^m\|x^i_{t-1}-\bar{x}_{t-1}\|^2
+ \theta_t\frac{4\eta\gamma^2}{1-\nu^2}\frac{1}{m}\sum_{i=1}^m\|g^i_{t-1}\|^2 \nonumber \\
& \quad + \vartheta_t\nu\frac{1}{m}\sum_{i=1}^m\|w^i_{t-1}-\bar{w}_{t-1}\|^2 + \frac{3\vartheta_t\nu^2}{1-\nu}\frac{1}{m}\sum_{i=1}^m\Big(L^2\eta^2\|\tilde{x}^i_t-x^i_{t-1}\|^2 + \beta_t^2\|\nabla f^i(x^i_{t-1})-u^i_{t-1}\|^2 \nonumber \\
& \quad + \frac{\beta_t^2}{bn}\sum_{k=1}^n\|z^i_{k,t-1}-\nabla f^i(x^i_{t-1})\|^2\Big) \Big] ,
\end{align}
where the inequality (i) is due to the above inequalities (\ref{eq:G1}), (\ref{eq:G4}) and (\ref{eq:G8}); and
the inequality (ii) holds by the above inequalities (\ref{eq:G2}), (\ref{eq:G3}), (\ref{eq:G5}) and (\ref{eq:G9}).

Then we have
\begin{align} \label{eq:G12}
\Phi_{t+1} & \leq  \mathbb{E}_{t+1}\Big[ F(\bar{x}_t) -\frac{\gamma\eta}{12\rho}\frac{1}{m}\sum_{i=1}^m\|w^i_t-\nabla F(x^i_t)\|^2 - \frac{\rho\gamma\eta}{12} \frac{1}{m}\sum_{i=1}^m\|g^i_t\|^2  -\frac{\gamma\eta L^2}{12\rho}\frac{1}{m}\sum_{i=1}^m\|x^i_t-\bar{x}_t\|^2  \nonumber \\
 & \quad + \lambda_t(1-\beta_t)\|\bar{u}_{t-1}-\overline{\nabla f(x_{t-1})}\|^2  + \big(\chi_t -\chi_t\beta_t + \frac{3\vartheta_t\beta^2_t\nu^2}{1-\nu} \big)\frac{1}{m}\sum_{i=1}^m\|u^i_{t-1}-\nabla f^i(x^i_{t-1})\|^2 \nonumber \\
 & \quad + L^2\eta^2\big( \frac{2\lambda_t}{b} + \frac{2\chi_t}{b}+\frac{2\alpha_tn}{b}+ \frac{3\vartheta_t\nu^2}{1-\nu}\big)\frac{1}{m}\sum_{i=1}^m\|\tilde{x}^i_t-x^i_{t-1}\|^2  \nonumber \\
 & \quad  + \big(\alpha_t-\frac{\alpha_tb}{2n}+ \frac{2\lambda_t\beta_t^2}{b}+\frac{2\chi_t\beta_t^2}{b}+\frac{3\beta_t^2\vartheta_t\nu^2}{b(1-\nu)}\big)\frac{1}{m}\sum_{i=1}^m\frac{1}{n}\sum_{k=1}^n\|\nabla f^i_k(x^i_{t-1})-z^i_{k,t-1}\|^2 \nonumber \\
 & \quad + \theta_t(1-\frac{(1-\nu^2)\eta}{2})\frac{1}{m}\sum_{i=1}^m\|x^i_{t-1}-\bar{x}_{t-1}\|^2
+ \theta_t\frac{4\eta\gamma^2}{1-\nu^2}\frac{1}{m}\sum_{i=1}^m\|g^i_{t-1}\|^2  + \vartheta_t\nu\frac{1}{m}\sum_{i=1}^m\|w^i_{t-1}-\bar{w}_{t-1}\|^2  \Big] \nonumber \\
& \mathop{\leq}^{(i)} \mathbb{E}_{t+1}\Big[ F(\bar{x}_t) -\frac{\gamma\eta}{12\rho}\frac{1}{m}\sum_{i=1}^m\|w^i_t-\nabla F(x^i_t)\|^2 - \frac{\rho\gamma\eta}{12} \frac{1}{m}\sum_{i=1}^m\|g^i_t\|^2  -\frac{\gamma\eta L^2}{12\rho}\frac{1}{m}\sum_{i=1}^m\|x^i_t-\bar{x}_t\|^2  \nonumber \\
 & \quad + \lambda_t(1-\beta_t)\|\bar{u}_{t-1}-\overline{\nabla f(x_{t-1})}\|^2  + \big(\chi_t -\chi_t\beta_t + \frac{3\vartheta_t\beta^2_t\nu^2}{1-\nu} \big)\frac{1}{m}\sum_{i=1}^m\|u^i_{t-1}-\nabla f^i(x^i_{t-1})\|^2 \nonumber \\
 & \quad + L^2\eta^2\big( \frac{2\lambda_t}{b} + \frac{2\chi_t}{b}+\frac{2\alpha_tn}{b}+ \frac{3\vartheta_t\nu^2}{1-\nu}\big)\Big((3+\nu^2)\frac{1}{m}\sum_{i=1}^m\|x^i_{t-1} -\bar{x}_{t-1}\|^2 + \frac{2(1+\nu^2)}{1-\nu^2}\gamma^2
 \frac{1}{m}\sum_{i=1}^m\|g^i_{t-1}\|^2\Big)  \nonumber \\
 & \quad  + \big(\alpha_t-\frac{\alpha_tb}{2n}+ \frac{2\lambda_t\beta_t^2}{b}+\frac{2\chi_t\beta_t^2}{b}+\frac{3\beta_t^2\vartheta_t\nu^2}{b(1-\nu)}\big)\frac{1}{m}\sum_{i=1}^m\frac{1}{n}\sum_{k=1}^n\|\nabla f^i_k(x^i_{t-1})-z^i_{k,t-1}\|^2 \nonumber \\
 & \quad + \theta_t(1-\frac{(1-\nu^2)\eta}{2})\frac{1}{m}\sum_{i=1}^m\|x^i_{t-1}-\bar{x}_{t-1}\|^2
+ \theta_t\frac{4\eta\gamma^2}{1-\nu^2}\frac{1}{m}\sum_{i=1}^m\|g^i_{t-1}\|^2  + \vartheta_t\nu\frac{1}{m}\sum_{i=1}^m\|w^i_{t-1}-\bar{w}_{t-1}\|^2  \Big],
\end{align}
where the inequality (i) is due to the above inequality (\ref{eq:G7}).

Since $0<\beta_t < 1$ for all $t\geq1$, $\lambda_t = \frac{9\gamma\eta}{2\rho\beta_t}\geq \frac{9\gamma\eta}{2\rho}$ and $\lambda_t \leq \lambda_{t-1}$, then we have $\lambda_t(1-\beta_t)\leq \lambda_{t-1} - \frac{9\gamma\eta}{2\rho}$. Let $\chi_t =\frac{3\vartheta_t\nu^2}{1-\nu}\geq \frac{3\vartheta_t\beta_t\nu^2}{1-\nu}$ and $\chi_t\leq \chi_{t-1}$, we have $\chi_t -\chi_t\beta_t + \frac{3\vartheta_t\beta^2_t\nu^2}{1-\nu} \leq \chi_{t-1}$. Let $\alpha_t=
\frac{2n\beta^2_t}{b^2}(2\lambda_t+2\chi_t+\frac{3\vartheta_t\nu^2}{1-\nu})
=\frac{2n\beta^2_t}{b^2}(2\lambda_t+\frac{9\vartheta_t\nu^2}{1-\nu})$ and $\alpha_t \leq \alpha_{t-1}$,
then we have
$\alpha_t-\frac{\alpha_tb}{2n}+ \frac{2\lambda_t\beta_t^2}{b}+\frac{2\chi_t\beta_t^2}{b}+\frac{3\beta_t^2\vartheta_t\nu^2}{b(1-\nu)} \leq \alpha_{t-1}$. Let $\vartheta_t=\frac{\gamma\eta}{\rho(1-\nu)}$ for all $t\geq1$, since $0<\nu<1$, we have $\vartheta_t\nu =\vartheta_t - (1-\nu)\vartheta_t \leq \vartheta_{t-1}-\frac{3\gamma\eta}{4\rho}$.
Meanwhile, let $\theta_t\leq \theta_{t-1}$ for all $t\geq1$, $\gamma\leq \min\big(\frac{\rho(1-\nu^2)}{48\theta_t},\frac{3\rho(1-\nu^2)\theta_t}{58L^2}\big)$, $\eta\leq \min\big(\frac{\rho\sqrt{1-\nu^2}}{2L\gamma\sqrt{6(1+\nu^2)}\sqrt{H_t}},\frac{\sqrt{\rho(1-\nu^2)\theta_t}}{2 L\sqrt{\gamma(3+\nu^2)}\sqrt{H_t}}\big)$ with $H_t=\frac{9}{b\beta_t}+\frac{6\nu^2}{b(1-\nu)^2}+\frac{4n^2\beta^2_t}{b^3}\big(\frac{9}{\beta_t}+
\frac{9\nu^2}{(1-\nu)^2}\big)+\frac{3\nu^2}{(1-\nu)^2}$ for all $t\geq1$, we can obtain
\begin{align}
& \theta_t(1-\frac{(1-\nu^2)\eta}{2}) + L^2\eta^2(3+\nu^2) \big( \frac{2\lambda_t}{b} + \frac{2\chi_t}{b}+\frac{2\alpha_tn}{b}+ \frac{3\vartheta_t\nu^2}{1-\nu}\big) \leq \theta_{t-1} - \frac{29\gamma\eta L^2}{6\rho}, \\
& \theta_t\frac{4\eta\gamma^2}{1-\nu^2} +\frac{2(1+\nu^2)}{1-\nu^2}L^2\eta^2\gamma^2\big( \frac{2\lambda_t}{b} + \frac{2\chi_t}{b}+\frac{2\alpha_tn}{b}+ \frac{3\vartheta_t\nu^2}{1-\nu}\big) \leq \frac{\rho\gamma\eta}{6}.
\end{align}

Based on the choice of these parameters and the above inequality (\ref{eq:G12}), we have
\begin{align} \label{eq:G13}
  \Phi_{t+1} & \leq \mathbb{E}_{t+1}\Big[ F(\bar{x}_t) -\frac{\gamma\eta}{12\rho}\frac{1}{m}\sum_{i=1}^m\|w^i_t-\nabla F(x^i_t)\|^2 - \frac{\rho\gamma\eta}{12} \frac{1}{m}\sum_{i=1}^m\|g^i_t\|^2  -\frac{\gamma\eta L^2}{12\rho}\frac{1}{m}\sum_{i=1}^m\|x^i_t-\bar{x}_t\|^2  \nonumber \\
 & \quad + (\lambda_{t-1}-\frac{9\gamma\eta}{2\rho})\|\bar{u}_{t-1}-\overline{\nabla f(x_{t-1})}\|^2  + \chi_{t-1}\frac{1}{m}\sum_{i=1}^m\|u^i_{t-1}-\nabla f^i(x^i_{t-1})\|^2 \nonumber \\
 & \quad  + \alpha_{t-1}\frac{1}{m}\sum_{i=1}^m\frac{1}{n}\sum_{k=1}^n\|\nabla f^i_k(x^i_{t-1})-z^i_{k,t-1}\|^2 +  (\theta_{t-1} - \frac{29\gamma\eta L^2}{6\rho})\frac{1}{m}\sum_{i=1}^m\|x^i_{t-1}-\bar{x}_{t-1}\|^2 \nonumber \\
 & \quad  + (\vartheta_{t-1}-\frac{3\gamma\eta}{4\rho})\frac{1}{m}\sum_{i=1}^m\|w^i_{t-1}-\bar{w}_{t-1}\|^2 +
\frac{\rho\gamma\eta}{6}\frac{1}{m}\sum_{i=1}^m\|g^i_{t-1}\|^2  \Big] \nonumber \\
 & = \Phi_t  -\frac{\gamma\eta}{12\rho}\frac{1}{m}\sum_{i=1}^m\|w^i_t-\nabla F(x^i_t)\|^2 - \frac{\rho\gamma\eta}{12} \frac{1}{m}\sum_{i=1}^m\|g^i_t\|^2  -\frac{\gamma\eta L^2}{12\rho}\frac{1}{m}\sum_{i=1}^m\|x^i_t-\bar{x}_t\|^2.
\end{align}
Then we can obtain
\begin{align} \label{eq:G14}
\frac{1}{\rho^2}\frac{1}{m}\sum_{i=1}^m\|w^i_t-\nabla F(x^i_t)\|^2 +\frac{L^2}{\rho^2}\frac{1}{m}\sum_{i=1}^m\|x^i_t-\bar{x}_t\|^2 + \frac{1}{m}\sum_{i=1}^m\|g^i_t\|^2  \leq \frac{12(\Phi_t-\Phi_{t+1})}{\gamma\rho\eta}.
\end{align}

Since $x^1_0=\tilde{x}^1_0=\cdots=x^m_0=\tilde{x}^m_0$, $z^i_{1,0}=z^i_{2,0}=\cdots=z^i_{n,0}=0$ and $u^i_0=w^i_0=0$ for any $i \in [m]$, we have
\begin{align} \label{eq:G15}
 \Phi_1 & = \mathbb{E}\big[ F(\bar{x}_1) + \alpha_0\frac{1}{m}\sum_{i=1}^m\frac{1}{n}\sum_{k=1}^n\|\nabla f^i_k(x^i_{0})\|^2 + (\lambda_{0}-\frac{9\gamma\eta}{2\rho})\|\overline{\nabla f(x_0)}\|^2
 + \chi_0\frac{1}{m}\sum_{i=1}^m\|\nabla f^i(x^i_0)\|^2 \big] \nonumber \\
 & \leq F(\bar{x}_1)+\big(\alpha_0+\chi_0+\lambda_0-\frac{9\gamma\eta}{2\rho}\big)\frac{1}{m}\sum_{i=1}^m\frac{1}{n}\sum_{k=1}^n\|\nabla f^i_k(x^i_{0})\|^2 \nonumber \\
 & = F(\bar{x}_1)+\big(\frac{18\gamma\eta\beta_0}{\rho}+\frac{18\beta^2_0\gamma\eta\nu^2}{\rho(1-\nu)^2}
 +\frac{3\gamma\eta\nu^2}{\rho(1-\nu)^2}
 +\frac{9\gamma\eta}{2\rho\beta_0}-\frac{9\gamma\eta}{2\rho}\big)
 \frac{1}{m}\sum_{i=1}^m\frac{1}{n}\sum_{k=1}^n\|\nabla f^i_k(x^i_{0})\|^2.
\end{align}

Let $\mathcal{M}^i_t = \|g^i_t\|
+ \frac{1}{\rho}\|\nabla F(x^i_t)-w^i_t\| + \frac{L}{\rho}\|\bar{x}_t-x^i_t\|$.
Then we have
\begin{align}
\mathcal{M}^i_t & = \|g^i_t\|
+ \frac{1}{\rho}\|\nabla F(x^i_t)-w^i_t\| + \frac{L}{\rho}\|\bar{x}_t-x^i_t\| \nonumber \\
& \mathop{=}^{(i)} \|(A^i_t)^{-1}w^i_t\|
+ \frac{1}{\rho}\|\nabla F(x^i_t)-w^i_t\| + \frac{L}{\rho}\|\bar{x}_t-x^i_t\| \nonumber \\
& = \frac{1}{\|A^i_t\|}\|A^i_t\|\|(A^i_t)^{-1}w^i_t\|
+ \frac{1}{\rho}\|\nabla F(x^i_t)-w^i_t\| + \frac{L}{\rho}\|\bar{x}_t-x^i_t\| \nonumber \\
& \geq \frac{1}{\|A^i_t\|}\|w^i_t\|
+ \frac{1}{\rho}\|\nabla F(x^i_t)-w^i_t\| + \frac{L}{\rho}\|\bar{x}_t-x^i_t\| \nonumber \\
& \mathop{\geq}^{(ii)} \frac{1}{\|A^i_t\|}\|w^i_t\|
+ \frac{1}{\|A^i_t\|}\|\nabla F(x^i_t)-w^i_t\| + \frac{L}{\|A^i_t\|}\|\bar{x}_t-x^i_t\| \nonumber \\
& \geq \frac{1}{\|A^i_t\|}\big(\|\nabla  F(x^i_t)\| + L\|\bar{x}_t-x^i_t\|\big),
\end{align}
where the equality $(i)$ holds by $g^i_t = (A^i_t)^{-1}w^i_t$, and the inequality $(ii)$ holds by $\|A^i_t\| \geq \rho$ for all $t\geq1$ due to Assumption \ref{ass:4}.
Then we have
 \begin{align} \label{eq:G16}
  \|\nabla F(x^i_t)\| + L\|\bar{x}_t-x^i_t\| \leq \mathcal{M}^i_t \|A^i_t\|.
 \end{align}

According to the above inequality (\ref{eq:G14}), we have
\begin{align} \label{eq:G17}
\frac{1}{T}\sum_{t=1}^T\frac{1}{m}\sum_{i=1}^m\mathbb{E}[\mathcal{M}^i_t]^2  &
\leq \frac{1}{T}\sum_{t=1}^T\big[\frac{3}{\rho^2}\frac{1}{m}\sum_{i=1}^m\|w^i_t-\nabla F(x^i_t)\|^2 +\frac{3L^2}{\rho^2}\frac{1}{m}\sum_{i=1}^m\|x^i_t-\bar{x}_t\|^2 + \frac{3}{m}\sum_{i=1}^m\|g^i_t\|^2 \big] \nonumber \\
& \leq \frac{1}{T}\sum_{t=1}^T\frac{36(\Phi_t-\Phi_{t+1})}{\gamma\rho\eta} \leq \frac{36(\Phi_1-F^*)}{T\gamma\rho\eta}.
\end{align}
By using Cauchy-Schwarz inequality, we have
 \begin{align} \label{eq:G18}
  \frac{1}{T}\sum_{t=1}^T\frac{1}{m}\sum_{i=1}^m\mathbb{E}[ \|\nabla F(x^i_t)\| + L\|\bar{x}_t-x^i_t\|] & \leq \frac{1}{T}\sum_{t=1}^T\frac{1}{m}\sum_{i=1}^m\mathbb{E}\big[\mathcal{M}^i_t \|A^i_t\|\big] \nonumber \\
  & \leq \sqrt{\frac{1}{T}\sum_{t=1}^T\frac{1}{m}\sum_{i=1}^m\mathbb{E}[\mathcal{M}^i_t]^2} \sqrt{\frac{1}{T}\sum_{t=1}^T\frac{1}{m}\sum_{i=1}^m\mathbb{E}\|A^i_t\|^2}.
 \end{align}

By plugging the above inequalities \eqref{eq:G18} into \eqref{eq:G17}, we can obtain
\begin{align}
  \frac{1}{T}\sum_{t=1}^T\frac{1}{m}\sum_{i=1}^m\mathbb{E}[ \|\nabla F(x^i_t)\| + L\|\bar{x}_t-x^i_t\|] \leq \frac{6\sqrt{\Phi_1-F^*}}{\sqrt{T\gamma\rho\eta}}\sqrt{\frac{1}{T}\sum_{t=1}^T\frac{1}{m}\sum_{i=1}^m\mathbb{E}\|A^i_t\|^2}.
\end{align}

Since $F(x)$ is $L$-smooth, we have
\begin{align}
 \|\nabla F(\bar{x}_t)\| = \|\nabla F(\bar{x}_t) - \nabla F(x^i_t) + \nabla F(x^i_t)\| \leq \|\nabla F(x^i_t)\| + L\|x^i_t - \bar{x}_t\|.
\end{align}
Meanwhile, let $G= \frac{F(\bar{x}_1)-F^*}{\rho\gamma\eta}+\big(\frac{18\beta_0}{\rho^2}+\frac{18\beta^2_0\nu^2}{\rho^2(1-\nu)^2}
 +\frac{3\nu^2}{\rho^2(1-\nu)^2}
 +\frac{9}{2\rho^2\beta_0}-\frac{9}{2\rho^2}\big)
 \frac{1}{m}\sum_{i=1}^m\frac{1}{n}\sum_{k=1}^n\|\nabla f^i_k(x^i_{0})\|^2$. Then we have
\begin{align}
  \frac{1}{T}\sum_{t=1}^T \mathbb{E}\|\nabla F(\bar{x}_t)\|
  & \leq \frac{1}{T}\sum_{t=1}^T\frac{1}{m}\sum_{i=1}^m\mathbb{E}[ \|\nabla F(x^i_t)\| + L\|\bar{x}_t-x^i_t\|] \nonumber \\
  & \leq \frac{6\sqrt{\Phi_1-F^*}}{\sqrt{T\gamma\rho\eta}}\sqrt{\frac{1}{T}\sum_{t=1}^T\frac{1}{m}\sum_{i=1}^m\mathbb{E}\|A^i_t\|^2} \nonumber \\
  & = \frac{6\sqrt{G}}{\sqrt{T}}\sqrt{\frac{1}{T}\sum_{t=1}^T\frac{1}{m}\sum_{i=1}^m\mathbb{E}\|A^i_t\|^2}.
\end{align}

Let $\theta_t=\theta\geq \frac{29\gamma\eta L^2}{6\rho}$ for all $t\geq1$. Let $b=\sqrt{n}$ and $\beta_t = \frac{b}{n}$ for all $t\geq1$, we have
\begin{align}
 H_t = \frac{9}{b\beta_t}+\frac{6\nu^2}{b(1-\nu)^2}+\frac{4n^2\beta^2_t}{b^3}\big(\frac{9}{\beta_t}+
\frac{9\nu^2}{(1-\nu)^2}\big)+\frac{3\nu^2}{(1-\nu)^2} \leq 45 +\frac{45\nu^2}{(1-\nu)^2}.
\end{align}
Then we have $\frac{1}{\sqrt{H_t}}\geq \frac{1}{\sqrt{45 +\frac{45\nu^2}{(1-\nu)^2}}}$,
and
\begin{align}
\min\big(\frac{\rho\sqrt{1-\nu^2}}{2L\gamma\sqrt{6(1+\nu^2)}\sqrt{H_t}},\frac{\sqrt{\rho(1-\nu^2)\theta}}{2 L\sqrt{\gamma(3+\nu^2)}\sqrt{H_t}}\big) \geq \min\big(\frac{\rho\sqrt{1-\nu^2}}{2L\gamma\sqrt{6(1+\nu^2)}},\frac{\sqrt{\rho(1-\nu^2)\theta}}{2 L\sqrt{\gamma(3+\nu^2)}}\big)\sqrt{45 +\frac{45\nu^2}{(1-\nu)^2}}. \nonumber
\end{align}
Thus we can let $\eta= \min\big(\frac{\rho\sqrt{1-\nu^2}}{2L\gamma\sqrt{6(1+\nu^2)}},\frac{\sqrt{\rho(1-\nu^2)\theta}}{2 L\sqrt{\gamma(3+\nu^2)}}\big)\sqrt{45 +\frac{45\nu^2}{(1-\nu)^2}}$ and
$\gamma=\min\big(\frac{\rho(1-\nu^2)}{48\theta},\frac{3\rho(1-\nu^2)\theta}{58L^2}\big)$.
Note that we set $\beta_t=\frac{b}{n}$ for all $t\geq1$, while we can set $\beta_0\in (0,1)$, which is independent on
$T$, $b$ and $n$. Thus we have $G= \frac{F(\bar{x}_1)-F^*}{\rho\gamma\eta}+\big(\frac{18\beta_0}{\rho^2}+\frac{18\beta^2_0\nu^2}{\rho^2(1-\nu)^2}
 +\frac{3\nu^2}{\rho^2(1-\nu)^2}
 +\frac{9}{2\rho^2\beta_0}-\frac{9}{2\rho^2}\big)
 \frac{1}{m}\sum_{i=1}^m\frac{1}{n}\sum_{k=1}^n\|\nabla f^i_k(x^i_{0})\|^2$ is independent on
$T$, $b$ and $n$.

Let $\rho=O(1)$, $\eta=O(1)$ and $\gamma=O(1)$, then we have $G= O(1)$ is independent on
$T$, $b$ and $n$. Given $\sqrt{\frac{1}{T}\sum_{t=1}^T\frac{1}{m}\sum_{i=1}^m\mathbb{E}\|A^i_t\|^2}=O(1)$,
set
\begin{align}
  \frac{1}{T}\sum_{t=1}^T \mathbb{E}\|\nabla F(\bar{x}_t)\|
  \leq \frac{6\sqrt{G}}{\sqrt{T}}\sqrt{\frac{1}{T}\sum_{t=1}^T\frac{1}{m}\sum_{i=1}^m\mathbb{E}\|A^i_t\|^2} \leq \epsilon,
\end{align}
we have $T= O(\epsilon^{-2})$. Since our Algorithm~\ref{alg:2} requires $b$ samples, it obtains a sample complexity of $Tb=O(\sqrt{n}\epsilon^{-2})$.

\end{proof}

\section{Additional Experiment Details}
\label{app:add}
\begin{figure}[ht]
\centering
 \subfloat{\includegraphics[width=0.3\textwidth]{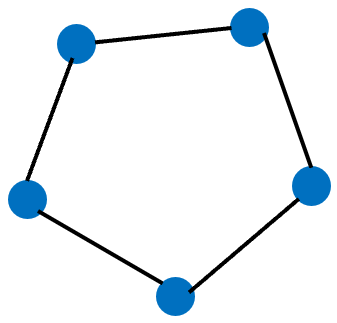}}
 \qquad
 \subfloat{\includegraphics[width=0.44\textwidth]{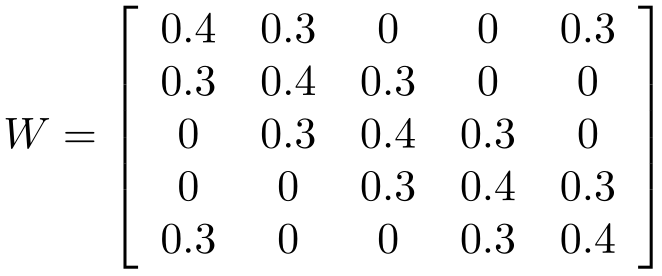}}
  \hfill
\caption{ An illustration of the ring network
with 5 nodes and its mixing matrix.}
\label{fig:ring}
\end{figure}

\begin{figure}[ht]
\centering
 \subfloat{\includegraphics[width=0.3\textwidth]{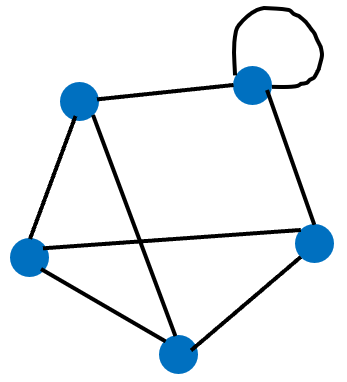}}
 \qquad
 \subfloat{\includegraphics[width=0.48\textwidth]{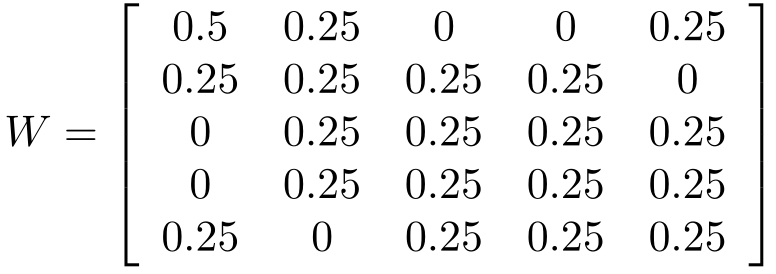}}
  \hfill
\caption{ An illustration of the 3-regular expander network
with 5 nodes and its mixing matrix.}
\label{fig:regular}
\end{figure}

In the experiments, we consider two classical undirected networks that
connect all clients, i.e., the \emph{ring} and \emph{3-regular} expander networks~\citep{hoory2006expander},
illustrated in Figures~\ref{fig:ring} and~\ref{fig:regular}, respectively.

\end{document}